\documentclass[manuscript, screen]{jair}

\setcopyright{cc}
\copyrightyear{2025}
\acmDOI{10.1613/jair.1.17400}

\JAIRAE{Nick Hawes}
\JAIRTrack{Surveys} 
\acmVolume{83}
\acmArticle{12}
\acmMonth{07}
\acmYear{2025}

\RequirePackage[
  datamodel=acmdatamodel,
  style=acmnumeric,
  backend=biber,
  giveninits=true
  ]{biblatex}

\usepackage{subcaption}
\usepackage{tabularx}

\begin{document}

\title{Towards Generalist Robot Learning from Internet Video: A Survey}

\author{Robert McCarthy}
\authornote{Corresponding Author.}
\email{robert.mccarthy.22@ucl.ac.uk}
\orcid{0000-0002-2140-6988}
\affiliation{%
  \institution{University College London}
  \country{United Kingdom}
}

\author{Daniel C.H. Tan}
\email{daniel.tan.22@ucl.ac.uk}
\orcid{0000-0003-1067-8432}
\affiliation{%
  \institution{University College London}
  \country{United Kingdom}
}

\author{Dominik Schmidt}
\orcid{0000-0001-8208-6087}
\email{schmidtdominik30@gmail.com}
\affiliation{%
  \institution{Weco AI}
  \country{United Kingdom}
}

\author{Fernando Acero}
\email{fernando.acero@ucl.ac.uk}
\orcid{0000-0002-5470-8122}
\affiliation{%
  \institution{University College London}
  \country{United Kingdom}
}

\author{Nathan Herr}
\email{nathan.herr.19@ucl.ac.uk}
\orcid{0009-0002-1915-6980}
\affiliation{%
  \institution{University College London}
  \country{United Kingdom}
}

\author{Yilun Du}
\email{yilundu@mit.edu}
\orcid{0000-0001-6792-5946}
\affiliation{%
  \institution{Massachusetts Institute of Technology}
  \country{United States of America}
}

\author{Thomas G. Thuruthel}
\email{t.thuruthel@ucl.ac.uk}
\orcid{0000-0003-0571-1672}
\affiliation{%
  \institution{University College London}
  \country{United Kingdom}
}

\author{Zhibin Li}
\email{alex.li@ucl.ac.uk}
\orcid{0000-0002-6357-7419}
\affiliation{%
  \institution{University College London}
  \country{United Kingdom}
}


\renewcommand{\shortauthors}{McCarthy, Tan, Schmidt, Acero, Herr, Du, Thuruthel, \& Li}

\begin{abstract}
Scaling deep learning to massive and diverse internet data has driven remarkable breakthroughs in domains such as video generation and natural language processing.
Robot learning, however, has thus far failed to replicate this success and remains constrained by a scarcity of available data.
Learning from Videos (LfV) methods aim to address this data bottleneck by augmenting traditional robot data with large-scale internet video.
This video data provides foundational information regarding physical dynamics, behaviours, and tasks, and can be highly informative for general-purpose robots.

This survey systematically examines the emerging field of LfV.
We first outline essential concepts, including detailing fundamental LfV challenges such as distribution shift and missing action labels in video data.
Next, we comprehensively review current methods for extracting knowledge from large-scale internet video, overcoming LfV challenges, and improving robot learning through video-informed training.
The survey concludes with a critical discussion of future opportunities.
Here, we emphasize the need for scalable foundation model approaches that can leverage the full range of available internet video and enhance the learning of robot policies and dynamics models.
Overall, the survey aims to inform and catalyse future LfV research, driving progress towards general-purpose robots.
\end{abstract}



\received{6 June 2024}
\received[revised]{8 November 2024}
\received[accepted]{25 March 2025}

\maketitle


\section{Introduction}

A \textit{generalist robot} should be capable of performing a broad range of physical tasks in unstructured real-world environments.
It should maintain high-level reasoning and planning abilities, along with low-level physical skills, such as dexterous manipulation.
It should adapt to unseen and unexpected scenarios.
It should operate in unstructured settings by perceiving the world through imperfect partial observations (e.g., visual and tactile sensing).

Such a robot would be highly useful in many practical applications (e.g., household or factory tasks).
Nevertheless, obtaining generalist robots remains a grand challenge in robotics.
Classical robotics techniques are insufficient as they cannot handle unseen and unstructured scenarios \cite{krotkov2018darpa}.
More recent machine learning (i.e., robot learning) techniques are more promising \cite{peters2016robot,ibarz2021train}.
These approaches have achieved skillful robot control in narrow settings \cite{zhuang2023robot,kaufmann2023champion}, but generalization to unseen settings remains a challenge.

Robot learning can take inspiration from advances in other domains.
Deep learning has recently provided remarkable improvements in natural language processing \cite{OpenAI2023GPT4TR}, image generation \cite{betker2023improving}, and video generation \cite{videoworldsimulators2024}.
This has been achieved by training expressive architectures \cite{vaswani2017attention} on massive, diverse datasets scraped from the internet.
Here, `scaling laws' have shown performances to consistently and predictably improve with increased computational power and data \cite{kaplan2020scaling}.
The use of diverse internet data has facilitated a transition from task-specific models to monolithic `foundation models' with more general capabilities \cite{OpenAI2023GPT4TR}.

\begin{figure}[h!]
    \centering
    \includegraphics[width=1\textwidth]{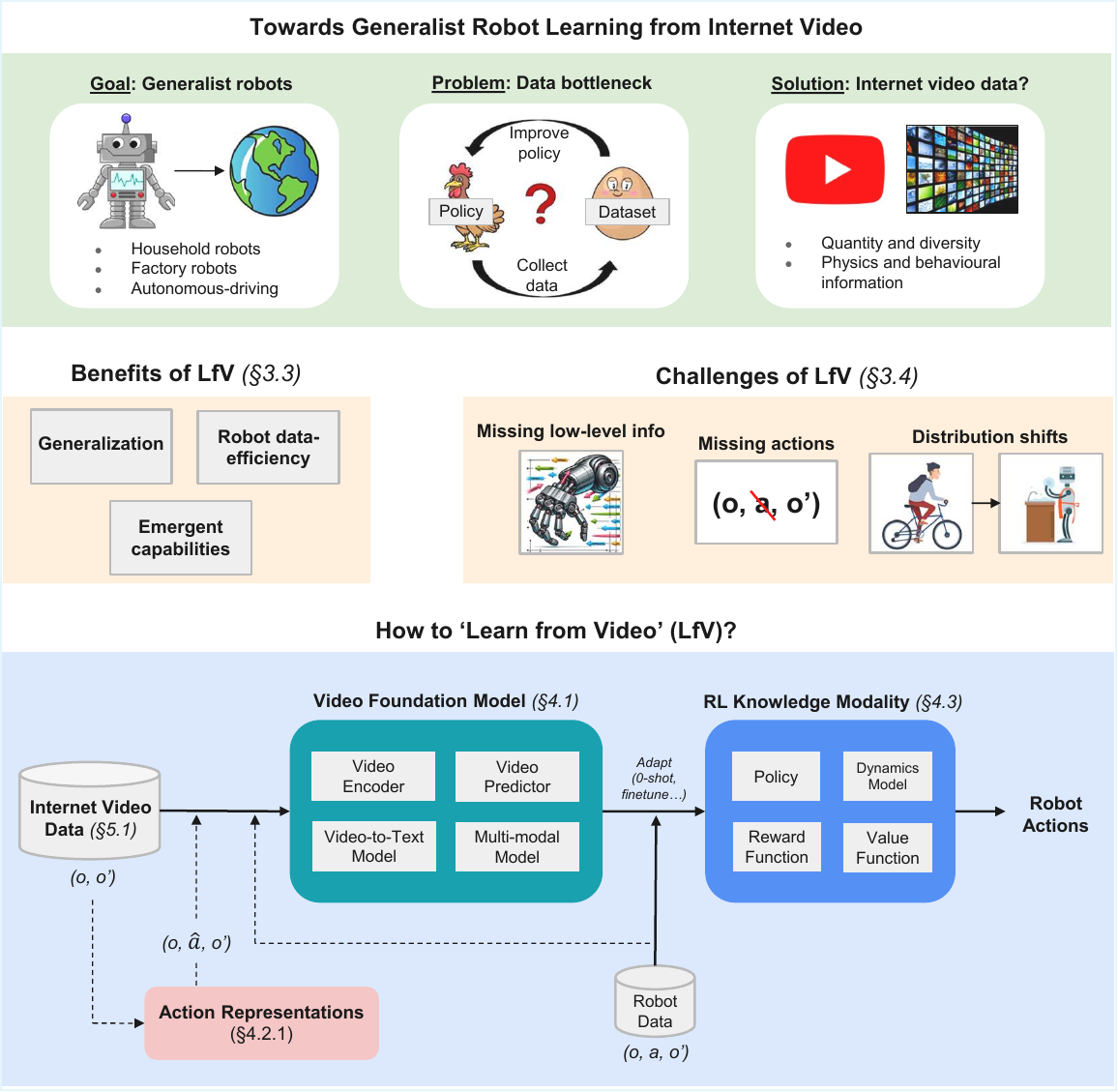}
    \caption{\textbf{An overview of the key concepts and taxonomies in this survey.} 
    The top green box presents the high-level motivation behind LfV. The middle orange boxes highlight the benefits (Section \ref{sec:benefits}) and challenges (Section \ref{sec:challenges}) of LfV.
    The bottom blue box visualises possible components in a pipeline for learning from large-scale internet video, as per the taxonomies presented in the survey.
    Large internet video datasets (Section \ref{sec:datasets}) can be used to pretrain (video) foundation models (Section \ref{sec:video_FMs}).
    These models can be adapted (e.g., via zero-shot transfer or finetuning) into reinforcement learning (RL) `knowledge modalities' \cite{wulfmeier2023foundations} for use in the robot domain (Section \ref{sec:applications}).
    The diagram additionally highlights that action representations (Section \ref{sec:alt_actions}) can be used to mitigate the issue of missing action labels in video.}
    \label{fig:fig_1}
\end{figure}

There is evidence that these deep learning techniques \cite{Brohan2022RT1RT,team2023octo} and scaling laws \cite{Hilton2023ScalingLF,Sartor2024NeuralSL,Bruce2024GenieGI} can transfer to robotics and control.
This offers a path towards more general-purpose robot capabilities.
However, obtaining large-scale data is a challenge in robotics.
Robotics faces a \emph{chicken-and-egg} problem:
data cannot be easily collected due to limited robot capabilities (deploying limited robots to collect real-world data can be ineffective and dangerous), and capabilities cannot easily be improved due to the lack of data (see Figure \ref{fig:fig_1}).

How can we overcome this data bottleneck? One possibility is to use humans to collect robot data \cite{Brohan2022RT1RT,khazatsky2024droid}.
However, this is expensive and can be difficult in tasks requiring skill.
Another option is to leverage simulation \cite{zhuang2023robot,kaufmann2023champion}.
However, simulation comes with issues related to flawed simulated physics and difficulties creating and training on a suitable diversity of simulated environments and tasks. A final option is, like previous deep learning successes \cite{OpenAI2023GPT4TR,betker2023improving}, to leverage the vast quantities of data available on the internet.

While any practical approach may leverage all complementary sources of data, in this survey we focus on learning from internet data.
Specifically, our interest lies in \emph{internet video data}.
Our reasoning here is threefold:

\begin{enumerate}
    \item \textit{Relevant information content.}
    In contrast to internet-scraped text or image data, internet video can uniquely offer information regarding the physics and dynamics of the world, and information regarding human behaviours and actions \cite{yang2024video}.
    Crucially, internet video has coverage over many behaviours and tasks relevant to a general-purpose robot (e.g., household chores).
    
    \item \textit{High in quantity and diversity.}
    There are huge quantities of video data freely available on the internet \cite{sjoberg_2023}.
    Importantly, this data is highly diverse.
    The largest open-source robot dataset \cite{padalkar2023open} pales in comparison, both in terms of quantity and diversity of the data  (see Figure \ref{fig:datasets}).

    \item \textit{Internet video is relatively untapped.}
    The use of real and simulated robot data has been extensively explored \cite{Brohan2022RT1RT,team2023octo,akkaya2019solving,kaufmann2023champion}.
    Meanwhile, leveraging pretrained text and image foundation models has become increasingly common in robotics \cite{Ahn2022DoAI,liang2023code,Brohan2023RT2VM,shah2023lm}.
    However, the use of internet video data in robotics is in more nascent stages.
\end{enumerate}

\begin{figure}[h]
\includegraphics[width=1\textwidth]{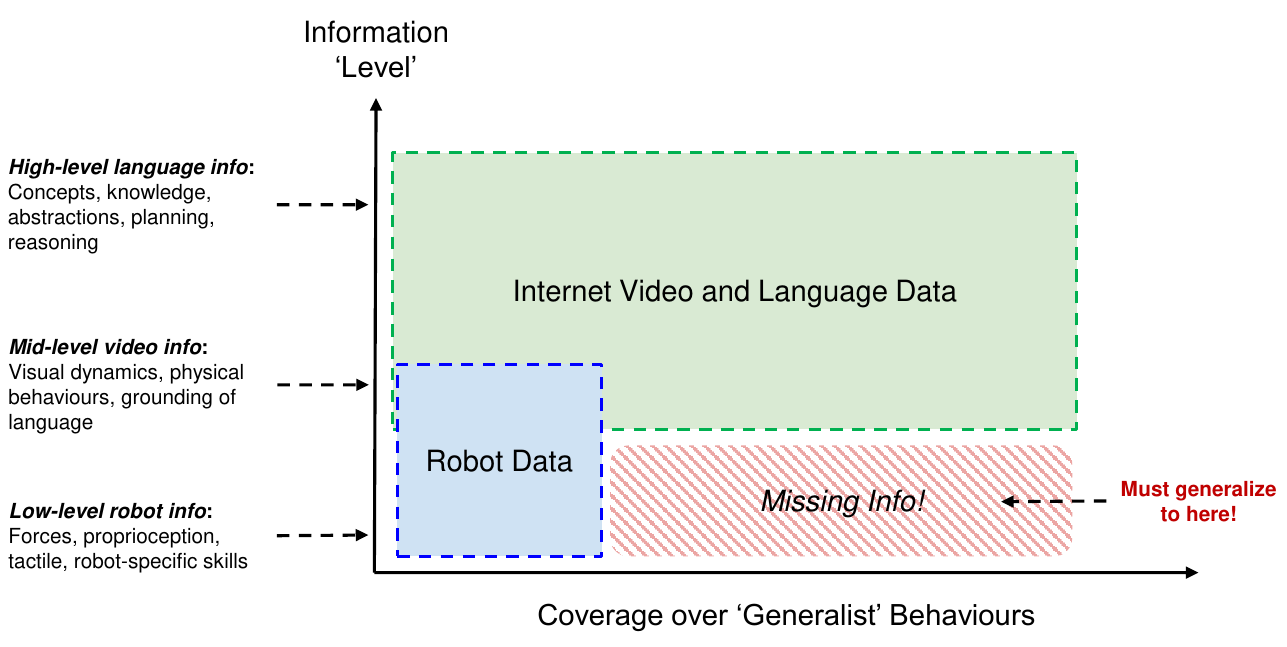}
\centering
\caption{\textbf{Generalization in the Learning from Videos (LfV) setting.} The x-axis indicates the range of behaviours expected from a generalist robot. The y-axis indicates the `levels' of information contained in data. The figure demonstrates that internet data has better coverage over desired behaviours than narrow robot datasets, but lacks crucial low-level information essential to robotics. Generalising beyond the robot data despite this missing low-level information is a key LfV challenge. See Sections \ref{sec:benefits} and \ref{sec:challenges} for further discussion.}
\label{fig:lfv_generalize}
\end{figure}

By leveraging large, diverse internet video data during robot learning, we hope to obtain a number of benefits (Section \ref{sec:benefits}). This includes obtaining improved generalization beyond the available robot data, versus approaches that rely solely on small, narrow robot datasets (see Figure \ref{fig:lfv_generalize}).
Indeed, recent progress in the emerging field of Learning from Videos (LfV) (Section \ref{sec:lfv_setting}) has been promising, demonstrating evidence of this.
This has included work leveraging large-scale video prediction models to act as robot dynamics models \cite{Yang2023LearningIR,Bruce2024GenieGI}, and work leveraging both robot data and internet video to train foundation models for robotics \cite{sohn2024introducing}.

Nevertheless, leveraging internet video is non-trivial and comes with a number of challenges (Section \ref{sec:challenges}).
First, video is a challenging data modality: it is high-dimensional, noisy, and poorly labelled.
Second, video lacks information critical to robotics, including action labels and low-level force and proprioceptive information.
Moreover, there may be various distribution shifts between internet video and the robot domain.
Given these challenges, two key questions for LfV research are: (i) `How to extract relevant knowledge from internet video?' and (ii) `How to apply video-extracted knowledge to robotics?'.

In this survey, we review literature that attempts to answer these questions \footnote{See Figure \ref{fig:fig_1} for more details regarding the structure of the survey.}.
For the first question, we survey video foundation model techniques promising for extracting knowledge from large-scale, heterogenous internet video (Section \ref{sec:video_FMs}).
These techniques may serve as the backbone of scalable LfV approaches.
For the second question, we perform a thorough analysis of literature that explicitly addresses LfV challenges (Section \ref{sec:mitigating}), before reviewing methods that directly utilise video data to aid robot learning (Section \ref{sec:applications}).
We conclude the survey by discussing challenges and opportunities for future LfV research (Section \ref{sec:challenges_opportunities}).
Here, we recommend focusing on scalable approaches that can leverage the full range of available internet video data, and focusing on learning policies and dynamics models to best obtain the benefits of LfV.

Existing LfV research has shown promising signs of life, and this is a field ripe for further progress.
Of course, it is unclear whether scaling deep learning will take us all the way to general-purpose robots (Section \ref{sec:opportunities_other}), and there are other research directions that should be pursued in parallel (Section \ref{sec:background}).
Nevertheless, with sufficient effort and innovation, it seems likely that scaling---with the help of internet video data---can provide significant gains in the near-future.
We hope this comprehensive survey can encourage and inform these future LfV research efforts, serving to facilitate our progress towards the creation of generalist robots.

\section{Background}
\label{sec:background}

This section provides further details on prior literature in deep learning and robot learning.
This serves to add context regarding where LfV approaches (and our survey) fit into the broader research landscape.

\paragraph{Scaling Deep Learning.}
Empirically-derived `scaling laws' have shown deep learning performances to consistently and predictably improve with increased computation, model parameter count, and data \cite{kaplan2020scaling}.
Recent successes in deep learning \cite{Brown2020LanguageMA,OpenAI2023GPT4TR,dubey2024llama,betker2023improving} have largely been driven by such scaling, aided by the use of improved model architectures -- transformer \cite{vaswani2017attention} and diffusion models \cite{Ho2020DenoisingDP}, and the use of self-supervised objectives that allow for learning from unlabelled internet data. 
The term `foundation model' is commonly used to refer to a large, general-purpose model, pretrained on large diverse data, that can solve wide varieties of downstream tasks \cite{bommasani2021opportunities}.

Large language models (LLMs) are transformer models with billions of parameters \cite{dubey2024llama}, pretrained via a simple next-token prediction objective on huge text datasets scraped from the internet \cite{Brown2020LanguageMA,narang2022pathways}.
This pretraining scheme allows the model to learn a `foundation' of declarative and procedural knowledge. The pretrained LLM can be finetuned into a more functional product via supervised learning on high-quality curated data and/or via reinforcement learning from human (or AI) feedback (RLHF) \cite{ouyang2022training,bai2022constitutional}.
Meanwhile, image generation has also benefited from the use of large-scale internet data and from diffusion architectures \cite{Rombach2021HighResolutionIS,betker2023improving}.
Multi-modal vision-language models (VLMs), trained on large-scale internet data to take both images and text as input, have progressed visual understanding \cite{alayrac2022flamingo,OpenAI2023GPT4TR,team2023gemini}.
We detail \emph{video} foundation model efforts in Section \ref{sec:video_FMs}.

\paragraph{Scaling Robot Learning.}
There is a hope that scaling robot learning, in a similar fashion to other domains, may improve the generality of our robots \cite{padalkar2023open}.
However, there are well-known challenges here.
Unlike in video games \cite{schrittwieser2020mastering}, scaling online reinforcement learning (RL) is not a practical solution for real-world robotics:
it is time-consuming, costly, and dangerous \cite{tang2024deep}.
Meanwhile, offline learning solutions \cite{Levine2020OfflineRL,Brohan2022RT1RT} lack suitably large-scale, high-quality robotic datasets \cite{ahn2024autort}.
A number of approaches have been proposed to address these challenges and scale up robotic datasets:

\begin{itemize}

\item \textit{Simulated data.}
The use of simulation is promising for scaling up online learning while avoiding avoiding difficulties associated with the real-world \cite{zhuang2023robot,kaufmann2023champion,nasiriany2024robocasa}.
However, the use of simulation presents a number of issues \cite{bharadhwaj2024position}.
(1) Inaccuracies in low-level physics create a `sim-to-real' gap \cite{zhao2020sim} that must be overcome.
A partial solution here is to employ domain randomization \cite{Tobin2017DomainRF}.
(2) Manually creating a suitable diversity of simulated environments and tasks for generalist robotics is a challenge.
Recent works seek to tackle this using procedural environment generation \cite{deitke2022,team2023human} and foundation-model-assisted environment design \cite{Xian2023TowardsGR,faldor2024omni,wang2023robogen}.
(3) We often lack a policy to collect high-quality data in the simulation.
Solutions here have included
the use of an automated curriculum during RL training \cite{ParkerHolder2022EvolvingCW,liang2024environment}
and collecting data using: specialist policies \cite{tang2024automate}, models with access to privileged simulation information \cite{ha2023scaling,wan2023unidexgrasp++,dalal2023imitating}, or bootstrapping data collection from human demonstrations \cite{mees2022calvin,mandlekar2023mimicgen}.

\item \textit{Real world data.}
Data collected via human teleoperation \cite{Brohan2022RT1RT,khazatsky2024droid} and pooled data from multiple academic labs \cite{padalkar2023open} have been used to train initial robot foundation models \cite{Brohan2022RT1RT,team2023octo}.
Other works have investigated methods for automating data collection to improve scalability \cite{bousmalis2023robocat,ahn2024autort,yang2023robot}.
Several companies have demonstrated evidence of infrastructure suitable for large-scale robot data-collection \cite{sohn2024introducing,jang2024all}.
However, the largest open-source robot dataset \cite{padalkar2023open} is still significantly smaller and less diverse than internet-scale data (see Figure \ref{fig:datasets}).

\item \textit{Internet data.}
Robot learning has benefited from the use of broad internet data.
Image and video data have been used to pretrain visual representations for robotics \cite{Wang2022VRL3AD,Nair2022R3MAU}.
Foundational VLMs and LLMs have been used to help define reward functions for the robot learner \cite{Tam2022SemanticEF,du2023vision,yu2023language,klissarov2023motif}.
LLMs have been employed as high-level planners in long-horizons tasks \cite{Ahn2022DoAI,Huang2022InnerME}.
Some works have finetuned a standard internet-pretrained foundation model into a robot foundation model \cite{Brohan2023RT2VM,kim2024openvla}.
Others have jointly pretrained on both internet data and action-labelled robot data \cite{Reed2022AGA,sohn2024introducing}.
We elaborate on how internet \emph{video} data has been used to aid robot learning in Section \ref{sec:applications}.

\end{itemize}

\paragraph{Other Paths to Generalist Robots.}
Scaling end-to-end deep learning approaches that train a monolithic `foundation' model is one potential path to general-purpose robot capabilities.
Other paths could involve combining learning with the use of stronger inductive biases, more structure, or classical robotics techniques.
These may help address robotics-specific challenges and potential limitations of scaling end-to-end deep learning (\textcite{ieee_solve_robotics}, Section \ref{sec:opportunities_challenges}).
Modular approaches have been proposed which use specialist models for different skills or tasks \cite{yang2020multi,Lee2021AdversarialSC},
or use a hierarchy to separate out high-level planning from low-level control \cite{Ji2022HierarchicalRL,lecun2022path,yuan2023hierarchical}.
Learning has been combined with explicit planning algorithms \cite{mishra2023generative,huang2023voxposer} and environment representation techniques \cite{Chaplot2020NeuralTS,Marza2022MultiObjectNW,peng2024q}.
Structure related to the robot embodiment \cite{Wang2018NerveNetLS,sferrazza2024body} or the physics of the system \cite{lutter2019deep,levy2024learning} can be leveraged, while some approaches carefully choose and process their observation and action spaces \cite{xu2024manifoundation,chisari2024learning}.
As these `alternative' approaches and deep learning are often combined, they can be complementary: advances in scaling deep learning can benefit alternative approaches, and vice versa (e.g., alternative approaches can collect data for monolithic deep learning models; \textcite{dalal2023imitating}).

\paragraph{Related Surveys.}
\textcite{Ravichandar2020RecentAI,gavenski2024imitation} survey imitation learning methods; \textcite{prudencio2023survey} review offline RL methods; while \textcite{wulfmeier2023foundations} highlight the promise of transferring knowledge from a source to a target domain in RL.
\textcite{Yang2023FoundationMF,hu2023toward} review the use of foundation models in decision-making and robotics.
Relevant works in the video ML literature have included reviews of self-supervised video learning \cite{schiappa2023self}, foundation models for video understanding \cite{madan2024foundation}, and LLMs for video understanding \cite{tang2023video,zhang2024mm}.

\textcite{Torabi2019RecentAI} review imitation learning from observational data, while \textcite{yang2024video} advocate for the use of video (and video generation) as a unified interface to absorb internet knowledge and represent diverse tasks.
\textcite{eze2024learning} review video-based learning approaches for robot manipulation.
In contrast to this work, our survey places a stronger focus on approaches that: (i) can scale to large, diverse internet video data, and (ii) can yield general-purpose robotic capabilities.

\section{Preliminaries}
\label{sec:preliminaries}

In this section, we introduce useful preliminary concepts related to LfV.
First, the reinforcement learning (Section \ref{sec:prelims_rl}) and LfV (Section \ref{sec:lfv_setting}) settings are formalized.
Next, we perform a deeper dive into the potential benefits of LfV (Section \ref{sec:benefits}) and the fundamental challenges that come with learning from videos in robotics (Section \ref{sec:challenges}).

\subsection{Reinforcement Learning (RL)}
\label{sec:prelims_rl}

\paragraph{Formalism.}

In reinforcement learning (RL), an agent observes its environment, takes an action, and receives a reward after the state of the environment changes. This can be formalised as a Markov Decision Process (MDP) consisting of the state space $\mathcal{S}$, the action space $\mathcal{A}$, and the transition probability $p(s_{t+1}|s_t,a_t)$ of reaching state  $s_t+1$ from state $s_t$ when executing action $a_t$. The agent’s behaviour is given in terms of a policy $\pi(a_t|s_t)$. When aspects of the state cannot be observed, the environment is termed a Partially Observable MDP (POMDP) and the agent only has access to observations $o_t \in \mathcal{O}$ that are partial mappings of the state $o_t = f(s_t)$. Unless stated otherwise, we will use always a POMDP setting.

The agent aims to maximise its sum of discounted future rewards, commonly referred to as its `returns'. This is captured by the objective in Equation \ref{eq:rl_returns}:

\begin{equation}
    J(\pi) = \mathbb{E}_{[p(s_0),p(s_{t+1}|s_t,a_t),\pi(a_t|s_t)]_{t=0...T}}\left[\sum_{t=0}^{\infty} \gamma^t r_t \right]
    \label{eq:rl_returns}
\end{equation}

where $\gamma$ is the discount factor, $r_t = r(s_t,a_t)$ is the reward, and $\rho(s_0)$ is the initial state distribution. The optimal policy $\pi^*(a_t|s_t)$ maximises Equation \ref{eq:rl_returns}.

In the LfV setting, the POMDP is some real-world environment and $\pi(a_t|s_t)$ controls a robot within this environment.

\paragraph{Knowledge Modalities.}
An RL Knowledge Modality (KM) is some function learned from data that represents specific types of RL-related knowledge (see the work of \textcite{wulfmeier2023foundations} for more details).
We make use of the notion of a KM throughout our paper, including to help define our main taxonomy of LfV methods in Section \ref{sec:applications}.
We will commonly refer to the following KMs:

\begin{itemize}
    \item A \textit{policy, $\pi(a_t|s_t)$} is a mapping from states to actions directly describing the agent’s behaviour.
    \item A \textit{state or state-action value function, $V_{\pi}(s_t)$ or $Q_{\pi}(s_t,a_t)$} maps from states or state-action pairs to the expected future return when acting under a particular policy $\pi$.
    \item A \textit{dynamics model, $p(s_{t+1}|s_t, a_t)$} predicts the next state given the current state and action.
    \item A \textit{reward model, $r(s_t, a_t)$} predicts the reward provided by the environment for a specific state-action pair.
\end{itemize}

\paragraph{Transfer Mechanisms.}
\label{sec:foundations_transfer}

RL KMs can be pretrained on a source MDP or dataset and can then be subsequently transferred to a target MDP \cite{wulfmeier2023foundations}.
In the case of LfV, this involves learning a KM from a video dataset and adapting it to the robot MDP (usually via the use of robot data).
We will commonly refer to the following transfer mechanisms in our paper (see the work of \textcite{wulfmeier2023foundations} for more details):
\emph{Generalisation and zero-shot transfer} involves using a pretrained KM directly in the target setting, without further fine-tuning or adaptation \cite{Escontrela2023VideoPM}.
We use \emph{Fine-tuning} and \emph{Representation transfer} interchangeably in this survey to refer to methods that train a model in the robot domain that is partially composed of parts of a pretrained KM \cite{Nair2022R3MAU,Majumdar2023WhereAW}.
\emph{Hierarchy: conditioning} involves using a pretrained KM to condition a new KM being trained in the target MDP \cite{Schmidt2023LearningTA,Wang2023MimicPlayLI,Wen2023AnypointTM}.

\subsection{Learning from Videos (LfV): Formalism and Assumptions}
\label{sec:lfv_setting}

\paragraph{Formalism.}
We assume an LfV method has access to a video dataset $D_\text{video}$.
We denote a video clip as $\tau = (o_0, o_1, ...,o_T)$, where $\tau$ is the full clip and each $o$ is an RGB image observation.
Optionally, $D_\text{video}$ may come paired with language annotations or annotations of other modalities.
We also assume access to a robot dataset $D_\text{robot}$. This contains trajectories of transition tuples $(o_t, a_t, r_t, o_{t+1})$, where $r_t$ may be missing, and where each $o$ contains an image observation and may also contain other information (e.g., tactile information).
The goal of LfV is to leverage our combined LfV dataset, $D_\text{LfV} = \{D_\text{robot}, D_\text{video}\}$, to obtain an improved $\pi(a_t|s_t)$ versus when learning from $D_\text{robot}$ alone.

\paragraph{LfV in the Generalist Robot Setting.}

Though $D_\text{video}$ may come from any source, in this survey we are interested in methods that can leverage large-scale video data gathered from the internet.
We will generally assume that such a dataset consists primarily of videos of humans and has broad coverage over many common physical tasks that humans perform.
We assume the robot must perform tasks that, to some extent, seen in $D_\text{video}$.

We loosely define a `generalist'/general-purpose robot as one that can perform a diverse range of every-day physical human tasks in unstructured real-world settings.
Such settings are POMDPs, where the robot must rely heavily on visual observations.
Throughout the survey, unless stated otherwise, we assume the general-purpose robot has an embodiment and affordances similar to those of a human.
We subsequently assume the robot can execute tasks in a physically similar manner to how a human would.

Under the above assumptions, internet video can be particularly informative to the robot:
it provides extensive information regarding how relevant embodiments can perform relevant tasks and behaviours.


\paragraph{Limitations of these assumptions.}
First, for certain robot tasks, embodiments very different from that of a human may prove more effective --- e.g., aerial drones \cite{Ali2022UnmannedAV} or quadrupedal robots \cite{Biswal2020DevelopmentOQ,yang2020multi} that can better traverse treacherous terrains.
Second, $D_\text{video}$ may not have good coverage over all the tasks we want the robot to perform: some important actions and tasks may be underrepresented in the video dataset --- e.g., specialized industrial tasks \cite{Urrea2025}.
Additionally, the robot may need to perform tasks not commonly performed by humans --- e.g., planetary exploration \cite{Schilling1995MobileRF}.
Nevertheless, even in these cases, internet video can still provide generally information regarding the world and physical behaviour.

\subsection{Benefits of LfV}

\label{sec:benefits}

Robotic datasets are expensive to acquire and  thus are currently task-specific or relatively narrow \cite{padalkar2023open}.
In contrast, diverse video data is freely available in vast quantities on the internet.
In this section, we outline the specific benefits we hope to obtain from methods that leverage this video data.


    \paragraph{Generalization beyond $D_\text{robot}$.}
    LfV offers the exciting possibility of improved generalization beyond narrow robot datasets, $D_\text{robot}$, to the full space of tasks covered in the video dataset, $D_\text{video}$ \cite{eze2024learning}.
    An argument for why this could be the case is as follows.
    First, consider a $D_\text{robot}$ that has good coverage over the low-level skills or `atomic' actions required from the robot (e.g., specific grasping motions or locomotion skills) \cite{chen2024rh20t}.
    Now consider a task unseen in $D_\text{robot}$ but seen in $D_\text{video}$.
    Our combined LfV dataset ($D_\text{LfV} = \{D_\text{robot}, D_\text{video}\}$) may contain most information required to complete the task.
    $D_\text{robot}$ provides information regarding how to execute the low-level robot skills,
    whilst the human actions in $D_\text{video}$ provides higher-level information regarding how to complete the overall task (e.g., visual information regarding required movements and steps).
    Thus, a suitable LfV method may be capable of leveraging $D_\text{video}$ to generalise beyond $D_\text{robot}$ and solve the task.
    There is preliminary evidence of such generalization in LfV-related literature \cite{Brohan2023RT2VM,Du2023LearningUP,Wu2023UnleashingLV,Wang2023MimicPlayLI}.
    Figure \ref{fig:lfv_generalize} explores this generalization setting in more detail.

    \paragraph{Emergent Capabilities.}
    Learning from internet-scale video may yield capabilities qualitatively beyond what can be obtained when learning only from a narrow $D_\text{robot}$.
    We expect this for two reasons.
    First, in other domains large quantities of internet data have allowed for unexpected `emergent' capabilities \cite{radford2019language,Brown2020LanguageMA}.
    Second, diverse internet video paired with language annotations offers a path towards stitching together the lower-level knowledge obtained from robot and video data with the rich abstractions and world knowledge that can be obtained from text data \cite{OpenAI2023GPT4TR}.

    \paragraph{Improvements in-distribution of $D_\text{robot}$.}
    Finally, we also expect video data to yield improvements in tasks that are in-distribution of the robot dataset. Utilising a large video dataset can allow for improved data-efficiency with respect to $D_\text{robot}$ \cite{Nair2022R3MAU}. Additionally, LfV approaches may obtain higher absolute task performance (e.g., higher success rates) in settings in-distribution of $D_\text{robot}$, versus non-LfV approaches \cite{Wu2023UnleashingLV}.
    

\begin{figure}[h]
\includegraphics[width=0.9\textwidth]{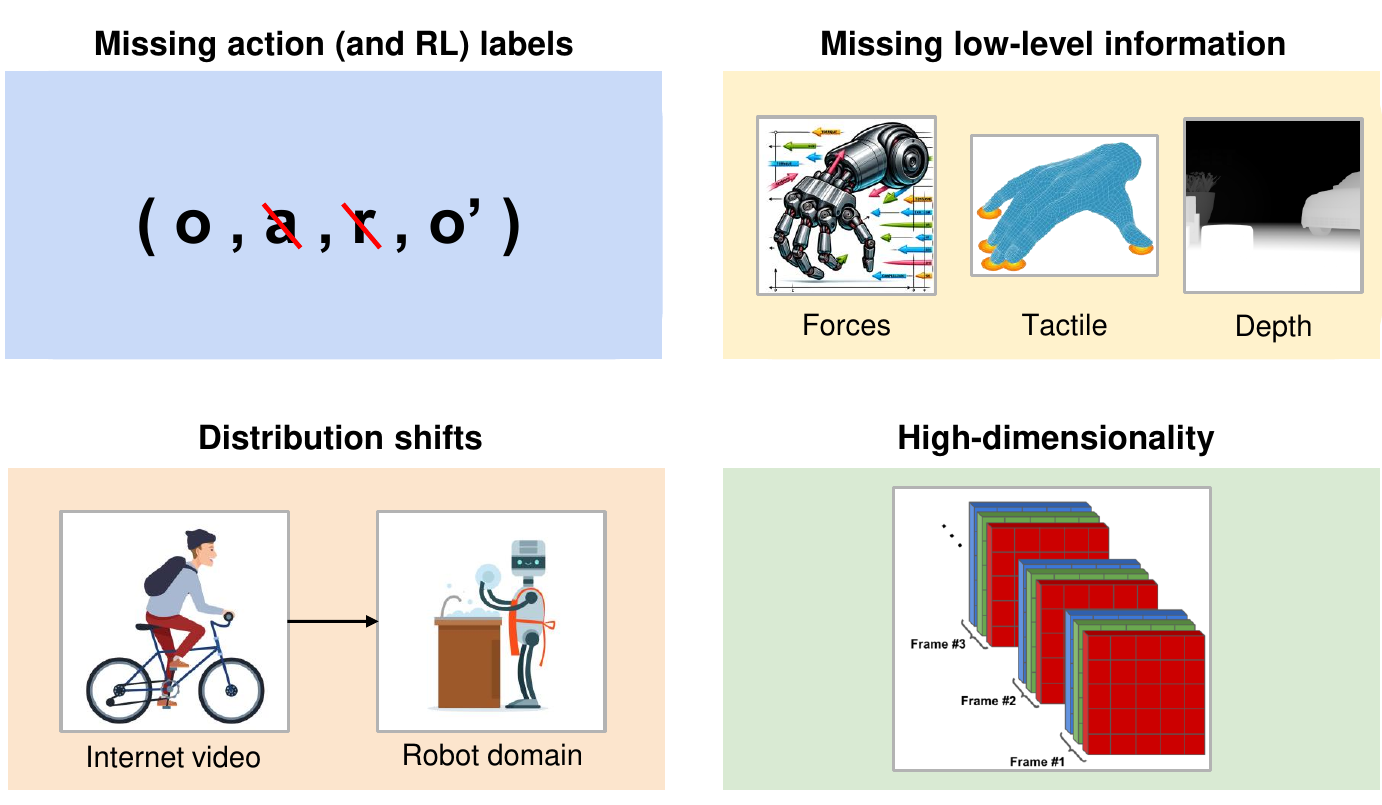}
\centering
\caption{\textbf{Key challenges in LfV} (see Section \ref{sec:challenges}) are visualised, including: missing (action and low-level) information in video, LfV distribution shifts, and the high-dimensional nature of video data.}
\label{fig:challenges}
\end{figure}

\subsection{Fundamental Challenges}
\label{sec:challenges}

Learning from internet video for robotics comes with some fundamental challenges.
These render LfV a non-trivial problem.
Awareness of these challenges is crucial for understanding the motivation behind existing LfV research (Section \ref{sec:LfV_methods}) and the future promising research directions we highlight (Section \ref{sec:challenges_opportunities}).
Note, none of these challenges have yet been fully solved.



\paragraph{Missing Action Labels.} Raw video data lacks the action labels required by existing imitation \cite{Brohan2022RT1RT} and offline RL \cite{chebotar2023q} methods for learning from demonstrations. Moreover, adding low-level robot-action labels retrospectively to internet video data is non-trivial (see Section \ref{sec:alt_actions}): the unrestricted action space of heterogeneous video data often cannot map cleanly to the robot action-space \cite{Ye2024LatentAP}.
It is also challenging to retrospectively add other missing RL-relevant metadata to raw video, such as reward labels which can inform on the quality of the data (see Section \ref{sec:reward_funcs}), goal labels which can be useful for goal-conditioning (see Section \ref{sec:value_funcs}), or end-of-episode labels.

\paragraph{Distribution-shift.} There may be significant distributional shifts between an internet video dataset and the downstream robot domain.
This can include differences in physical embodiments, camera viewpoints, tasks, and environments.
Humans in videos may perform behaviours which are sub-optimal or irrelevant to the downstream robot.
Certain useful tasks and behaviours may be underrepresented in internet video (e.g., coverage over specific factory tasks may be poor).
These shifts present a challenge to deep learning methods.

\paragraph{Missing Low-level Information.}
For certain skillful or dexterous behaviours, robots require low-level percepts such as tactile sensing, forces, proprioception, or depth sensing. This crucial low-level information is not explicitly available in internet video. A key challenge in LfV is to obtain generalization beyond $D_\text{robot}$ despite the missing low-level information in $D_\text{video}$ (see Figure \ref{fig:lfv_generalize}). 

\paragraph{Controllability, Stochasticity, and Partial Observability.}
In unlabelled video, it can be difficult to disentangle which parts of a transition are affected (i.e., controlled) by a specific agent's actions or which are due to the external environment or noise \cite{Horan2021WhenIU}.
This can be problematic for methods that attempt to extract action information from video (Section \ref{sec:alt_actions}). Furthermore, the stochastic nature and partial-observability of the underlying environments in video can make accurate video prediction a challenge \cite{babaeizadeh2017stochastic}.

\paragraph{High-dimensionality, Noise, and Redundancy.}
Methods that learn from or generate video data are typically computationally demanding due to the high-dimensional nature of video data. Additionally, video can contain significant noise and redundant information. These characteristics make it challenging to extract meaningful information and representations from video data \cite{bardes2023v}.


\section{Methods}
\label{sec:LfV_methods}

This section reviews methods for learning from video data for robotics.
We first review video foundation models as a general-purpose means for extracting knowledge from internet video (Section \ref{sec:video_FMs}), noting that advances in these approaches will directly serve advances in scalable LfV methods.
We then review common methods used to tackle key LfV challenges (Section \ref{sec:mitigating}).
Finally, we move to our primary taxonomy of the LfV literature, categorizing LfV methods according to which RL knowledge modality benefits from the use of video data (Section \ref{sec:applications}).

\begin{figure}[h]
    \centering
    \includegraphics[width=\textwidth]{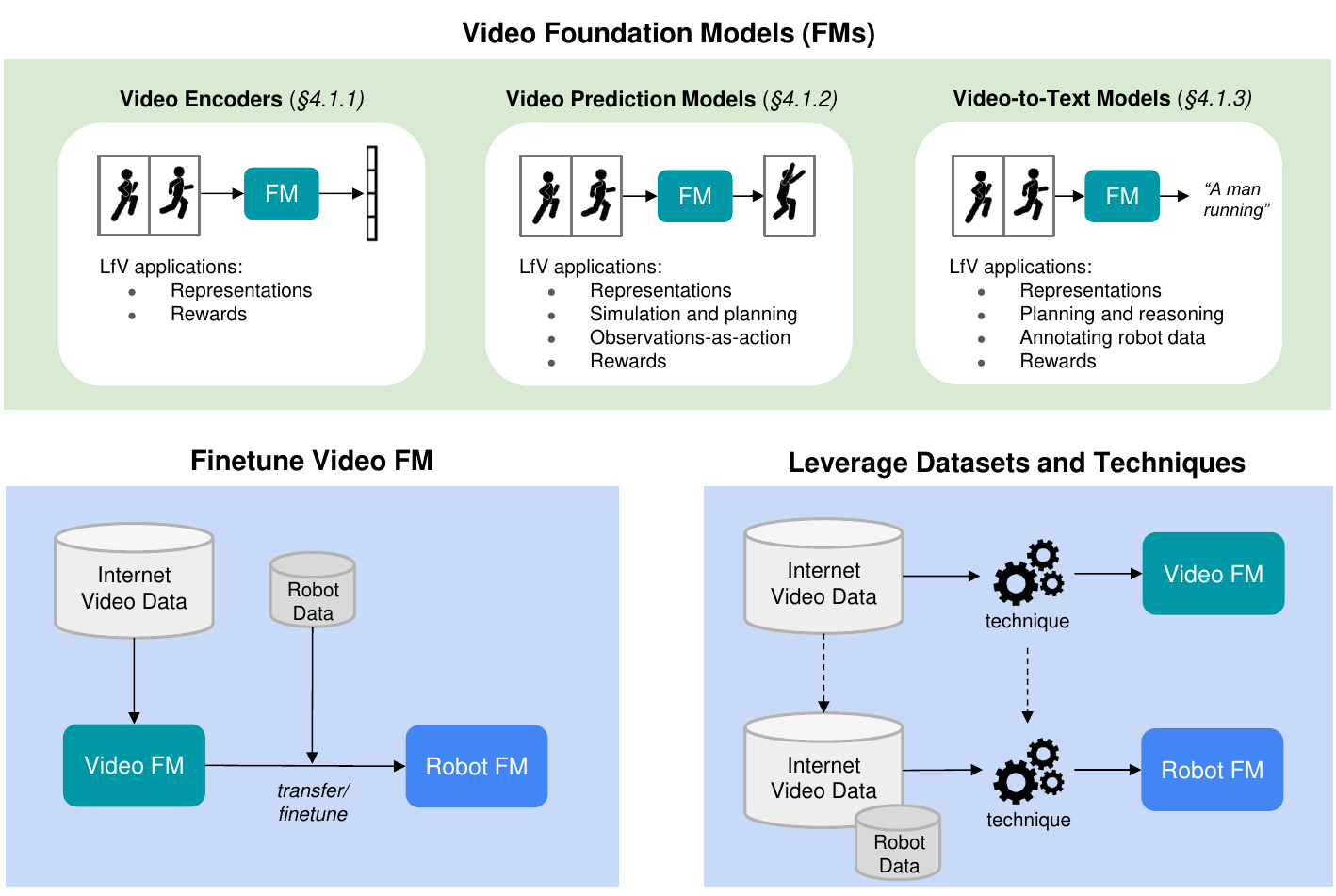}
    \caption{\textbf{Video Foundation Modelling for LfV} (see Section \ref{sec:video_FMs}). The top green box outlines different categories of video foundation models and their applications to robotics. The bottom blue boxes illustrate ways video foundation models can contribute to LfV: (left) pretrained video foundation models can be finetuned into robot foundation models; (right) video foundation model techniques and datasets can be used to train robot foundation models.}
    \label{fig:video_fms}
\end{figure}

\subsection{Extracting Foundational Knowledge from Internet Video}
\label{sec:video_FMs}

Internet video lacks action labels and has various distribution shifts with the robot domain.
Given these LfV challenges (see Section \ref{sec:challenges}), a key LfV question is:
\textit{How to extract robotics-relevant knowledge from large-scale, heterogenous internet videos?}
The video foundation model (FM) literature offers potential answers here.

Video FMs are large deep learning models trained on large internet-scraped video datasets.
From this data, they extract general knowledge and capabilities useful in a wide range of downstream settings \cite{madan2024foundation}.
This includes robotics-relevant knowledge related to semantics, physics, and behaviours.
As such, video FMs and corresponding techniques can be highly applicable for the following LfV use-cases (see Figure \ref{fig:video_fms}):

\begin{itemize}
    \item \textit{Using pretrained models:}
    A pretrained video FM can be transferred or adapted (e.g., finetuned) for downstream robot applications, allowing the robot to leverage its broad world knowledge.
    For example, a video prediction model can be adapted into a robot dynamics model \cite{Yang2023LearningIR}.
    \item \textit{Using techniques and datasets:}
    Techniques and datasets originally used for video FMs can be customized for robotics purposes.
    For example, a robot FM can be trained on both video and robot data, using techniques inspired by video FMs \cite{sohn2024introducing}.
    These techniques can allow LfV to better extract representations and knowledge from internet video, thus improving the generality of the robot.
\end{itemize}

The promise of these video FM use-cases is backed up by prior work demonstrating that text and image FMs can benefit robotics.
Scalable architectures transferred from these other domains have allowed robot techniques to better fit large, diverse robot data \cite{Brohan2022RT1RT,team2023octo}.
Knowledge and representations from text and image FMs have been transferred to the robot domain to improve capabilities, generalization, and common-sense \cite{Ahn2022DoAI,Brohan2023RT2VM,kim2024openvla}.
The use of video FMs may lead to further improvements due to the unique robotics-relevant information video data can offer. 



Video foundation models and techniques will likely be key driver to the success of scalable LfV approaches. As such, this section provides LfV researchers with essential information regarding potential use-cases and the lower-level workings of these techniques.
We loosely categorise the video foundation model literature into: video encoding methods (Section \ref{sec:video_FMs_ssl}); video prediction methods (Section \ref{sec:video_prediction}); and video-to-text generation methods (Section \ref{sec:video_FMs_to-text}).

\subsubsection{Video Encoders}
\label{sec:video_FMs_ssl}

Foundational video encoders ($p(z|\tau)$, where $z$ is a representation and $\tau$ is a video clip) can provide rich and robust video representations for downstream robotic applications.
Specifically, they can useful as follows:

\begin{itemize}
    \item \textit{Representations:} LfV representation transfer approaches take a pretrained video encoder and finetune it into an RL KM (e.g., into a policy, see Section \ref{sec:policy}).
    \item \textit{Rewards:} LfV approaches have used frozen pretrained video representations to help define robot reward functions \cite{fan2022minedojo,Sontakke2023RoboCLIPOD,Nair2022R3MAU} (see Section \ref{sec:reward_funcs}).
\end{itemize}

\paragraph{Methods.}

We now provide a general overview of representation learning from video, before providing more details on state-of-the-art (SOTA) approaches:

\begin{itemize}

\item \textit{Overview of video representation learning.}
Let $\bar X$ be a video clip $X$ paired with some corresponding labels and data modalities.
Most methods for learning video encoders can be framed as first deriving a ‘label’ $Y$ from $\bar X$ and then using $Y$ to define a learning objective.
$Y$ could be some transform of the video $X$ itself (e.g., a masked version of the video, see \textcite{Wang2023VideoMAEVS}).
Otherwise, $Y$ could be derived from additional information paired with the video such as: labels (e.g., object masks and bounding boxes, see \textcite{ding2023mose}); meta-data \cite{schiappa2023self}; or corresponding data of another modality (e.g., language annotations, see \textcite{Grauman2021Ego4DAT}).
Once $Y$ is obtained, there are a number of ways it can be used.
Prediction objectives predict $Y$ from $X$.
For example, predicting the next frame from the previous frame (i.e., video prediction) \cite{gao2022simvp,jang2024visual}.
Joint-embedding approaches first embed $X$ and $Y$ before calculating the loss \cite{oord2018representation,Xu2021VideoCLIPCP,bardes2023v}. 
Here, contrastive learning approaches have been heavily explored \cite{schiappa2023self}.

\item \textit{SOTA Approaches.}
The literature has proposed a number of approaches for learning rich spatio-temporal representations from video in a \textit{scalable} manner.
Video-text losses leverage text annotations to learn semantic representations from video.
This can include: video-text contrastive losses \cite{Xu2021VideoCLIPCP,Zhao2024VideoPrismAF,Papalampidi2023ASR,Li2023UnmaskedTT,Wang2023InternVidAL}, video-text matching \cite{Li2023UnmaskedTT}, masked language modelling \cite{Li2023UnmaskedTT}, and video-to-text losses \cite{Yan2022VideoCoCaVM,Papalampidi2023ASR}.
Auto-encoding approaches have also proven popular.
State-of-the-art methods have used masked auto-encoding (MAE) \cite{Tong2022VideoMAEMA,Wang2023VideoMAEVS,Girdhar2022OmniMAESM} and vector-quantized auto-encoding (VQ-AE) \cite{Seo2022HARPAL,Yan2021VideoGPTVG,Yu2022MAGVITMG,Villegas2022PhenakiVL,Bruce2024GenieGI}.
A number of works have explored the use of student-teacher distillation losses \cite{Wang2022MaskedVD,Zhao2024VideoPrismAF,Li2023UnmaskedTT}.
Joint-embedding prediction architectures (JEPA) have been proposed to mitigate issues related to video's high-dimensionality and noise in video \cite{bardes2023v}.

\end{itemize}

\subsubsection{Video Prediction Models}

\label{sec:video_prediction}  

This section is primarily concerned with models that can perform next-frame video prediction: $p(o_{t+1}|o_{t-k:t})$.
Also relevant are models that can more generally perform some form of conditional video generation: $p(\tau | c)$ (where $\tau$ is a video clip and $c$ is some conditioning information) \cite{yang2024video}.
Via training on internet video, such models can learn information regarding the world dynamics and human behaviours.
They can thus be useful for downstream robotic applications in the following ways:

\begin{itemize}
    \item \textit{Dynamics:} A video prediction model can be adapted into a robot dynamics model $p(o_{t+1}|o_{t-k:t},a_t)$ to serve as a planner \cite{Du2023VideoLP} or simulator \cite{Yang2023LearningIR} (see Section \ref{sec:dynamics}).
    \item \textit{Policies:} The video prediction objective implicitly models the distribution of behaviours in the video dataset \cite{Escontrela2023VideoPM}. As such, video prediction models can act as policies by generating `observations-as-actions', videos of future behaviours the robot should execute \cite{Du2023LearningUP} (see Section \ref{sec:policy}).
    \item \textit{Representations:} Due to the relevant information they represent, video predictors can be used for LfV representation transfer approaches \cite{Wu2023UnleashingLV}.
    \item \textit{Rewards:} Finally, a reward signal can be defined that encourages the robot to match the behaviour expected by the video predictor \cite{Escontrela2023VideoPM} (see Section \ref{sec:reward_funcs}).
\end{itemize}

\paragraph{Methods.}
Diffusion models, autoregressive transformers, and masking transformers have proven to be particularly effective and scalable architectures in recent years.

\begin{itemize}

\item \textit{Diffusion models} can easily model continuous output spaces and can sample multiple frames in parallel.
However, their sampling speeds can be slow and long video generation remains a challenge \cite{yang2024video}.
Computational efficiency can be aided by performing the diffusion process in a learned latent space \cite{videoworldsimulators2024,bar2024lumiere,blattmann2023stable} rather than in pixel space \cite{ho2022video,singer2022make,ho2022imagen}.
Many diffusion video prediction models augment their training data by making use of image data (which can come in higher quality and quantity than video data) \cite{ho2022video,ho2022imagen,bar2024lumiere,ge2023preserve,blattmann2023align,dai2023emu}.
The diffusion video prediction pipeline can often involve several intermediate steps, including: key-frame generation, interpolation between key-frames, and spatial super-resolution upsampling \cite{ho2022imagen,ge2023preserve,zhou2022magicvideo}.
However, \textcite{bar2024lumiere} generate the entire temporal duration in a single forward pass, obtaining improved global temporal consistency.

\item \textit{Autoregressive and masking transformers} tend to make predictions in a learned latent-space
\cite{Yan2021VideoGPTVG,Yu2023LanguageMB,ge2022long,kondratyuk2023videopoet}.
Predicting in latent space can improve computational efficiency and can mitigate issues related to pixel-level noise and redundancy \cite{Yan2021VideoGPTVG}.
The latent space is generally pretrained via vector-quantized auto-encoding (VQ-AE), providing a discrete token-space suitable for the transformer architecture.
Versus autoregressive methods \cite{Yan2021VideoGPTVG,Hu2023GAIA1AG,Bruce2024GenieGI}, masking transformers train the model to decode masked tokens in parallel \cite{Yu2022MAGVITMG,yu2023language,gupta2022maskvit}.
This is usually achieved via the use of MaskGIT decoding \cite{Chang2022MaskGITMG}.
Masked models are more computationally efficient and do not suffer from the `drifting' effect that can occur in auto-regressive models \cite{yang2024video}.

\item \textit{Conditioning information} can simplify the video prediction problem and allow for more control over generated videos.
Such conditioning is valuable for downstream robotics: it can allow us to simulate the effects of different `action' strategies. \textcite{yang2024video} outline various popular conditioning schemes for video generation models $p(\tau|c)$. 
In particular, language conditioning can allow for flexible and intuitive control over generated video, at varying levels of detail \cite{videoworldsimulators2024,bar2024lumiere,kondratyuk2023videopoet}.
Amongst other possibilities, video generation can also be conditioned on future images (i.e, video in-filling, see \textcite{hoppe2022diffusion}) or on action representations \cite{Bruce2024GenieGI}.
We outline various action representations that may be suitable here in Section \ref{sec:alt_actions}.

\end{itemize}

\subsubsection{Video-to-Text Models}
\label{sec:video_FMs_to-text}


This section details to models with video-to-text capabilities, $p(\text{text}|\tau)$. A capable video-to-text model can perform, for example, video question-answering or video summarization. Such a video-to-text foundation model could be valuable to robotics in the following ways.

\begin{itemize}
    \item \textit{High-quality representations:} A capable video-to-text model will have robust, high-quality video representations. Compared to a video-only model, it will have improved high-level semantic representations. Compared to an image-to-text model, it will have improved temporal-dynamic representations. Robotics can bootstrap from such models via representation transfer \cite{Brohan2023RT2VM}, or by adding robot data directly into the model's pretraining corpus \cite{Reed2022AGA}.
    
    \item \textit{Grounded reasoning and planning:} LLMs have proven useful as planning modules in robotics \cite{Ahn2022DoAI}, but their lack of grounding in the physical world is limiting. In contrast, video-to-text models can perceive the environment through information-rich video, allowing for improved and closed-loop reasoning and planning.
    
    \item \textit{Annotating robot data:} High-quality language annotations can provide valuable conditioning information in many ML domains \cite{betker2023improving,videoworldsimulators2024}. Robotics is no different \cite{team2023octo}. Capable video-to-text models could serve as useful language annotators for robotic datasets \cite{blank2024scaling}.
    
    \item \textit{Rewards:} A sufficiently capable video-to-text model can provide reward or value estimates for a robot learner. For example, this could be achieved through a visual-question answering framework \cite{du2023vision}, or via an RL-from-AI-feedback framework \cite{klissarov2023motif} (see Section \ref{sec:reward_funcs}).
\end{itemize}

\paragraph{Methods.}
We now give an overview of existing video-to-text methods and models. We note that research here is somewhat preliminary: low-quality video captions and the difficulties of the video modality means progress lags behind foundation models in other domains.

\begin{itemize}

\item \textit{Compositional approaches} have often been used to obtain video-to-text pipelines \cite{chen2023video,shang2024traveler,zeng2022socratic,li2022composing}. For example, \textcite{chen2023video} use an image-language model to answer questions about individual video frames, and an LLM to synthesize this information to produce a global summary. However, the lack of end-to-end video training in compositional approaches mean they can lack rich, nuanced video representations.

\item \textit{Leveraging pretrained LLMs via adaptors and finetuning.}
Approaches here typically first obtain a pretrained LLM and an (often) pretrained video encoder, and then define an adaptor module to channel information from the video encoder output into the LLM \cite{lin2023video,maaz2023video,zhang2023video,Papalampidi2023ASR}.
Once the new video-to-text architecture is defined, a common training scheme is:
(i) convert large, diverse video-text data into token sequences and perform next-token-prediction pretraining, then
(ii) perform supervised instruction-tuning on small, high-quality instruction datasets \cite{lin2023video,zhang2023video}.
Following trends in LLMs, future work could investigate a third RLHF finetuning stage \cite{kaufmann2023survey}.
During training, the LLM and video encoder may be finetuned \cite{lin2023video}, or kept frozen \cite{maaz2023video}.

\item \textit{Natively multi-modal models.}
Previously discussed methods have involved combining and finetuning models not initially intended for video-to-text purposes.
Recent multi-modal any-to-any sequence models---where video-to-text is formulated as an interleaved video and text sequence modelling problem---represent a step towards more \emph{natively} multi-modal models. \textcite{Liu2024WorldMO,team2023gemini,jin2023unified} all train multi-modal any-to-any (or any-to-text) autoregressive transformers via next token prediction.

\end{itemize}


\subsubsection{Challenges}
\label{sec:video_fm_challenges}

Scaling datasets and model sizes has recently led to impressive advances in video ML \cite{Zhao2024VideoPrismAF,Yang2023LearningIR,wang2024internvideo2}.
Most notably, the recent Sora model \cite{Yang2023LearningIR} demonstrated significant improvements in video prediction visual quality, physical realism, and generation lengths.
However, there are still major limitations in the capabilities of current video foundation models.
Models can hallucinate, and can struggle with spatial relationships, fine-grained spatio-temporal details, and long-term understandings and generation \cite{lin2023video,maaz2023video,Yang2023LearningIR}.

One bottleneck to further progress here is the quality (and quantity) of available video-data (see Section \ref{sec:datasets}).
In particular, improved annotations with precise low-level details will enhance robotics-relevant fine-grained understandings. 
The computational demands of processing high-dimensional and long video data present is another challenge.
Improved model architectures can help here \cite{yang2024video}.
For example, progress in long video understanding will be aided by efficient architectures for handling longer contexts \cite{liu2023ring,gu2023mamba,balavzevic2024memory}.
Other directions to improve video foundation models include using RL finetuning to address hallucination issues \cite{black2023training},
and leveraging 3D information to improve the physical realism of video generations \cite{zhen20243d}.
Improved methods for fine-grained, single-step conditioning of video predictions will be relevant to robotics \cite{Bruce2024GenieGI}.
Finally, we advocate for progress in open-sourced foundational video prediction models: open-sourced models make LfV research more accessible to the wider community.

\subsection{Video-to-Robot Transfer: Addressing Missing Action Labels and Distribution Shift}
\label{sec:mitigating}

The fundamental challenges associated with LfV are outlined in Section \ref{sec:challenges}. In this section, we detail two categories of techniques that each address a fundamental challenge: (1) The use of action representations to mitigate missing action labels in video data (Section \ref{sec:alt_actions}). (2) The use of representations designed to explicitly address LfV distribution-shift issues (Section \ref{sec:reps_address_shifts}). These techniques are often used as a single component within a larger LfV pipeline. We thus describe these technique in isolation here, in advance of our main analysis of the LfV literature in Section \ref{sec:applications}.

\subsubsection{Action Representations}

\label{sec:alt_actions}

\begin{figure}
    \centering
    \begin{subfigure}[b]{\textwidth}
        \centering
        \includegraphics[width=0.9\textwidth]{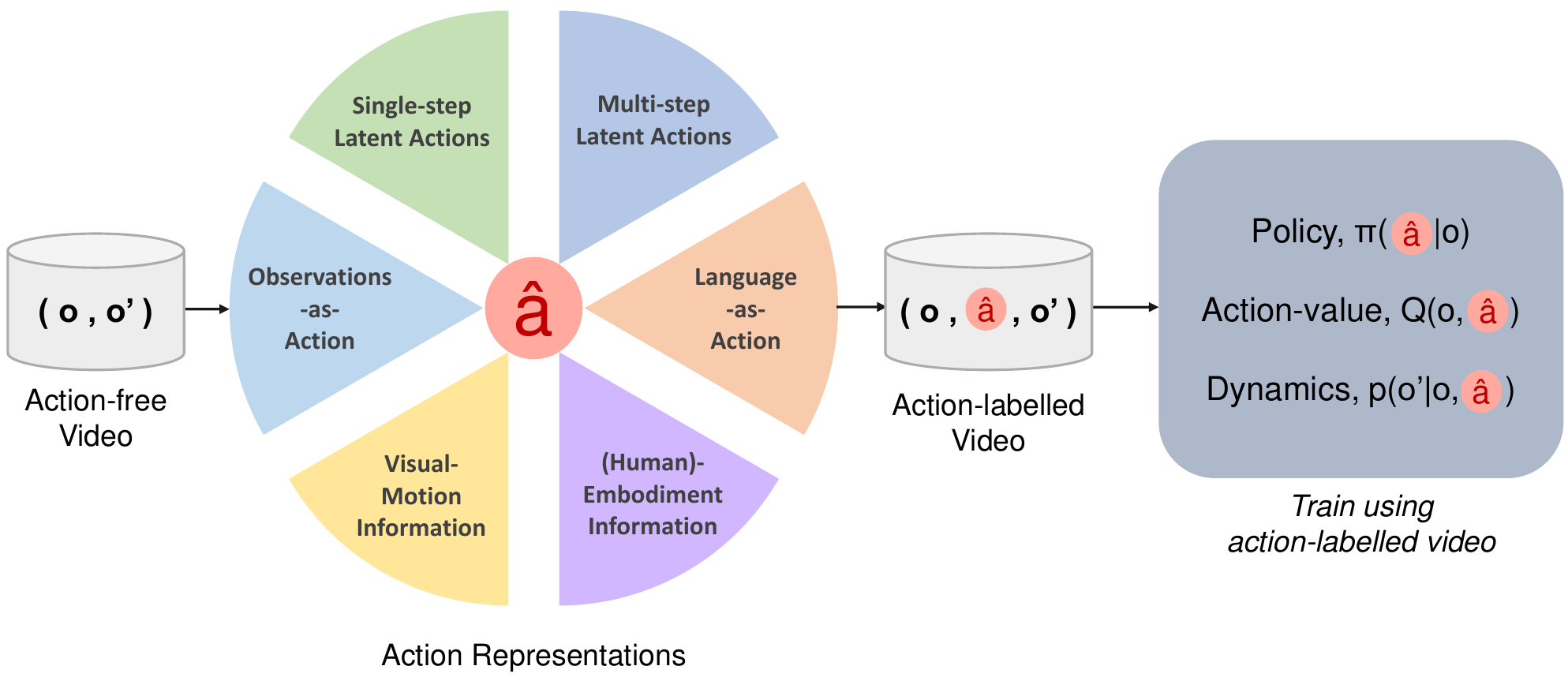}
        \caption{Action-free videos can be labelled with alternative-action representations $\hat a$. This labelled data can be used to train an alternative-action RL knowledge modality (e.g., a policy, value function, or dynamics model).}
        \label{fig:alt_acts_sub1}
    \end{subfigure}
    
    \vspace{0.5cm} 
    
    \begin{subfigure}[b]{\textwidth}
        \centering
        \includegraphics[width=0.8\textwidth]{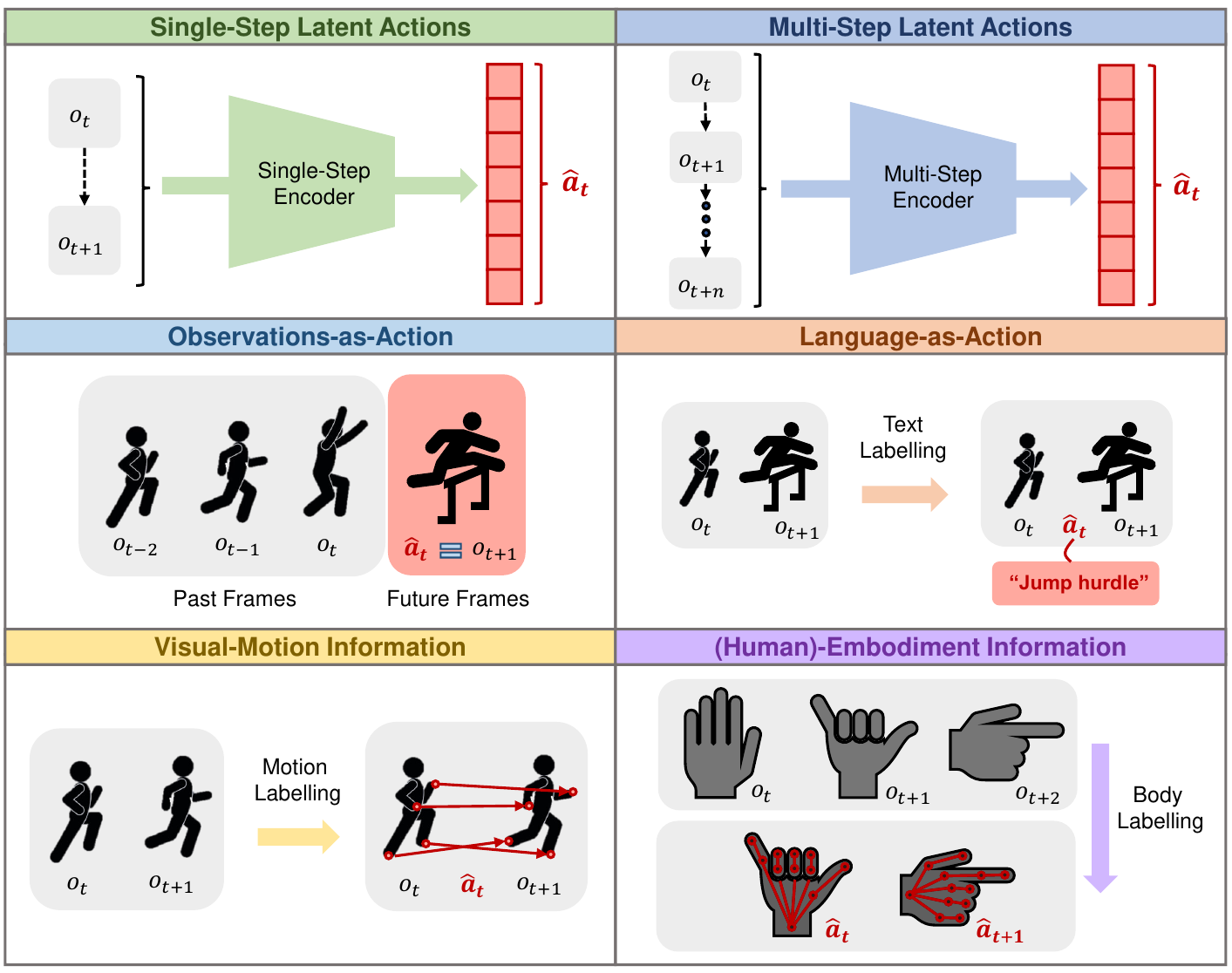}
        \caption{Categories of action representations that can be learned or obtained from video data.}
        \label{fig:alt_acts_sub2}
    \end{subfigure}
    
    \caption{\textbf{Recovering action representations from video} to overcome the missing action label problem in LfV (Section \ref{sec:alt_actions}).}
    \label{fig:alt_acts}
\end{figure}

Video data transition tuples are missing action labels: they come in the form of $(o_t, o_{t+1})$. In this section, we review works that define or learn some representation space that is analogous to the notion of an action; an \textit{alternative action representation}.
We use $\hat a \in \mathcal{\hat A}$ to denote a single alternative-action and its underlying alternative-action space.
With $\mathcal{\hat A}$, video data can be relabelled from $(o_t, o_{t+1})$ to $(o_t, \hat a_t, o_{t+1})$ and used to train an `alternative-action' version of an RL KM (see Figure \ref{fig:alt_acts_sub1}).
Alternative-action policies $\pi_{\text{alt}}(\hat a_t|o_t)$ \cite{Schmidt2023LearningTA}, dynamics models $p_{\text{alt}}(o_{t+1}|\hat a_t,o_t)$ \cite{Bruce2024GenieGI}, and value functions $Q_{\text{alt}}(o_t, \hat a_t)$ \cite{Bhateja2023RoboticOR} have all been previously trained.
These alternative-action KMs can be useful downstream if a mapping to the robot action-space, $f: \mathcal{A} \rightarrow \mathcal{\hat A}$, is obtained \cite{Wang2023MimicPlayLI,Wen2023AnypointTM,Du2023LearningUP}. Otherwise, their representations can prove highly useful \cite{Bhateja2023RoboticOR,Schmidt2023LearningTA}.

There are a number of characteristics we desire from our action representations $\mathcal{\hat A}$:

\begin{itemize}

    \item \textit{$\mathcal{\hat A}$ should be inferable from video data.} Action representations that can be obtained from raw, unlabelled video are appealing as they can leverage the full range of available video data. However, $\mathcal{\hat A}$'s that require labelled video (e.g., language captions) are also valid. 

    \item \textit{$\mathcal{\hat A}$ should contain `action' information.} We can operationalize this in two ways. First, $\mathcal{\hat A}$ should be predictive of future video frames: learning $p_{\text{alt}}(o_{t+1}|\hat a_t,o_t)$ should be easier than $p(o_{t+1}|o_t)$. Second, $\mathcal{\hat A}$ should be predictive of robot actions: learning $\pi(a_t|\hat a_t, o_t)$ should be easier than $\pi(a_t|o_t)$.

    \item \textit{$\mathcal{\hat A}$ should be transferable across embodiments.} It should capture a high-level, general notion of an action, and transfer from the embodiments in the video data (i.e., humans) to the target robot embodiment(s).
    
    \item \textit{$\mathcal{\hat A}$ should be well-structured.} It may be beneficial for $\mathcal{\hat A}$ to contain minimal information, to be well disentangled, and to maintain consistent meaning across the state space. This can allow $\pi(\hat a_t|o_t)$ and $\pi(a_t|\hat a_t, o_t)$ to be learned in a more data-efficient and generalizable manner \cite{Rybkin2018LearningWY,Bruce2024GenieGI}.

\end{itemize}

We now outline the main categories of $\mathcal{\hat A}$ we have identified in the literature (see Figure \ref{fig:alt_acts_sub2}). We focus on detailing how $\mathcal{\hat A}$ can be defined and learned from video data. More details on how $\mathcal{\hat A}$ can be used downstream for robotics are found throughout Section \ref{sec:applications}.

\paragraph{Single-step Latent Actions.}
Here, we consider \emph{learned} representations where $\hat a_t$ (i.e., the latent action) only contains information regarding the action that took us from $o_t$ to $o_{t+1}$. Such a latent representation is commonly learned using a next-observation prediction objective $p(o_{t+1}|\hat a_t, o_t)$, such that $\hat a_t$ is informative for the predictions of a forward dynamics model (FDM) \cite{Edwards2018ImitatingLP,Rybkin2018LearningWY,Schmidt2023LearningTA,Bruce2024GenieGI,Ye2024LatentAP}. A trainable (latent) inverse dynamics model (IDM) can be used to encode past-frames and infer $\hat a_t$ \cite{Rybkin2018LearningWY,Bruce2024GenieGI}.
Here, some form of regularisation, such as a vector-quantization bottleneck \cite{Schmidt2023LearningTA,Bruce2024GenieGI}, can be employed to prevent the IDM from copying the entire observation $o_{t+1}$ into the latent action.

We note some potential limitations of these approaches. First, these latent actions model all visual changes that occur in a video. Ideally, $\hat a_t$ would only represent changes due to actions taken by a single agent. Second, as these approaches model environment transitions on a visual level, thus they may omit important non-visual low-level information (such as forces). We finally note that more research into learning single-step latent action-spaces from realistic internet video is required.

\paragraph{Multi-step Latent Actions.}

This refers to \emph{learned} action representations, $\hat a$, that contain information regarding the actions taken across \emph{multiple} time-steps of a video (i.e., from $o_t$ to $o_{t+k}$). Any means of representing video segments could be applicable here, including the video representation methods seen in Section \ref{sec:video_FMs_ssl}. Here, we only detail methods used directly in the LfV literature. Building upon the `latent plans' literature \cite{lynch2020learning,Cui2022FromPT,RoseteBeas2022LatentPF}, \textcite{Wang2023MimicPlayLI} learn latent plan representations from action-free videos via variational auto-encoding. However, they use 3D human hand trajectories as the decoder target (rather than raw video). \textcite{fan2022minedojo,Lifshitz2023STEVE1AG,Sontakke2023RoboCLIPOD} use video-language contrastive losses on language-captioned video. \textcite{ChaneSane2023LearningVP,Chen2021LearningGR} perform supervised contrastive learning on video labelled with the action being performed. \textcite{Xu2023XSkillCE} use a self-supervised learning clustering approach to learn video representations. Further approaches for learning multi-step latent actions are seen in the works of \textcite{tomar2023video,Pertsch2022CrossDomainTV,Cai2023GROOTLT}.


\paragraph{Observations-as-action.}
\label{sec:alt_act_obs_as_act}

Future observations (or an encoding thereof) provide information regarding what actions will be taken next in a video.
Thus, they can be used as action representations.
Observations-as-actions can be implemented on varying time-horizons, as we now describe.
(1) \emph{Next-observation-as-action:}
These methods use the next observation $o_{t+1}$ as the $\hat a_t$ label \cite{Du2023LearningUP,Thomas2023PLEXMT,he2024large}.
$o_{t+1}$ provides clear information regarding the action that should be taken at $o_t$.
(2) \emph{Observations-as-subgoals:}
A subgoal can be thought of as a high-level action.
In `observations-as-subgoals' methods \cite{Black2023ZeroShotRM,park2024hiql,Bhateja2023RoboticOR}, an observation (or embedding thereof) from $k$ time-steps into the future is used as a sub-goal / action representation.
Some simple strategies for defining sub-goals in the video data include: choosing a fixed time-horizon $k$ \cite{Black2023ZeroShotRM,Du2023LearningUP}, or randomly sampling observations beyond the current timestep \cite{Bhateja2023RoboticOR}.
More complex strategies include using key-frame identification to identify bottleneck states in video \cite{Pertsch2019KeyframingTF,liu2023learning}.

\paragraph{Language-as-action.}
Natural language can be used as a flexible, high-level action-space (e.g., $\hat a_t =$ ``pick up the cube") \cite{belkhale2024rt,shi2024yell,xiang2024pandora},
and can allow for interfacing with other language models \cite{Du2023VideoLP}.
Some video datasets come with language-action annotations \cite{Grauman2021Ego4DAT}.
Otherwise, standard annotations can be further processed (e.g., using LLMs) \cite{Mu2023EmbodiedGPTVP}.
If the video does not come with any language annotations, manual or automated captioning methods can be used (see Section \ref{sec:dataset_curating}).
One downside here is that language is coarse and may omit important lower-level action information.

\paragraph{Visual Motion Information.}

Other works have used visual motion information in video to define $\mathcal{\hat A}$. \textcite{Wen2023AnypointTM} use 2D point trajectories, obtained by tracking points on objects throughout the video via an off-the shelf point tracker \cite{karaev2023cotracker}. \textcite{Yuan2024GeneralFA} similarly use 3D point trajectories obtained from 3D annotated datasets. Elsewhere, \textcite{Ko2023LearningTA} use an off-the-shelf model \cite{xu2022gmflow} to predict optical flow, giving a pixel-level dense correspondence map between two frames. \textcite{Nasiriany2024PIVOTIV} represent actions via visual arrows in images. Finally, \textcite{Wang2023ManipulateBS} use structure-from-motion \cite{schonberger2016structure} to recover action information, \textcite{Yuan2021DMotionRV} use the motions of object-centric representations, and \textcite{Yang2023LearningIR} optionally use camera frame motion information as conditioning information for a video predictor.

\paragraph{(Human)-embodiment Information.}
\label{sec:alt_action_human}
Off-the-shelf human-hand detection models \cite{rong2020frankmocap,Shan2020UnderstandingHH} can extract hand poses or affordances from videos which can act as an action representation \cite{Bharadhwaj2023TowardsGZ,Bahl2023AffordancesFH,Shaw2022VideoDexLD,Qin2022FromOH,Qin2021DexMVIL}. For example, $\hat a_t$ can be defined as the pose that should be reached at time $t+1$. \textcite{Bharadhwaj2023TowardsGZ} use object masks in addition to human poses to define $\mathcal{\hat A}$. Animal embodiment information could also be used \cite{peng2020learning}. More details on methods for detecting human hand poses and affordances from video are provided in Section \ref{sec:reps_address_shifts}.

\paragraph{IDM Pseudo-action Labels.}
\label{sec:alt_actions_idm}
Though not technically an \emph{alternative} action representation, these methods train an inverse dynamics model $p^{-1}(a_{t}|o_{t},o_{t+1})$ on action-labelled robot data, and use it to provide pseudo-action labels for action-free video data \cite{Baker2022VideoP,Torabi2018BehavioralCF,Schmeckpeper2020ReinforcementLW}.
However, these approaches are unlikely to scale to diverse internet video as they require either:
(i) minimal domain-shift between the video data and the robot domain (e.g., \textcite{Baker2022VideoP} assume identical embodiments),
or (ii) an explicit mechanism to deal with domain-shift that may not scale well to diverse internet video \cite{Schmeckpeper2020ReinforcementLW,Kim2023GivingRA} (see Section \ref{sec:reps_address_shifts}).

\paragraph{Discussion.}
\label{sec:alt_actions_discuss}
We now discuss key considerations regarding the feasibility and utility of alternative action representations for LfV, and possible directions for future work:

\begin{itemize}

\item \textit{Can $\mathcal{\hat A}$ be easily obtained from internet video?}
Observations-as-action can be obtained from any raw video and can (in theory) leverage the full range of available video data.
Single-step latent actions, and (some) multi-step latent actions can also be obtained from raw video, but an additional learning step (that may face optimisation difficulties) is required.
Language-as-action, visual motion information, and (human)-embodiment information all require labelled video.
Nevertheless, often off-the-shelf models can provide labels, and curated video datasets often come paired with language annotations (see Section \ref{sec:dataset_existing}).
Visual motion information and (human)-embodiment information approaches may require higher degrees of structure in the video (e.g., a relatively fixed frame); it is unclear how sensible these representations will be when applied to heterogenous internet video data.

\item \textit{Once obtained, how useful is $\mathcal{\hat A}$?}
The information content of a given $\mathcal{\hat A}$ will determine how useful it will be.
A longer-horizon or more high-level $\mathcal{\hat A}$ (e.g., language-as-actions) may ultimately aid downstream performances in long-horizon tasks.
However, shorter-horizon $\mathcal{\hat A}$'s with more low-level information may be more informative for decoding $a_t$ from $\hat a_t$.
Note, video-based actions representations will not usually contain all information required to accurately decode the exact low-level robot action.
As such, in practice, any decoder may need to be defined as $\pi(a_t|\hat a_t,o_t)$ rather than $\pi(a_t|\hat a_t)$.

\item \textit{Future directions.}
There is no work comprehensively comparing the utility of different action representations for LfV. We advocate for empirical research to help answer the questions we have posed above in the two paragraphs above.

\end{itemize}

\subsubsection{Representations to Address LfV Distribution-Shift}

\label{sec:reps_address_shifts}

Distribution shift between internet video and the target robot domain poses a challenge to LfV approaches (see Section \ref{sec:challenges}).
Such distribution shifts can hinder our ability to transfer knowledge from the video data to the robot.
A line of LfV research has attempted to overcome these issues by explicitly designing representations of (human) video that are more transferable to the robot domain.
Once the transferable representation is obtained, works have most commonly used it to:
(i) define a reward function encouraging the robot to match the behaviour in the video \cite{Zakka2021XIRLCI},
or (ii) to help train a policy via behaviour cloning of the represented video \cite{Bharadhwaj2023TowardsGZ}.

In this section, we focus solely on describing methods for obtaining representations that explicitly address LfV distribution-shift. Details regarding how the obtained representation can be used downstream to aid robot learning are found throughout Section \ref{sec:applications}.

\paragraph{Method: (Human)-embodiment-aware Approaches.}
In LfV, we are most often interested in learning from human behaviour; in particular in replicating the effects of the human hand. There is a line of LfV research that explicitly detects human-embodiment information in video, and subsequently transfers this information to the robot. Note, animal embodiment information can also be used \cite{peng2020learning}.

\begin{itemize}
\item \emph{What types of human-embodiment information can be detected?}
    (1) Poses:
    Several works explicitly estimate the pose of the human body or hand in videos \cite{Shaw2022VideoDexLD,Sivakumar2022RoboticTL,li2024okami}.
    This involves estimating the positions and orientations of various joints or key points on the body via off-the-shelf models.
    Once a human-hand pose is detected, it may need to be retargeted to the robot embodiment.
    This can be achieved by directly optimising a loss function at inference \cite{Shaw2022VideoDexLD}, or by training a retargeting network to minimise a retargeting loss function \cite{Sivakumar2022RoboticTL}.
    (2) Affordances:
    A common affordance seen in the LfV literature is the combined use of grasp points and grasp trajectories \cite{Bahl2023AffordancesFH,Mendonca2023StructuredWM} -- this abstraction transfers naturally from human to robot embodiment.
    This affordance information can be extracted from videos using an off-the-shelf model \cite{Shan2020UnderstandingHH}, and using further tricks to ensure accurate, usable affordance labels \cite{Bahl2023AffordancesFH}.
    Note, 2D affordances can be converted to 3D using depth estimation \cite{Mendonca2023StructuredWM}.    
    (3) Masks and Bounding boxes: Masks or bounding boxes of human hands and objects can be used \cite{Bharadhwaj2023TowardsGZ}. These can be obtained from videos using off-the shelf models \cite{kirillov2023segment,zhang2022fine,li2024okami} or via labelled datasets \cite{darkhalil2022epic}.

\item \emph{How to use this human embodiment-centric information?}
A video dataset labelled with the human-centric information described above can allow for one of the following to be performed:
(1) The original process for extracting the human-centric information from the video/image can be distilled into a single model \cite{Mendonca2023StructuredWM}.
This can be useful when the original method is convoluted or unlikely to generalize.
This process may also be a good auxiliary representation learning objective for robot manipulation \cite{Bahl2023AffordancesFH}.
(2) The extracted embodiment information can be treated as an `alternative action representation' \cite{Bharadhwaj2023TowardsGZ,Bahl2023AffordancesFH,Shaw2022VideoDexLD,Qin2022FromOH,Qin2021DexMVIL} (see Section \ref{sec:alt_action_human}), or as a state-observation when retargeted to the robot embodiment \cite{radosavovic2024humanoid}.

\item \emph{Limitations.} There can still be difficulties transferring this human-embodiment information to the robot. For example, a model trained to propose future poses solely on human images is unlikely to generalise zero-shot to robot images. This has led to tricks being used in the LfV literature, such as predicting affordances only when the embodiment is out of frame \cite{Bahl2023AffordancesFH}, or in-painting human embodiments over robot embodiments \cite{Bharadhwaj2023TowardsGZ}.

\end{itemize}

\paragraph{Method: Learned Invariant Representations.}
These are methods that \emph{learn} a representation that is invariant to a specific distribution shift between the video dataset and robot domain.

\begin{itemize}

\item \textit{Approaches.}
(1) Domain confusion techniques can learn representations that are invariant across viewpoints \cite{stadie2017third} and embodiments \cite{Schmeckpeper2020ReinforcementLW}.
(2) Contrastive learning and related techniques can encourage invariance across a particular axis; such as viewpoints \cite{sermanet2018time} and visual changes \cite{Aytar2018PlayingHE}.
(3) Temporal cycle-consistency objectives can learn embodiment invariant representations \textcite{Zakka2021XIRLCI}.
(4) Factorized representations learn one representation that is common across the video and robot distributions, and one that is unique to each distribution \cite{Schmeckpeper2019LearningPM,Shang2021SelfSupervisedDR}. Similarly, \textcite{Chang2023LookMN} use embodiment segmentation and in-painting removal to obtain explicit factorized representations of the agent and environment. (5) Image translation methods can convert an image from the source distribution (e.g., human embodiment) to the target (e.g., robot embodiment) \cite{Smith2019AVIDLM,Bahl2022HumantoRobotII,Bharadhwaj2023TowardsGZ,li2024ag2manip}.

\item \textit{Limitations.}
(1) Invariant representations can omit useful information, such as important differences between human and robot embodiments. (2) These methods can have overly strict requirements on the video data. For example, \textcite{sermanet2018time} assume access to multiple videos of the same scene but from different viewpoints. (3) Many of these methods only provide invariance to a single type of distribution shift (e.g., embodiment differences only). In reality, there may be many distinct shifts between any given video and the robot's setting.

\end{itemize}

\paragraph{Method: Transferable Abstractions.}
Some methods exploit abstractions that naturally transfer well from human to robot.
This includes:
object-centric graphical representations \cite{sieb2020graph,Kumar2022GraphIR};
object-centric activity-context priors \cite{Nagarajan2021ShapingEA};
key-points in video \cite{Karnan2021VOILAVI,Xiong2021LearningBW};
and embodiment-agnostic affordances \cite{Bahl2023AffordancesFH,Mendonca2023StructuredWM}.
These methods all respectively benefit from the use of off-the-shelf object, key-point, and human-hand detectors.
Language has also been used as a transferable abstraction \cite{Chen2021LearningGR,Mu2023EmbodiedGPTVP,Pertsch2022CrossDomainTV}.


\paragraph{Discussion.}

These methods highlight the trade-off that must be made between:
(i) imposing structure and inductive biases to improve performance in narrow settings, and
(ii) opting for end-to-end learning methods that are more scalable, but are less effective in the small data regime and narrow settings.
The central focus of this survey is on scaling to diverse internet data in order to tackle unstructured environments with generalist robots.
As such, we advocate for methods following the spirit of (ii).
Through this lens, we regard some of the methods in this section to be less promising, and have commented on their limitations above.


\subsection{Robot Learning and Reinforcement Learning from Video}
\label{sec:applications}

\begin{figure}[t]
    \includegraphics[width=\textwidth]{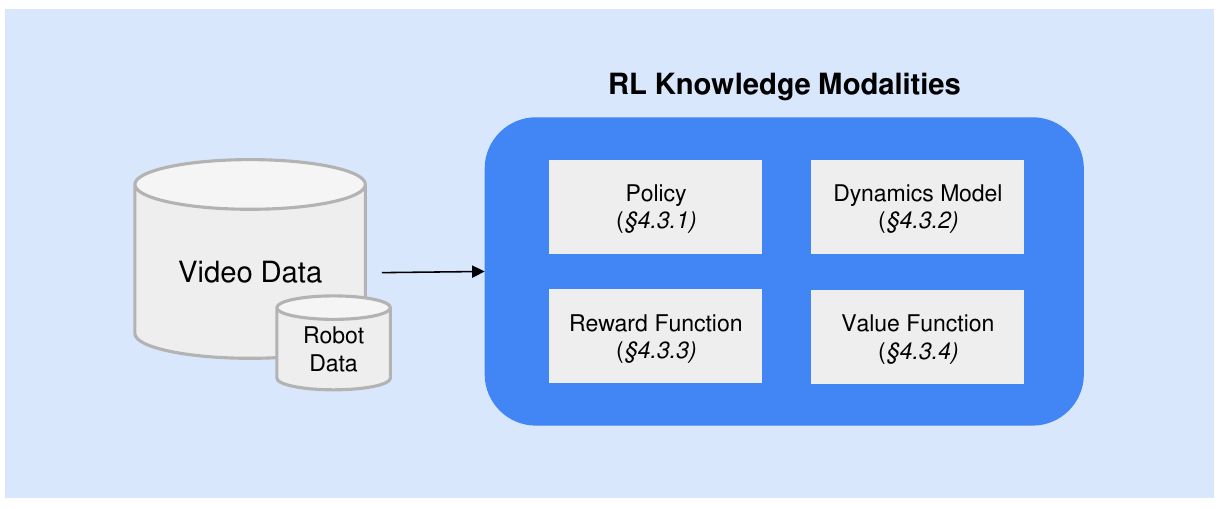}
    \centering
    \caption{\textbf{Leveraging video to benefit RL and robot learning} (Section \ref{sec:applications}). We categorise the LfV literature according to which Reinforcement Learning (RL) Knowledge Modality (KM) benefits from the use of video data, resulting in the taxonomy illustrated in the above figure.}
\label{fig:rl_kms}
\end{figure}

This section presents our main analysis of the LfV literature.
We taxonomize this literature according to which downstream RL knowledge modality (KM) most directly benefits from the use of video data by the LfV method (see Figure \ref{fig:rl_kms}).
The KMs we consider here are: policies (Section \ref{sec:policy}), dynamics models (Section \ref{sec:dynamics}), reward functions (Section \ref{sec:reward_funcs}), and value functions (Section \ref{sec:value_funcs}).
For brief descriptions of each of these KMs, we refer the reader back to Section \ref{sec:prelims_rl}.

\subsubsection{Policies}
\label{sec:policy}

\begin{figure}[b]
    \includegraphics[width=0.9\textwidth]{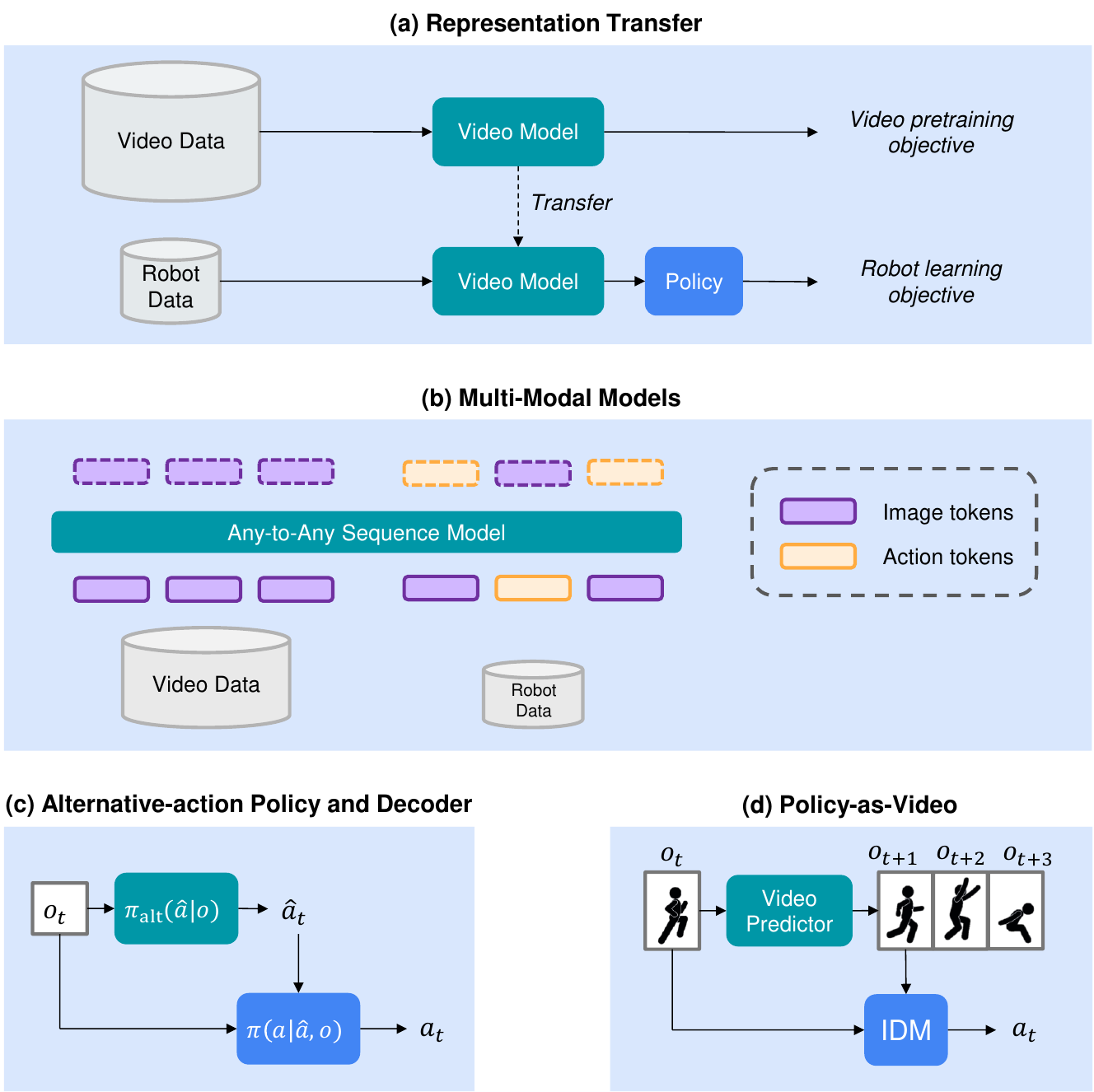}
    \centering
    \caption{\textbf{Learning policies from video} (Section \ref{sec:policy}). (a) Representations can be pretrained from video data and used downstream whilst training a policy on robot data. (b) Models can be trained jointly on video and robot data to predict robot actions. (c) A $\pi_{alt}$ can be trained on video data to output an action representation (see Section \ref{sec:alt_actions}) which conditions an action-decoding $\pi$ trained on robot data. (d) Policy-as-video is an specific instantiation of this hierarchical setup where the action representation is a video and the decoder is (often) an inverse dynamics model (IDM).}
    \label{fig:policy}
\end{figure}

Ultimately, the goal of LfV is to use $D_\text{video}$ to help obtain a policy $\pi(a_t|o_t)$. In this section, we detail literature where $\pi$ is the RL KM that most directly benefits from the use of video data. We first identify and detail several distinct categories of methods (see Figure \ref{fig:policy}), before moving to a brief discussion.

\paragraph{Method: Representation Transfer.}
These methods pretrain visual representations on video data using some learning objective.
Subsequently, the representations are transferred (either frozen or to be finetuned) to aid downstream learning of a policy.
Figure \ref{fig:policy} visualizes this process.
There have been several large-scale analyses of representation transfer from video data \cite{Jing2023ExploringVP,Silwal2023WhatDW,Zhao2022WhatMR,Burns2023WhatMP,Hu2023ForPV,Majumdar2023WhereAW}.

A number of learning objectives have been explored in the literature.
This includes:
frame-level objectives such as masked-autoencoding (MAE) \cite{Radosavovic2022RealWorldRL}, contrastive learning \cite{Chen2021AnES,Jing2023ExploringVP}, and others \cite{Zhao2022WhatMR}; time-contrastive learning objectives \cite{Ma2022VIPTU,Ma2023LIVLR,Nair2022R3MAU};
video prediction objectives \cite{Wu2023UnleashingLV,he2024large,jang2024visual};
temporal-difference learning \cite{Bhateja2023RoboticOR};
predicting affordances from videos \cite{Bahl2023AffordancesFH};
predicting latent actions \cite{Schmidt2023LearningTA};
and objectives that leverage language labels, such as image-language contrastive learning \cite{Ma2023LIVLR}, video-language alignment \cite{Nair2022R3MAU} and video captioning \cite{Karamcheti2023LanguageDrivenRL}.
Works have often combined multiple objectives \cite{Nair2022R3MAU,Ma2023LIVLR,Karamcheti2023LanguageDrivenRL}.


The use of video data relevant to the robot is important \cite{Jing2023ExploringVP}.
\textcite{Nair2022R3MAU,Ma2023LIVLR,Karamcheti2023LanguageDrivenRL,Majumdar2023WhereAW} leverage ego-centric human and/or navigation video.
\textcite{Radosavovic2022RealWorldRL} leverage third-person video of humans interacting with objects \cite{Shan2020UnderstandingHH,Miech2019HowTo100MLA}.
Others have leveraged videos of robots \cite{Dasari2023AnUL}, or first-person videos of humans operating robot-like grippers \cite{shafiullah2023bringing}.
Combining datasets to improve data diversity has proven beneficial \cite{Majumdar2023WhereAW,Dasari2023AnUL,Radosavovic2022RealWorldRL}.


\paragraph{Method: Multi-modal Models.}
These methods train a monolithic model jointly on video and robot data.
\textcite{Reed2022AGA} train a sequence model on multi-modal data, including image, text, and robot data, but do not use video data.
\textcite{radosavovic2024humanoid} train a sequence model for locomotion on both action-labelled robot data and action-free pose trajectories extracted from YouTube videos. \textcite{sohn2024introducing} present an 8 billion parameter any-to-any transformer trained on text, images, videos, and robot perceptions and actions (but experimental details are not published).
These recent methods may obtain positive transfer by learning from different modalities of data, and are promising due to their simplicity and scalability.

\paragraph{Method: Alternative-action Policies $\pi_{\text{alt}}(\hat a_t| o_t)$.}
In Section \ref{sec:alt_actions}, we outlined different alternative-action representations $\mathcal{\hat A}$.
Here, we detail methods that use such representations to train a policy $\pi_{\text{alt}}(\hat a_t| o_t)$ from video data.
Such a pretrained $\pi_{\text{alt}}(\hat a_t| o_t)$ can be useful for representation transfer, or to condition a decoder $\pi(a_t|\hat a_t,o_t)$ (see the following paragraph \cite{Schmidt2023LearningTA}).

Works have trained a $\pi_{\text{alt}}(\hat a_t| o_t)$ using:
single-step latent actions \cite{Edwards2018ImitatingLP,Schmidt2023LearningTA,Bruce2024GenieGI};
multi-step latent actions \cite{Wang2023MimicPlayLI,Pertsch2022CrossDomainTV};
language-actions \cite{Ajay2023CompositionalFM,Yang2023LearningIR,Du2023VideoLP,Mu2023EmbodiedGPTVP};
observations-as-action, including next-observation-as-action policies \cite{Du2023LearningUP,Thomas2023PLEXMT,he2024large} and sub-goal policies \cite{Black2023ZeroShotRM,park2024hiql};
visual motion information \cite{Wen2023AnypointTM,Yuan2024GeneralFA};
and human-embodiment information \cite{Bahl2023AffordancesFH,Shaw2022VideoDexLD,Qin2022FromOH,Qin2021DexMVIL,Peng2018SFVRL,Bharadhwaj2023TowardsGZ}.

A common training process involves assuming the video data contains relatively expert behaviour, labelling the video data with the alternative-action, and training $\pi_{\text{alt}}(\hat a| o)$ using a supervised behaviour cloning objective \cite{Schmidt2023LearningTA,Bruce2024GenieGI,Wang2023MimicPlayLI,Du2023VideoLP,Black2023ZeroShotRM,Wen2023AnypointTM,Bharadhwaj2023TowardsGZ}.
When the behaviour in the data is suboptimal, language \cite{Du2023VideoLP,Wen2023AnypointTM} or goal-conditioning \cite{Wang2023MimicPlayLI} can be beneficial.
Offline RL is another option when the data is suboptimal (though this is currently underexplored).

To obtain language-as-action policies, \textcite{Yang2023LearningIR,Du2023VideoLP,Mu2023EmbodiedGPTVP} supervised finetune internet-pretrained VLMs or LLMs.
\textcite{Black2023ZeroShotRM} supervised finetune an image-editing diffusion model to edit the current observation into a subgoal image.

\paragraph{Method: Alternative-action Decoders $\pi(a_t|\hat a_t,o_t)$.}
\label{sec:policy_alt_decode}

Several works train a low-level robot policy $\pi(a_t|\hat a_t,o_t)$ that decodes $\hat a_t$ to $a_t$. This decoder can be useful when $\hat a_t$ is: (1) obtained from a $\pi_{\text{alt}}(\hat a_t| o_t)$ -- i.e., \emph{hierarchical conditioning via $\pi_{\text{alt}}(\hat a_t| o_t)$} \cite{Schmidt2023LearningTA,Du2023LearningUP,Wen2023AnypointTM}; or (2) $\hat a_t$ is a video instruction, usually a representation of a human demonstration video -- i.e., \emph{video-as-instructions} \cite{ChaneSane2023LearningVP,Cai2023GROOTLT,Lifshitz2023STEVE1AG,Xu2023XSkillCE}.
The decoding policy $\pi(a_t|\hat a_t,o_t)$ can either be learned with data, or can be manually crafted.

\begin{itemize}

\item \emph{Learning the decoder}:
The $D_{\text{robot}}$ can be labelled with alternative actions to give tuples of the form $(o_t, a_t, \hat a_t, o_{t+1})$, and the decoder can be trained via supervised learning on these tuples.
This has been done with single-step latent actions \cite{Schmidt2023LearningTA,Bruce2024GenieGI}, multi-step latent actions \cite{Wang2023MimicPlayLI}, point-trajectory latent actions \cite{Wen2023AnypointTM}, next-observation-as-actions \cite{Du2023LearningUP} and sub-goals-as-actions \cite{Black2023ZeroShotRM}.
$\pi(a_t|\hat a_t,o_t)$ can also be trained via online RL with the compositional policy, $\pi(a_t|\pi_{\text{alt}}(\hat a_t|o_t),o_t)$, being trained end-to-end (as opposed to keeping $\pi_{\text{alt}}$ frozen) \cite{Schmidt2023LearningTA}.
\textcite{jain2024vid2robot} end-to-end learn a mapping from human demonstrations to robot actions, but require a dataset of human video demonstrations paired with equivalent robot trajectories.

\item \emph{Manually defining the decoder}: \textcite{Ko2023LearningTA} infer low-level robot actions from optical flow, \textcite{Yuan2024GeneralFA} infer actions from 3D flow predictions, while \textcite{Nasiriany2024PIVOTIV} map from robot action to visual arrows and back.
Other works have retargeted human-hand pose actions to robot poses \cite{Qin2021DexMVIL,Sivakumar2022RoboticTL}.

\end{itemize}

\paragraph{Method: Policy-as-video.}
These methods are a distinct sub-section of the alternative-action methods mentioned above.
They use observations-as-actions in a `hierarchical conditioning via $\pi_{\text{alt}}(\hat a_t| o_t)$' scheme.
A common approach here involves using a language-conditioned video predictor to generate plausible video trajectories that complete a language-specified task, before decoding actions from the generated video via an action-decoding inverse-dynamics model (IDM) $p^{-1}(a_t|o_t,o_{t+1})$ trained on $D_{\text{robot}}$ \cite{Du2023LearningUP}.
To improve upon this scheme:
\textcite{Ajay2023CompositionalFM} enforce compositional consistency between LLM plans, video generations, and the IDM actions;
and \textcite{Du2023VideoLP} improve action decoding model via the use of goal-conditioned behaviour-cloning.

\paragraph{Method: IDM Pseudo-actions.} These methods train an IDM $p^{-1}(a_t|o_t,o_{t+1})$ on action-labelled robot data and use it to label video data with pseudo-actions. The pseudo-labelled video data can then be used to help train a policy $\pi(a|o)$  \cite{Baker2022VideoP}. However, as mentioned in Section \ref{sec:alt_actions_idm}, naive IDM pseudo-action approaches are unlikely to scale to diverse internet videos.


\paragraph{Discussion.}
Learning policies from videos has shown potential. Though it comes with some cons as well as pros, there are several promising directions for future research.

\begin{itemize}

\item \textit{Performance gains:} The methods above have contributed to improvements in task-success rates \cite{Radosavovic2022RealWorldRL,Nair2022R3MAU,Parisi2022TheUE}, robot data efficiency \cite{Nair2022R3MAU,Schmidt2023LearningTA,Ye2023FoundationRL}, and generalization beyond $D_{\text{robot}}$ \cite{Du2023LearningUP,Wang2023MimicPlayLI,Black2023ZeroShotRM,Thomas2023PLEXMT,Zhao2022WhatMR,lin2024spawnnet}.
Nevertheless, this has often been in toy (non-generalist) settings or with toy datasets \cite{Schmidt2023LearningTA,Wen2023AnypointTM,Thomas2023PLEXMT}.
Gains when using diverse human video datasets have often been more modest \cite{Du2023LearningUP,Shaw2022VideoDexLD,Black2023ZeroShotRM}.

\item \textit{Pros and cons:}
A robot policy is ultimately what we want, and initial results suggest this is a highly promising KM to target.
However, there are some disadvantages to note.
Policies must output low-level actions and so the missing low-level information in video will present a challenge.
Additionally, policies prefer expert data, but the behaviours in internet video may often be diverse and non-expert.

\item \textit{Future directions:}
Jointly pretraining multi-modal foundation models on both video and robot data is still underexplored.
Representation transfer approaches may be further improved by:
training encoders that fuse spatio-temporal information (rather than embedding individual images separately);
using representations from video foundation models (see Section \ref{sec:video_FMs});
or training on larger internet datasets (see Section \ref{sec:dataset_existing}).
Methods that leverage representations $\mathcal{\hat A}$ will be limited by the scalability of their $\mathcal{\hat A}$ (as discussed in Section \ref{sec:alt_actions}).
An open question is whether $\mathcal{\hat A}$'s are best utilised explicitly in a hierarchical setup, or if they are most useful for representation transfer.
In hierarchical setups, where the policy is defined as $\pi(a_t|\pi_{\text{alt}}(\hat a_t|o_t),o_t)$, we should explore the extent to which either $\pi_{\text{alt}}$ or $\pi$ is the performance bottleneck.

\end{itemize}

\subsubsection{Dynamics Models}

\label{sec:dynamics}

Video prediction models, $p(o_{t+1}|o_t)$, capture temporal dynamics information -- information about the dynamics and physics of the world.
A line of LfV research has sought to use video prediction objectives on a $D_{\text{video}}$ to aid the learning of a robot dynamics model $p(o_{t+1}|o_t, a_t)$.
This can involve pretraining $p(o_{t+1}|o_t)$ on $D_{\text{video}}$ and adapting it into $p(o_{t+1}|o_t, a_t)$ using $D_{\text{robot}}$.
Alternatively, it could involve jointly pretraining on $D_{\text{video}}$ and $D_{\text{robot}}$.
In this section, we detail the various video prediction pretraining schemes seen in the LfV literature, before outlining how the pretrained models can be adapted and utilized for downstream robotics.

\paragraph{Pretraining: Architectures and Datasets.}
Video diffusion architectures have recently been scaled to increasingly large human video datasets for LfV \cite{Yang2023LearningIR,Du2023LearningUP,Du2023VideoLP}.
Autoregressive transformer architectures have also been employed \cite{Hu2023GAIA1AG,Wu2023UnleashingLV}, including on large-scale internet (video-game) data \cite{Bruce2024GenieGI}.
\textcite{Bruce2024GenieGI,Hu2023GAIA1AG} find their video prediction models to scale well with increased compute and model size.
Other works \cite{Seo2022ReinforcementLW,Mendonca2023StructuredWM,Wu2023PretrainingCW} use the recurrent state-space model of Dreamer \cite{Hafner2023MasteringDD}, a state-of-the-art model-based RL architecture.

\paragraph{Pretraining: Action-conditioning.}
To add action-conditioning information to the video predictor, \textcite{Yang2023LearningIR,sohn2024introducing} pretrain jointly on both video and robot data (the robot data contains action labels used for conditioning).
Otherwise, other works have pretrained an alternative-action video predictor $p_{\text{alt}}(o_{t+1}|o_t, \hat a_t)$ \cite{Yang2023LearningIR,Bruce2024GenieGI}.
Conditioning on an $\mathcal{\hat A}$ allows for more control over video generations, and often be a mapping can be obtained from $\mathcal{\hat A}$ to the robot action-space \cite{Rybkin2018LearningWY,Schmidt2023LearningTA}. 
LfV video predictors have been trained to be conditioned on:
language-as-actions \cite{Yang2023LearningIR,Du2023VideoLP};
single-step latent actions \cite{Rybkin2018LearningWY,Bruce2024GenieGI};
goal images \cite{Du2023LearningUP};
future grasp-location and post-grasp waypoint affordance information \cite{Mendonca2023StructuredWM};
or motion information \cite{Yuan2021DMotionRV,Wang2023ManipulateBS}.
\textcite{Yang2023LearningIR} condition on several different action-spaces, including robot actions, language actions, and camera motions.

\paragraph{Downstream: Adapting the Video Predictor.}
A video prediction model can be adapted via finetuning on $D_{\text{robot}}$ \cite{Du2023LearningUP,Ajay2023CompositionalFM}.
However, few works directly finetune an action-free video predictor $p(o_{t+1}|o_t)$ into an action-conditioned dynamics model $p(o_{t+1}|o_t,a_t)$.
\textcite{Mendonca2023StructuredWM} pretrain on video with affordance-based conditioning, and add optional robot action-conditioning during finetuning.
\textcite{Seo2022ReinforcementLW} find naive finetuning on robot data to result in the erasure of pretraining knowledge, and instead `stack' an action-conditioned model on top of the pretrained model.
In related work, \textcite{Yang2023ProbabilisticAO} demonstrate that the score function of a large pretrained video diffusion model can guide the generations of a smaller task-specific video model.
Adaptation can involve obtaining a mapping from $\mathcal{\hat A}$ to the robot action space \cite{Rybkin2018LearningWY} (Section \ref{sec:policy_alt_decode} elaborates on methods that map from $\mathcal{\hat A}$ to $\mathcal{A}$).

\paragraph{Downstream: Using the dynamics model.}
Once the dynamics model has been obtained it has been used in several ways in the LfV literature.
(1) \textit{As a simulator}: $p(o_{t+1}|o_t,a_t)$ can be used to generate synthetic data to help train a policy \cite{Yang2023LearningIR,Bruce2024GenieGI}.
(2) \textit{As a differentiable simulator}: similarly, one can backpropagate through model-generated rollouts to help train a policy \cite{Seo2022ReinforcementLW,Wu2023PretrainingCW,Hafner2023MasteringDD}. 
(3) \textit{For planning}: the dynamics model can be used for standard model predictive control \cite{Rybkin2018LearningWY,Mendonca2023StructuredWM} or tree-search \cite{Du2023VideoLP}.

These use-cases often require evaluation of the model-generated trajectories.
Some works learn reward/value estimates via a downstream finetuning stage, making use of reward-labelled robot data \cite{Seo2022ReinforcementLW}.
Others use video data to help learn a reward function \cite{Mendonca2023StructuredWM,Rybkin2018LearningWY,Yang2023LearningIR,Du2023VideoLP}.
See Sections \ref{sec:reward_funcs} and \ref{sec:value_funcs} for details regarding methods for learning reward and value functions from video.


\paragraph{Discussion.}
\textit{Performance gains:}
\textcite{Bruce2024GenieGI} show their video-game foundation model to exhibit initial signs of impressive generalization.
When pretraining with large human video datasets, several works have demonstrated moderate but significant performance gains \cite{Mendonca2023StructuredWM,Wu2023PretrainingCW}.
\textcite{Yang2023LearningIR,Du2023VideoLP} demonstrate that large-scale pretrained video diffusion models can be particularly useful in long-horizon tasks.

\textit{Pros and cons:}
These results, combined with the possibility that dynamics models can learn from non-expert and off-policy internet data more easily than policies \cite{yu2020mopo}, make LfV dynamics model approaches appear promising.
However, we note that, like policies, certain approaches may suffer from missing low-level information in video.

\textit{Future directions:}
(1) An obvious route to improved performances here is to continue to scale our video dataset and model sizes \cite{Bruce2024GenieGI,Hu2023GAIA1AG}.
Experimenting with scalable alternative-action techniques \cite{Bruce2024GenieGI} will be useful for conditioning foundational video predictors.
(2) A hierarchical world/dynamics model \cite{lecun2022path}, where the higher levels are learned from video data and the lower-levels are learned from robot data, is another direction to explore. (3) An underexplored direction is to combine standard analytic simulation \cite{todorov2012mujoco,makoviychuk2021isaac} with video prediction simulation \cite{Yang2023LearningIR} to obtain the benefits of both.
(4) Following synthetic data trends in other fields \cite{wang2024helpsteer2,numina_math_datasets}, future work may seek to use foundational video predictors to generate synthetic video rollouts \cite{Yang2023LearningIR,Bruce2024GenieGI}, or augment and improve the coverage of existing data via video editing \cite{yu2023scaling}.

\subsubsection{Reward Functions}

\label{sec:reward_funcs}


The reward function is an essential component of an RL algorithm.
However, manual reward design is difficult to scale to complex and unstructured real-world generalist robot settings.
LfV research has sought to tackle this issue by extracting visual reward functions from video data.
In this section, we detail any method that uses video data to help relabel transition tuples $(o_t, a_t, o_{t+1})$ to $(o_t, a_t, r_t, o_{t+1})$, thus allowing the tuples to be used for online or offline RL.

\paragraph{Extracting reward functions from video.} We now describe the main clusters of methods for extracting and constructing reward functions from video data.

\begin{itemize}
    \item \textit{Video-language Model Rewards.}
    These methods specify the task via language and use a video-language model (VidLM) (see Section \ref{sec:video_FMs}) to provide a reward signal.
    We identify two categories of methods here.
    \textit{Video-text similarity}:
    A dual encoder model can be trained to embed videos and language into the same representation-space \cite{Xu2021VideoCLIPCP,Zhao2024VideoPrismAF}.
    From such a model, a reward can be defined as the similarity (in embedding space) between a language task-description and a video of the robot's behaviour \cite{fan2022minedojo,Ding2023CLIP4MCAR,Sontakke2023RoboCLIPOD}.
    \textcite{baumli2023vision} study dual-encoder image-text rewards in detail.
    \textit{Visual question answering (VQA)}:
    Reward information can be obtained from a video-to-text model (see Section \ref{sec:video_FMs_to-text}) by asking it whether a task has been completed in a video.
    A dense reward could be obtained by asking the model to score the robots progress, or by using the model to provide feedback within an `RL-AI-F' framework \cite{klissarov2023motif}.
    VQA rewards have been used with images as input \cite{du2023vision,yang2023robot}.
    \textcite{liu2024enhancing} employ video-based VQA rewards, but the limited capabilities of current video-to-text models has hindered progress here.

    \item \textit{Video-predictors as Reward Functions.} A video prediction model $p(o_{t+1}|o_t)$ can be converted into a reward function \cite{Zhu2023GuidingOR,Escontrela2023VideoPM,huang2023diffusion}. These methods define a reward based on the likelihood of the robot video under the video predictor. This encourages the robot to match the behaviour distribution of the video data. Language-conditioned video prediction may allow these approaches to scale to diverse, non-expert internet video \cite{Escontrela2023VideoPM}.

    \item \textit{Representational Similarity to a reference.}
    These methods define the reward as the similarity between representations of the robot's observation(s) and a reference observation (i.e., a goal image or demonstration video).
    We outline two distinct approaches here.
    \textit{Standard deep representations:}
    `Standard' learning objectives on video data can provide representations useful for measuring the similarity between the current observation and a goal image.
    \textcite{Ma2022VIPTU} use a time-contrastive objective, which gives implicit measurements of the temporal distance between two images.
    \textcite{Hu2023ForPV} find that many standard deep representations can be effective, though masked autoencoding-based representations perform poorly.
    \textit{Representations that address LfV distribution shift:}
    Distribution-shifts (such as embodiment differences) can prevent meaningful comparison between human and robot videos.
    Thus, previous work has used representations designed to explicitly ignore such distributions shifts (see Section \ref{sec:reps_address_shifts}) when defining their LfV reward.
    This has included the use of:
    embodiment-invariant representations \cite{Zakka2021XIRLCI};
    object-centric representations \cite{Kumar2022GraphIR,sieb2020graph,Chang2020SemanticVN};
    the retargeting of poses from human to robot embodiments \cite{Mandikal2022DexVIPLD,Qin2021DexMVIL};
    and viewpoint invariant representations \cite{sermanet2018time,Aytar2018PlayingHE}.

    \item \textit{Potential-based shaping with value functions.}
    A value function, $V(o_t)$, pretrained from video can provide a dense reward via `potential-based shaping': $r_t = V(o_{t+1}) - V(o_t)$. In the LfV literature, such rewards have been defined using value functions pretrained on video data via TD learning \cite{Chang2022LearningVF}, or time-contrastive learning \cite{Ma2022VIPTU}.
    
    \item \textit{Generative adversarial imitation.}
    Generative adversarial imitation learning \cite{ho2016generative} has been used to encourage the robot to match the behaviour distribution of a video dataset during online learning \cite{Torabi2018GenerativeAI}. This has often required the use of representations that explicitly ignore LfV distribution shifts \cite{stadie2017third,Qin2021DexMVIL}.
    
    \item \textit{Other methods.} A task classifier can be trained on task-labelled video data to provide downstream rewards \cite{Chen2021LearningGR,Shao2020Concept2RobotLM}. Several methods use the number of steps-to-completion in the video as a proxy label to train a video reward model \cite{Yang2023LearningIR,edwards2019perceptual}. Others encourage similarity to a video-obtained behavioural prior: such as a video-pretrained policy \cite{Ye2023FoundationRL}, or a human affordance distribution \cite{Bahl2023AffordancesFH}.
    
\end{itemize}

\paragraph{Downstream Usage: Transfer Mechanisms.}
A video-pretrained reward function can be used zero-shot in the downstream robot domain \cite{Escontrela2023VideoPM}.
Other works further finetune the reward function on in-domain robot data \cite{Sontakke2023RoboCLIPOD}.
The LfV reward could be used as the sole reward for the robot, or it could be used as an exploration or shaping bonus, in addition to a sparse task reward \cite{Ye2023FoundationRL,Chang2022LearningVF}.
\textcite{Adeniji2023LanguageRM} pretrain the policy using the LfV reward, before finetuning it on a manually defined task reward.

\paragraph{Discussion.}
\textit{Pros and cons:}
Learning accurate reward functions purely from video data may be more feasible than other RL KMs.
Policies or dynamics models may need access to non-visual information (e.g., forces) in the downstream robot domain, whereas reward functions can often operate solely using visual information and thus can be used zero-shot after video pretraining \cite{Escontrela2023VideoPM}.
However, targeting other KMs via LfV (i.e., policies and dynamics models) may better reduce demands on the robot data and better aid generalization beyond $D_{\text{robot}}$.

\textit{Future directions:}
(1) The most promising approaches are likely those that can leverage pretrained video foundation models: i.e., video-language model rewards and video-predictors as reward functions.
(2) One avenue is to combine LfV reward functions with other LfV KMs. For example, finetuning an LfV policy via online RL and LfV rewards.
(3) An underexplored direction is to use LfV rewards to augment reward labels in offline RL data.
(4) LfV reward functions may prove useful for detecting safety-related metrics when deploying robots in the real world \cite{guan2024task}.
(5) Many LfV reward functions are differentiable, making them suitable for use in certain model-based RL algorithms \cite{Hafner2023MasteringDD,JyothirS2023GradientbasedPW}.
(6) Shaped reward functions could be extracted from video-to-text models via RL from AI feedback (RL-AI-F) methods \cite{klissarov2023motif}.

\subsubsection{Value Functions}

\label{sec:value_funcs}

Value functions are an essential component of most deep RL algorithms \cite{schulman2017proximal,haarnoja2018soft}.
A small but distinct line of LfV research pretrains models that closely resemble value functions using video data. We outline methods that do so below.

\paragraph{Pretraining: TD-learning.}
The temporal-difference (TD) learning objective is commonly used to learn value functions in RL \cite{sutton2018reinforcement}. However, video data is missing important action, reward, and goal labels that are often required during TD-learning.

\begin{itemize}
    \item \textit{Missing action labels:}
    Action labels are required to obtain a state-action value function $V(o_t, a_t)$, which can be more useful than a `state' value function $V(o_t)$ \cite{sutton2018reinforcement}.
    Thus, LfV research has sought to train value functions conditioned on `alternative-actions' (see Section \ref{sec:alt_actions}): $V(o_t, \hat a_t)$.
    This has been achieved by conditioning on:
    sub-goals \cite{Bhateja2023RoboticOR,ghosh2023reinforcement};
    next observations \cite{Edwards2020EstimatingQS};
    single-step latent actions \cite{Chang2022LearningVF};
    and IDM-obtained pseudo-actions \cite{Chang2020SemanticVN}.
    
    \item \textit{Missing reward and goal labels:}
    TD-learning requires reward labels.
    Meanwhile, TD-learning of multi-task/goal value functions requires task/goal labels.
    To obtain reward labels for TD learning, \textcite{Bhateja2023RoboticOR,ghosh2023reinforcement,park2024hiql} use hindsight goal relabelling: a sparse reward is defined as $r=(o==g)$ (where $o$ is the current observation and $g$ is the goal observation).
    \textcite{Chang2020SemanticVN} leverage object labels and off-the-shelf object detectors to provide goal and reward labels in navigational video data.
    Other works assume a single task setting \cite{Edwards2020EstimatingQS}, or assume access to reward labels in the video data \cite{Edwards2020EstimatingQS,Chang2022LearningVF}. These are not scalable assumptions.
    The LfV reward functions from Section \ref{sec:reward_funcs} could be applicable here, while video-to-text models (see Section \ref{sec:video_FMs_to-text}) could provide goal labels in textual form.
\end{itemize}

\paragraph{Pretraining: Time-contrastive learning.}

Time-contrastive objectives induce a temporally smooth representation, and a value function can be defined by measuring the distance between the current observation and a goal image in the representation space \cite{Ma2022VIPTU}.
Importantly, these objectives do not require action labels for video pretraining.
Quasimetric functions \cite{wang2023optimal} and temporal cycle-consistency objectives \cite{Zakka2021XIRLCI} have been used to learn similar representations.
Note, the requirement for goal images is a limitation of these approaches.

\paragraph{Pretraining: Others.}
\textcite{edwards2019perceptual} regress a value function towards a heuristic value; the number of timesteps remaining in the video. This approach assumes expert behaviour in the video. \textcite{Du2023VideoLP} take a similar approach, finetuning a VLM to give the heuristic value estimates. \textcite{liu2023learning} use critical state identification to aid value predictions, but this assumes access to reward labels in the video data.

\paragraph{Downstream Usage.}
We now briefly outline how video-pretrained value functions have been used in the downstream robot domain in the literature.
(1) \textit{As a value function:}
A video-pretrained value function can be directly used downstream in a standard fashion if: it is an action-conditioned $V(o_t, a_t)$ \cite{Chang2020SemanticVN}; or it is an alternative-action-conditioned $V_{\text{alt}}(o_t, \hat a_t)$ and a mapping from $\hat a$ to $a$ can be obtained; or an action-conditioned dynamics model $p(o_{t+1}|o_t, a_t)$ is available, allowing an unconditioned $V(o)$ to be used for tree-search planning \cite{Du2023VideoLP,Chang2022LearningVF}.
(2) \textit{Representation transfer:}
\textcite{Bhateja2023RoboticOR} initialise their downstream value function and policy representations from a video-pretrained value function.
(3) \textit{Potential-based reward shaping:} 
A reward function can be defined as: $r = V(o_{t+1}) - V(o_t)$, and be used for downstream online RL \cite{edwards2019perceptual,Ye2023FoundationRL,Chang2022LearningVF}.
It may be desirable to do this if:
the pretrained value function is not fully reliable but can provide useful auxiliary rewards to guide exploration; or the value function is not action-conditioned, so cannot be used as a $V(o_t, a_t)$ for Q-learning \cite{sutton2018reinforcement}.
(4) \textit{TD bootstrapping:}
A $V(o_t)$ can still be used to accelerate downstream TD-learning of a state-action value function $V(o_t, a_t)$ by using its estimates for the bootstrap term in the bellman backup \cite{edwards2019perceptual}.

\paragraph{Discussion.}
Research into learning value functions from video has been relatively scarce.
This is perhaps due to several associated challenges.
First, as noted above, there are issues related to missing action, reward, and goal labels in video.
Second, value functions estimate returns under a particular policy, but the behaviour in video data is highly multi-modal.
Third, TD learning from video may run into common issues seen in the offline RL literature \cite{Levine2020OfflineRL}, though the offline RL literature does present potential solutions \cite{kumar2020conservative,zhou2021plas}.
We also note that, if the goal is to obtain a policy, it may be easier to attempt this directly from video (see Section \ref{sec:policy}).
Nevertheless, \textcite{Bhateja2023RoboticOR} show that TD-learning of value functions from large-scale human video is a promising direction for real-world robotics.

\section{Datasets and Benchmarks}
\label{sec:datasets_benchmarks}

This section first provides details and discussions regarding video datasets in LfV (Section \ref{sec:datasets}), before moving to the benchmarking of LfV methods (Section \ref{sec:benchmarks}).

\subsection{Datasets}
\label{sec:datasets}

\begin{figure}[t]
    \includegraphics[width=0.7\textwidth]{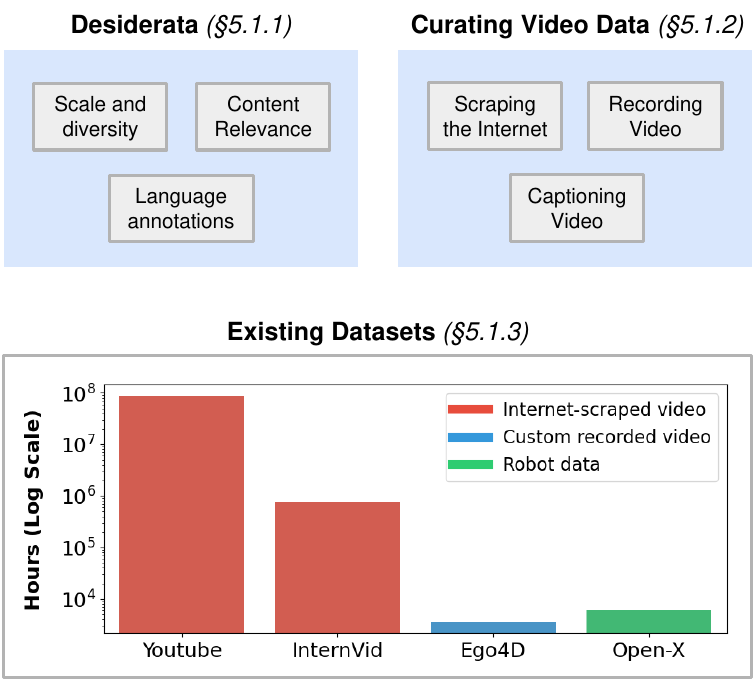}
    \centering
    \caption{\textbf{LfV Datasets} (Section \ref{sec:datasets}). The log-scale plot at the bottom compares the sizes of the largest curated open-source datasets in three different categories. InternVid \cite{Wang2023InternVidAL}, Ego4D \cite{Grauman2021Ego4DAT}, and Open X-Embodiment \cite{padalkar2023open} are the largest internet-scraped video, custom-recorded video, and robot datasets, respectively (to the best of our knowledge). We estimated the hours in the Open X-Embodiment dataset by assuming the average trajectory length is 10 seconds (at the time of writing, the dataset contained $\sim$2 million trajectories). The number of hours of video on YouTube is a rough calculation \cite{sjoberg_2023}.}
    \label{fig:datasets}
\end{figure}

We now turn our attention to the video datasets themselves.
In this section (see Figure \ref{fig:datasets}), we will discuss the desired properties of video datasets (Section \ref{sec:dataset_desiderata}), summarise methods for curating video data (Section \ref{sec:dataset_curating}), and review existing datasets and their limitations (Section \ref{sec:dataset_existing}).

\subsubsection{Desiderata}

\label{sec:dataset_desiderata}

We now discuss the key properties and characteristics desired from our LfV video datasets.

\paragraph{Scale and Diversity.}
The dataset should be large in scale and high in diversity. Scale can be measured in terms of total video duration.
Diversity refers to variety in the content of the data.
Increased scale and diversity of training data can reliably improve deep learning performances and generalization \cite{Brown2020LanguageMA}.
LfV is promising precisely due to the massive scale and diversity of video data available on the internet.

\paragraph{Content Relevance.}
The content and information in the video data should be relevant to downstream generalist robot settings \cite{khazatsky2024droid}.
For example, the videos should include information regarding the dynamics of the world and how embodied agents can complete physical tasks.
Crucially, the video data should have good coverage over the tasks and environments the generalist robot is likely to encounter.

\paragraph{Language Annotations.}
Language annotations can be useful for training language-conditioned models \cite{betker2023improving,videoworldsimulators2024} and to aid semantic representation learning from video \cite{Zhao2024VideoPrismAF}.
High-level annotations can be useful for learning abstracted representations. 
Granular annotations---for example, detailed descriptions of actions, spatial information, and object relations---can aid the learning of low-level features that are particularly relevant to robotics.

\paragraph{Other Desiderata.}
(1) Many LfV methods assume continuity within a clip (i.e., an absence of sharp scene transitions) \cite{Bruce2024GenieGI,Yang2023LearningIR}.
(2) The use of high-resolution video may benefit finer-grained downstream behaviours \cite{li2024visual}.
(3) Longer video clips may benefit long-horizon representations and memory useful in multi-step tasks \cite{Liu2024WorldMO,Wang2023MimicPlayLI}. (4) Video can be paired with other useful information, besides language annotations.
This includes object bounding boxes, object segmentations, or human-pose estimates --- these labels can aid representation learning efforts \cite{Bahl2023AffordancesFH}, serve as action labels \cite{hoque2025egodex}, or inform automated language captioning \cite{blank2024scaling}.
Modalities such as audio or depth information \cite{Grauman2021Ego4DAT} can provide additional information beyond what is contained in RGB pixels.

\subsubsection{Curating Video Data}

\label{sec:dataset_curating}

Here, we give concrete details on techniques for curating video datasets. We discuss techniques for scraping video from the internet, techniques for recording custom video, and techniques for (manually and automatically) annotating video.

\paragraph{Scraping Video from the Internet.} 
Diverse, large-scale video datasets can be obtained from internet repositories of pre-existing video data.
Techniques here focus on ensuring only relevant and high-quality videos are scraped from these repositories.
A typical pipeline for scraping internet video involves:
(1) \textit{Formulating a pool of query prompts} used to search the repository for candidate video with relevant content.
Previous works have constructed this pool, for example, via:
surveys of human time-use \cite{Grauman2021Ego4DAT,Wang2023InternVidAL};
crowd-sourcing \cite{sermanet2023robovqa,Goyal2017TheS};
or by using LLMs to parse action prompts from text corpora \cite{Wang2023InternVidAL}
(2) \textit{Post-processing the pool of candidate videos.}
This has included
processing videos into clips which contain no `cuts' between scenes \cite{blattmann2023stable}
and using video meta data (e.g., view counts) to identify and remove low-quality video \cite{xue2022advancing,Miech2019HowTo100MLA,nagrani2022learning}.
Automated metrics such as optical flow \cite{blattmann2023stable}, image-language model embeddings \cite{schuhmann2022laion,blattmann2023stable}, and video `realism' estimates \cite{Shan2020UnderstandingHH} have also been used to filter video data.
(3) \textit{Ensuring adherence to legal, ethical, and safety standards.}
This should include only selecting appropriately licensed data \cite{Smith2019AVIDLM}, ensuring privacy by anonymizing faces \cite{Smith2019AVIDLM}, filtering out unsafe content \cite{ju2024miradata}, and minimizing and explicitly acknowledging potential biases \cite{Smith2019AVIDLM,Miech2019HowTo100MLA}.
(4) \textit{Re-annotating the videos to improve the quality of labels.} We describe video captioning methods in detail in the below paragraphs.

\paragraph{Recording New Video.}


Manually recording custom videos can be an effective (but expensive) means of collecting video relevant to specific robot tasks \cite{sermanet2023robovqa,grauman2023ego} or embodiments \cite{sermanet2023robovqa,shafiullah2023bringing}.
Large-scale recording of custom video datasets has mainly been achieved via crowd-sourcing \cite{Goyal2017TheS,Grauman2021Ego4DAT}.
\textcite{damen2018scaling,Grauman2021Ego4DAT} ask participants to record their daily lives,
while \textcite{Goyal2017TheS,sermanet2023robovqa,grauman2023ego} provide participants with text instructions regarding the behaviours they should perform.
Video can be collected with a diversity of embodiments, including human arms, human-controlled manual grippers, and robot arms \cite{sermanet2023robovqa,shafiullah2023bringing}.
Recording first-person video \cite{Grauman2021Ego4DAT}---video which may be less common on the internet---could be particularly relevant to a robot with first-person observations.
Attempts to record custom video should adhere to ethical standards --- including minimizing bias by ensuring diversity across various demographic and contextual dimensions, and ensuring individuals in the video give consent or have their privacy protected (e.g., via face blurring) \cite{grauman2023ego}.

\paragraph{Manual Captioning.}
\label{para:manual_annotation}

The benefits that language captioning of video can provide are outlined in Section \ref{sec:dataset_desiderata}.
Previous works have often obtained descriptions of the contents of video data via manual human labelling \cite{blattmann2023stable,damen2018scaling,grauman2023ego,Grauman2021Ego4DAT}.
This can be performed by third-party workers or (where applicable) the person recording the video.
Versus to typed annotations, \textcite{damen2018scaling} find spoken annotations to result in higher quantity and quality of captions.
\textcite{grauman2023ego} have the person recording the video explain their thought processes whilst performing actions, and then have third-party annotators provide retrospective expert commentaries.
After initial annotation is performed, \textcite{maaz2023video} prompt workers to augment the descriptiveness of the video captions.

\paragraph{Automated Captioning.}
Manual human annotations are expensive \cite{Grauman2021Ego4DAT}.
For larger-scale datasets, automated annotation pipelines are promising.
Various methods have been proposed here.

\begin{itemize}

    \item \textit{Automatic speech recognition (ASR), metadata, and others.}
    ASR converts speech in the video's audio into text \cite{Miech2019HowTo100MLA,xue2022advancing,zellers2021merlot}.
    However, raw ASR captions can be noisy and unrelated to the contents of the video.
    Another source of information is corresponding metadata; such as descriptions, tags, and titles \cite{Wang2023InternVidAL}.
    \textcite{bain2021frozen} obtain captions from the alt-text HTML attribute associated with web images and videos \cite{bain2021frozen} (`Alt-text' in Table \ref{table:overview_of_existing_datasets}).
    \textcite{nagrani2022learning} start with a dataset of high-quality image-caption pairs, then mine video clips with similar frames and transfer the captions to those clips (`Transfer' in Table \ref{table:overview_of_existing_datasets}).
    
    \item \textit{Off-the-shelf vision models}.
    Video-to-text captioning models are an option \cite{blattmann2023stable}, but poor captioning capabilities have limited previous usage.
    Alternatively, more reliable image-captioning \cite{blattmann2023stable,maaz2023video} or object-detection \cite{zeng2022socratic,maaz2023video,ranasinghe2024understanding,blank2024scaling} can be employed to gather information regarding the content in a video.
    The methods in this bullet are referred to as `Generated' in Table \ref{table:overview_of_existing_datasets}.
    
    \item \textit{LLM processing.}
    \textcite{blattmann2023stable,maaz2023video,Wang2023InternVidAL} use LLMs to summarise keyframe captions and full-clip captions into a single video caption.
    LLM post-processing can be used to filter out inconsistent captions across sources or frames \cite{maaz2023video}.
    Combining multiple sources of information via LLM processing may result in more detailed and accurate captions.

\end{itemize}

\subsubsection{Existing Datasets}

\label{sec:dataset_existing}

In this section, we provide an overview of existing video datasets relevant to LfV. We aim to highlight datasets that satisfy key desiderata from Section \ref{sec:dataset_desiderata}, and thus are promising for training foundational video and/or robotics models. 

\paragraph{Overview of Existing Datasets.}
Table \ref{table:overview_of_existing_datasets} presents details of a representative set of existing video datasets. We now discuss these datasets in more detail.

\begin{itemize}
    \item \textit{Large-scale, internet-scraped datasets.} 
    These can span up to several decades worth of video data \cite{Wang2023InternVidAL,xue2022advancing}.
    They are diverse, capturing human behaviours in tasks and environments sampled from a global population.
    These datasets have been constructed via:
    query-based searches of YouTube \cite{Wang2023InternVidAL,Miech2019HowTo100MLA,stroud2020learning};
    aggregating video from previous datasets \cite{zellers2021merlot};
    or scraping video from various webpages \cite{bain2021frozen}.
    Language captions of these large datasets are usually obtained via automated methods.
    These datasets have commonly been used in initial attempts at training video foundation models \cite{Zhao2024VideoPrismAF,Wang2023VideoMAEVS} (see Section \ref{sec:video_FMs}).

    \item \textit{Manually collected datasets.}
    These are generally not as large or diverse as the largest web-obtained video datasets.
    However, many contain content highly relevant to robotics, and these datasets have often been used in past LfV research \cite{Nair2022R3MAU,Yang2023LearningIR,Wu2023UnleashingLV}.
    For example, Ego4D \cite{Grauman2021Ego4DAT} contains egocentric video of participants going about their daily lives, and RoboVQA \cite{sermanet2023robovqa} contains video of teleoperated robots and humans performing long-horizon household tasks.
    Here, it is common for captions to be provided manually by crowd-sourced workers.

    \item \textit{Annotations.}
    All of the datasets in Table \ref{table:overview_of_existing_datasets} come with text annotations.
    Other annotation modalities are sometimes provided.
    \textcite{Grauman2021Ego4DAT,grauman2023ego} occasionally collect corresponding audio, 3D meshes, eye-gazes, stereo, and synchronized video from different viewpoints.
    \textcite{Shan2020UnderstandingHH} provide human-hand labels for 100k images in their video dataset.
    Note, internet-obtained video datasets generally lack these other annotations as they may be costly to add manually, or may require customized video recording setups.
    
\end{itemize}

\begin{table}[t]
\small
\centering
\caption{\textbf{Existing video datasets}. Listed are: (top) large-scale, internet-scraped video datasets, and (bottom) robotics-relevant, manually-recorded video datasets. The datasets are ordered by decreasing total video duration. Details regarding `Caption Type' can be found in Section \ref{sec:dataset_curating}.}
\begin{tabularx}{\textwidth}{p{4cm}|XXXXXX}
\toprule
\textbf{Dataset} & \textbf{Content} & \textbf{Size (hours)} & \textbf{\# Clips} & \textbf{Caption Type} & \textbf{Collection Method} \\
\midrule
InternVid \cite{Wang2023InternVidAL} & YouTube & 760,000 & 230M & Generated & Internet \\ 
HD-VILA-100M \cite{xue2022advancing} & YouTube & 370,000 & 103M & ASR & Internet \\ 
YT-Temporal-180M \cite{zellers2021merlot} & YouTube & - & 180M & ASR & Internet \\ 
WTS-70M \cite{stroud2020learning} & YouTube & 190,000 & 70M & Metadata & Internet \\
HowTo100M \cite{Miech2019HowTo100MLA} & Instruction & 134,000 & 136M & ASR & Internet \\
WebVid-10M \cite{bain2021frozen}\footnotemark[1] & YouTube & 52,000 & 10M & Alt-text & Internet \\ 
VideoCC3M \cite{nagrani2022learning}\footnotemark[1] & YouTube & 18,000 & 6M & Transfer & Internet \\ 
100 Days of Hands \cite{Shan2020UnderstandingHH} & Actions & 3,100 & 27k & Metadata & Internet \\ 
\midrule
Ego-4D \cite{Grauman2021Ego4DAT} & Everyday & 3,600 & 28k & Manual & Manual \\ 
Ego-Exo-4D \cite{grauman2023ego} & Skilled & 1,400 & 6k & Manual & Manual \\ 
SS-v2 \cite{Goyal2017TheS} & Actions & 245 & 221k & Manual & Manual \\
RoboVQA \cite{sermanet2023robovqa} & Everyday & 230 & 98k & Manual & Manual \\ 
Epic-Kitchens-100 \cite{damen2022rescaling} & Cooking & 100 & 700 & Manual & Manual \\
\bottomrule
\end{tabularx}
\label{table:overview_of_existing_datasets}
\end{table}

\setcounter{footnote}{0}  
\footnotetext[1]{These datasets are not publicly available at the time of writing.}

\paragraph{Discussion.} 

The largest internet-curated video datasets are two orders of magnitude larger than the largest manually-recorded video datasets.
Yet, these still barely scratch the surface of the full range of video content available online (see Figure \ref{fig:datasets}).
As such, we advocate for continued efforts into curating ever-larger internet video datasets for LfV (whilst upholding high legal, ethical, and safety standards).
Crucially, new curation efforts should optimise for the full range of desiderata discussed in Section \ref{sec:dataset_desiderata}).
Low-quality language annotations are a key limitation of current internet-curated video datasets.
This can be addressed via improved automated captioning methods.
Finally, we note that efforts to curate smaller-scale, higher-quality video datasets can still provide value.
For example, such high-quality datasets can be used for finetuning models after pretraining on larger, lower-quality data \cite{blattmann2023stable}.

\subsection{Benchmarks}
\label{sec:benchmarks}

Benchmarks can be crucial for catalysing rapid research progress in a given area \cite{deng2009imagenet}.
In this section, we first give recommendations for how an LfV benchmark should be designed (Section \ref{sec:benchmarks_designing}).
We subsequently review relevant benchmarks from the literature, commenting on limitations and proposing improvements (Section \ref{sec:benchmarks_existing}).

\subsubsection{Designing LfV Benchmarks}
\label{sec:benchmarks_designing}

In the paragraphs below, we first give recommendations regarding the metrics an LfV benchmark should measure, before outlining different possible categories of LfV benchmarks and their corresponding desiderata.

\paragraph{Metrics.}
An LfV benchmark should serve to evaluate either
the capabilities of a policy obtained via an LfV approach,
or the effectiveness of an LfV algorithm at producing a policy, under certain constraints (e.g., when constrained to use a fixed dataset).
More specifically, the benchmark can evaluate the extent to which the potential benefits of LfV (Section \ref{sec:benefits}) have been obtained.
This can be achieved via the following quantitative metrics:

\begin{enumerate}
    \item \textit{Performance in-distribution of $\mathcal{D}_{robot}$.} Performance (e.g., task success rate) of the robot in settings in-distribution of $\mathcal{D}_{robot}$, after training on $\mathcal{D}_{video}$ and a fixed $\mathcal{D}_{robot}$.
    \item \textit{Data-efficiency in-distribution of $\mathcal{D}_{robot}$.} The quantity of data in $\mathcal{D}_{robot}$ required to reach a certain performance level in settings in-distribution of $\mathcal{D}_{robot}$, when also training on $\mathcal{D}_{video}$.
    \item \textit{Generalization beyond $\mathcal{D}_{robot}$.} The performance of the robot in settings out-of-distribution of $\mathcal{D}_{robot}$, after training on $\mathcal{D}_{video}$ and a fixed $\mathcal{D}_{robot}$.
\end{enumerate}

These metrics cover the benefits of `generalization beyond $\mathcal{D}_{robot}$' and `improvements in-distribution of $\mathcal{D}_{robot}$', as outlined in Section \ref{sec:benefits}.
However, the benefit of `emergent capabilities' likely must be evaluated qualitatively.

Finally, the benchmark should also inform us on
how scalable the LfV method is to diverse internet data and the generalist robot setting,
and how well it can handle fundamental LfV challenges (Section \ref{sec:challenges}).
Specific design choices that can be made here are discussed in later paragraphs.

\paragraph{Categories.}
All LfV benchmarks should include a fixed set of evaluation environments, $\mathcal{M}_{\text{eval}}$, in which the LfV policy, $\pi_{\text{lfv}}$, is to be evaluated. 
However, LfV benchmarks can differ based on whether they specify the datasets $\mathcal{D}_{\text{video}}$ and $\mathcal{D}_{\text{robot}}$ that are to be used to train the policy.
Concretely, we identify the following possible categories of LfV benchmark:

\begin{itemize}
    \item \textbf{$B_{\text{e}}$ = \{$\mathcal{M}_{\text{eval}}$\}.} A $\pi_{\text{lfv}}$ trained on any $\mathcal{D}_{\text{video}}$ and any $\mathcal{D}_{\text{robot}}$ is evaluated on a fixed $\mathcal{M}_{\text{eval}}$.

    \item \textbf{$B_{\text{e-r}}$ = \{$\mathcal{M}_{\text{eval}}$, $\mathcal{D}_{\text{robot}}$\}.} A $\pi_{\text{lfv}}$ trained on any $\mathcal{D}_{\text{video}}$ and a \emph{fixed} $\mathcal{D}_{\text{robot}}$ is evaluated on a fixed $\mathcal{M}_{\text{eval}}$.

    \item \textbf{$B_{\text{e-r-v}}$ = \{$\mathcal{M}_{\text{eval}}$, $\mathcal{D}_{\text{robot}}$,  $\mathcal{D}_{\text{video}}$\}.}
    A $\pi_{\text{lfv}}$ trained on a \emph{fixed} $\mathcal{D}_{\text{video}}$ and a \emph{fixed} $\mathcal{D}_{\text{robot}}$ is evaluated on a fixed $\mathcal{M}_{\text{eval}}$.

\end{itemize}

Benchmarks that fix the datasets (e.g., $B_{\text{e-r-v}}$) can provide a fair comparison of LfV algorithms.
However, they may end up being `toyish' in practice.
Benchmarks that do not fix the data (i.e., $B_{\text{e}}$) are useful for evaluating the performance of an obtained $\pi_{\text{lfv}}$.

\paragraph{Desiderata.} We now give details regarding the desiderata for each potential component of an LfV benchmark.

\begin{itemize}
    \item \textbf{$\mathcal{M}_{\text{eval}}.$} The evaluation environments should be analogous to the generalist robot settings we are interested in.
    (1) \emph{Relevance:} The environments and tasks should resemble those in generalist robot settings.
    (2) \emph{Diversity} in $\mathcal{M}_{\text{eval}}$ allows us to measure how $\pi_{\text{lfv}}$ will handle diverse and unseen real-world scenarios.
    (3) \emph{Realism:} The benchmark should include challenges that may be faced in real-world environments (such as noisy observations), and its physics should be sufficiently realistic.
    To ensuring $\mathcal{M}_{eval}$ bears on the challenge of `missing low-level information in video', the tasks in $\mathcal{M}_{eval}$ could be setup to require perception of information not contained in video data.

    \item \textbf{$\mathcal{D}_{\text{robot}}$.}
    (1) $\mathcal{D}_{\text{robot}}$ should be drawn from a \emph{subset} of the tasks and environments in $\mathcal{M}_{\text{eval}}$, allowing us to measure performance both in and out-of-distribution of $\mathcal{D}_{\text{robot}}$.
    (2) To ensure LfV generalization (see Figure \ref{fig:lfv_generalize}) is indeed possible in the benchmark, the robot dataset should have good coverage over all possible low-level ``atomic" actions.

    \item \textbf{$\mathcal{D}_{\text{video}}$.} Firstly, $\mathcal{D}_{\text{video}}$ should not contain action labels. Other desiderata depend on the extent to which we wish $B_{\text{e-r-v}}$ to be a `toy' setting for testing LfV algorithms.
    
    \emph{Toy setups} can allow for excellent control over which LfV challenges (Section \ref{sec:challenges}) are faced.
    (1) A toy setup can ensure $\mathcal{D}_{\text{video}}$ has good coverage over the environments and tasks in $\mathcal{M}_{\text{eval}}$.
    (2) Characteristics of $\mathcal{D}_{\text{video}}$ can be controlled to establish distribution shifts between $\mathcal{D}_{\text{video}}$ and $\mathcal{M}_{\text{eval}}$ (e.g., embodiment gaps, viewpoint differences).
    (3) The challenge of `controlability' (see Section \ref{sec:challenges}) can be toggled by managing the extent to which changes in the video are due to effects beyond a single agent's actions.

    \emph{Approximating the scaled-up LfV setting:} Here, $\mathcal{D}_{\text{video}}$ should resemble internet video in terms of its scale, diversity, and content. Thus, $\mathcal{D}_{\text{video}}$ should ideally consist of real-world human videos scraped from the internet. Doing so will inherently present several LfV challenges, and allow for evaluation of the scalability of the LfV method.

\end{itemize}

\subsubsection{Existing Benchmarks}

\label{sec:benchmarks_existing}

This section briefly details existing LfV benchmarks, before discussing future directions.

\paragraph{$B_{\text{e}} = \{\mathcal{M}_{\text{eval}}\}$.} 
A number of benchmarks provide an $\mathcal{M}_{\text{eval}}$ relevant to LfV research.
(1) \textit{Toy settings:}
The diversity of certain complex or open-ended video game environments \cite{fan2022minedojo,kuttler2020nethack} can provide an excellent setting for testing LfV policy generalization.
(2) \textit{Robotics simulators:}
Robotics-relevant simulated environments include benchmarks focused on low-level motor control and object interaction \cite{mees2022calvin,yu2020meta,gu2023maniskill2,liu2024libero,kumar2024robohive,makoviychuk2021isaac,pumacay2024colosseum}.
(3) \textit{Embodied AI simulators:} Some benchmarks focus on household and everyday mobile manipulation tasks while abstracting away low-level control to focus on higher-level planning \cite{puig2023habitat,kolve2017ai2,li2024behavior}.
(4) \textit{Real-robot setups:}
Real-world robot evaluations are also available \cite{ahn2020robel}, but may be more costly and time-consuming than simulated evaluation.
To address this, \textcite{li2024evaluating} design a simulated benchmark specifically for evaluating real-world policies.

\paragraph{$B_{\text{e-r}} = \{\mathcal{M}_{\text{eval}}, \mathcal{D}_{\text{robot}}\}$.}
Several benchmarks pair a $\mathcal{D}_{\text{robot}}$ with an $\mathcal{M}_{\text{eval}}$.
\textcite{mees2022calvin,lynch2023interactive,liu2024libero} provide demonstration or play data for simulated tabletop manipulation tasks.
\textcite{gu2023maniskill2} includes mobile manipulation tasks.
\textcite{pumacay2024colosseum,xie2023decomposing,nasiriany2024robocasa} provide multiple different tasks and environments (along with corresponding $\mathcal{D}_{\text{robot}}$ generation methods), which may be useful for testing LfV generalization.
Lastly, there exist relatively large real-world robot datasets which are not paired with any specific benchmark \cite{padalkar2023open,khazatsky2024droid}.

\paragraph{$B_{\text{e-r-v}} = \{\mathcal{M}_{\text{eval}}, \mathcal{D}_{\text{robot}}, \mathcal{D}_{\text{video}}\}$.} 
To the best of our knowledge, there are few benchmarks that specify a fixed $\mathcal{D}_{\text{video}}$ and a fixed $\mathcal{D}_{\text{robot}}$.
\textcite{fan2022minedojo} provide a $\mathcal{D}_{\text{video}}$ containing 730k YouTube videos in a Minecraft setting.
Many previous LfV works have constructed a $\mathcal{D}_{video}$ by stripping action labels from a $\mathcal{D}_{robot}$ \cite{Seo2022ReinforcementLW,Schmidt2023LearningTA,Wen2023AnypointTM}.
This can allow for easy setup of a scenario where $\mathcal{D}_{\text{video}}$ has good coverage over $\mathcal{M}_{\text{eval}}$, but can neglect distribution shifts seen between the video data and the robot domain in realistic LfV settings.

\paragraph{Discussion.}

We provide recommendations for improvements in LfV benchmarking.

\begin{itemize}
    \item \emph{Improving the diversity in $\mathcal{M}_{\text{eval}}$}. The diversity in most suitable robotic simulators \cite{mees2022calvin,lynch2023interactive,liu2024libero,gu2023maniskill2,li2024behavior} is still limited in terms of the tasks, environments, and objects presented.
    Improving diversity will allow us to better assess the applicability of the LfV method to generalist robot settings.
    One possibility here is to use procedural generation \cite{deitke2022} or LLM-assisted environment design \cite{Xian2023TowardsGR}.
    Another is to aggregate multiple $\mathcal{M}_{\text{eval}}$'s within a common framework \cite{kumar2024robohive}.
    
    \item \emph{Establishing popular $B_{\text{e-r}}$ and $B_{\text{e-r-v}}$ LfV benchmarks.}
    Establishing benchmarks in these categories (currently there are no popular options) would provided an improved ability to compare LfV algorithms;
    past works have often chosen different $\mathcal{D}_{\text{video}}$'s \cite{Nair2022R3MAU,Yang2023LearningIR,Schmidt2023LearningTA} and thus are difficult to compare.
    Any new LfV benchmark should be designed following the recommendations outlined in Section \ref{sec:benchmarks_designing}. 
    
\end{itemize}

\section{Challenges and Opportunities}
\label{sec:challenges_opportunities}
We now provide a comprehensive discussion of challenges and opportunities for future LfV research, based on our analysis of the existing literature. First, we give high-level recommendations for future LfV research (Section \ref{sec:opportunities_recommendations}). Second, we detail promising directions for utilising video foundation models and techniques for LfV (Section \ref{sec:opportunities_video_fms}). Third, we highlight approaches for overcoming previously identified LfV challenges (Section \ref{sec:opportunities_challenges}). We conclude by discussing other challenges in generalist robotics may not be solved via scaling to larger datasets (Section \ref{sec:opportunities_other}).

\subsection{High-level Recommendations}


\label{sec:opportunities_recommendations}

Following our the analysis of the LfV literature in Section \ref{sec:LfV_methods}, we now provide some high-level recommendations for future LfV research.

\paragraph{Focus on scalable approaches.}
Methods that can scale well to large, diverse internet video may be particularly effective.
Many previous LfV works limit scalability by making strong assumptions on the nature of the video data or the downstream robot setting \cite{Baker2022VideoP,stadie2017third}.
Instead, we advocate for methods that can use general learning objectives to extract knowledge from heterogenous video data (see the foundation model approaches in Section \ref{sec:video_FMs}).
Methods that can learn from unlabelled and suboptimal video data are most scalable in theory \cite{Bruce2024GenieGI}, though methods that leverage language labels can still be highly effective \cite{videoworldsimulators2024,Yang2023LearningIR}.
Finally, focusing initially on LfV approaches that can learn from offline robot data (or that can perform online RL in simulation) may be sensible, due to the impracticalities of online real-world RL.

\paragraph{Target the key benefits of LfV via policies and dynamics models.}
Future LfV research should focus on obtaining the most promising LfV benefits (Section \ref{sec:benefits}); namely, improving robot data efficiency and improving generalization beyond $\mathcal{D}_{\text{robot}}$.
Much past LfV research has sought to extract reward functions from video (Section \ref{sec:reward_funcs}).
Whilst these approaches are useful, learning policies (Section \ref{sec:policy}) and dynamics models (Section \ref{sec:dynamics}) is likely most promising for reducing reliance on robot data (as is discussed in their respective sections).

\paragraph{Improve LfV datasets.}
Improving and scaling video datasets is a reliable way to improve our LfV models.
In Section \ref{sec:dataset_existing}, we recommend scraping larger-scale video data from the internet, and captioning this data with high-quality language annotations.
We advocate for the open-sourcing of such curated video datasets.

\paragraph{Combine LfV with alternative approaches, like simulation.}
Any practical LfV approach should seek to collect as useful a $\mathcal{D}_{\text{robot}}$ as possible.
In some cases, video data could aid the collection of (real and simulated) robot data.
Automatic generation of diverse and realistic simulated environments \cite{Xian2023TowardsGR,faldor2024omni} could be aided by video data \cite{Yang2024SimEndoGSED} and video models \cite{videoworldsimulators2024}.
Here, video foundation models could be used to provide dense reward signals in diverse simulated environments (Section \ref{sec:reward_funcs}, \textcite{li2024sds}).
Elsewhere, we note that the use of video data is not restricted to monolithic foundation model approaches or pure end-to-end learning approaches.
Components in modular approaches, such as specialist agents \cite{yang2020multi} or vision systems, may benefit from the use of internet video data.
In hybrid approaches, where deep learning is combined with more classical or structured control methods \cite{dalal2023imitating,huang2023voxposer,mishra2023generative}, video data can benefit the deep learning component.

\paragraph{Improve evaluation of LfV methods.}
Research should explicitly measure the extent to which a method can handle LfV challenges and can provide the potential benefits of LfV.
In Section \ref{sec:benchmarks_designing}, we outline concrete metrics for this.
It is often difficult to identify these metrics in existing LfV research.
In Section \ref{sec:benchmarks_existing}, we advocate for the design of improved LfV benchmarks that can offer these metrics off-the-shelf.

\subsection{Video Foundation Models for LfV}

\label{sec:opportunities_video_fms}

A promising LfV direction is to utilise large-scale internet video data to help train large foundation models (FMs) for robotics.
Here, we outline related research avenues and key points of discussion.

\paragraph{Improving Video FMs.}
Advancing video FMs and techniques may be a key driver of progress for scalable LfV approaches (Section \ref{sec:video_FMs}).
Key issues to tackle here include addressing dataset limitations and prohibitive computational requirements.
More details on avenues for improving video FMs are discussed in Section \ref{sec:video_fm_challenges}.

\paragraph{Leveraging Pretrained Video FMs.}
Generic video foundation models can be adapted into RL knowledge modalities to aid robot learning.
Strong baselines here will include finetuning the video FM into an action-outputting policy using offline robot data \cite{Brohan2023RT2VM,Wu2023UnleashingLV}, or (if the video FM has video generation capabilities) finetuning it into a robot dynamics model.
Data-efficient and computationally efficient adaptation mechanisms \cite{hu2021lora,Yang2023ProbabilisticAO} may be useful here.

\paragraph{Customising Video FMs for LfV.}

The most effective LfV approaches will likely use pipelines fully optimized for robotics.
We now touch on directions for customising video FMs and techniques for LfV.

\begin{itemize}

    \item \textit{Improve the capabilities of generic video FMs in areas relevant to robotics.}
    We could, for example, improve the physics realism of generic video prediction models (e.g., via RL finetuning; see \textcite{black2023training}), or the fine-grained understandings of generic video-to-text models (e.g., via improved fine-grained language captioning of the video data).
    The inclusion of domain-specific robot videos in generic video FM pretraining could improve its performance in the downstream robot setting.
    
    \item \textit{Pretrain control-centric video FMs on $\mathcal{D}_{\text{video}}$.}
    There are various options for obtaining models that are more `control-centric' from video data.
    The use of action representations (see Section \ref{sec:alt_actions}) can add explicit action information to the video FM \cite{Bruce2024GenieGI,Schmidt2023LearningTA}.
    The model could be trained using control-centric objectives, such as TD-learning \cite{Bhateja2023RoboticOR}.
    Video data could be paired with robotics-relevant low-level information (where available); e.g., 3D depth information can improve crucial 3D understandings \cite{zhen20243d,Yuan2024GeneralFA}, or frozen video features could be combined with features from different modalities \cite{lin2024spawnnet}.
    
    \item \textit{Pretrain control-centric FMs on $\mathcal{D}_{\text{video}}$ and $\mathcal{D}_{\text{robot}}$.}
    Any-to-any sequence models can be pretrained jointly on video and robot data, and can subsequently act as policies, dynamics models, and high-level planners \cite{sohn2024introducing}.
    Video prediction models can be pretrained on video and robot data, allowing them to be optionally conditioned on robot actions \cite{Yang2023LearningIR}.
    Action labels in robot data could be used in addition to action representations in video data.
    It is not yet clear whether pretraining on both $\mathcal{D}_{\text{video}}$ and $\mathcal{D}_{\text{robot}}$ is preferable to pretraining on $\mathcal{D}_{\text{video}}$ before finetuning on $\mathcal{D}_{\text{robot}}$.
    Future research should clarify this.
\end{itemize}

\paragraph{Monolithic Vs Compositional Models.}

Any-to-any sequence modelling is an example monolithic LfV approach \cite{sohn2024introducing}.
The use of a video generation model to hierarchically condition an action-prediction model is an example compositional LfV approach \cite{Ajay2023CompositionalFM}.
Versus monolithic approaches, \textcite{Du2024CompositionalGM} argue that compositional approaches are more computationally efficient, more data-efficient, and can obtain improved generalization \cite{zhou2024robodreamer}.
These advantages are very relevant to the LfV setting.
Nevertheless, monolithic models may benefit from positive transfer between the different data modalities and tasks they handle, and can be optimized end-to-end. 
Future research should seek to better compare monolithic and compositional LfV approaches.

\paragraph{Open-sourcing Video FMs.}
We advocate for increased open-sourcing of video foundation models.
Open-sourced models will make cutting-edge LfV research more accessible to academic researchers, accelerating progress in LfV \cite{kim2024openvla}.

\subsection{Overcoming LfV Challenges}
\label{sec:opportunities_challenges}


In Section \ref{sec:challenges}, we outlined the key challenges in LfV. Whilst recent LfV research has partially addressed some of these, here we discuss further promising directions.

\paragraph{Bridging the gap to low-level robot information.}


A key LfV challenge is to minimise the quantity of robot data required, despite the missing low-level information (e.g., forces and tactile information) in the video data (see Figure \ref{fig:lfv_generalize}). While it is possible that further scaling the robot and video datasets will implicitly solve this issue, here we discuss more explicit solutions.

\begin{itemize}

\item \emph{Incorporate low-level information into video pretraining.}
As discussed in Section \ref{sec:opportunities_video_fms}, low-level information could be introduced during video pretraining via:
the use of video paired with low-level information (e.g., depth info);
or the use of robot data during pretraining.
Simulated robot data could be cheap source of low-level information.

\item \emph{Bypass the need to incorporate low-level information into video pretraining.}
A hierarchical approach can mitigate the need for a video-pretrained model to make use of non-visual information.
For example, a high-level policy trained on $\mathcal{D}_{\text{video}}$ can propose alternative-actions, and a low-level policy trained on $\mathcal{D}_{\text{robot}}$ can decode these to robot actions \cite{Du2023LearningUP,Schmidt2023LearningTA,Wen2023AnypointTM} (see Section \ref{sec:policy}).
The use of a coarse-action space (e.g., cartesian control for manipulation; see \textcite{Brohan2022RT1RT}) may achieve a similar effect.
One possibility to collect robot data with the hierarchical/coarse action-space policy, and subsequently finetune for finer-grained control.

\end{itemize}

\paragraph{Recovering action information from video.}

The missing action label problem in LfV can be partially addressed using alternative-action representations (see Section \ref{sec:alt_actions}). We recommend future research to:
(1) Scale existing methods to realistic and diverse internet video.
(2) Explicitly compare the pros and cons of different action representations, and measure the extent to which each can help obtain the benefits of LfV (see Section \ref{sec:benefits}).
(3) Develop improved, novel action representations. Perhaps by combining the benefits of existing options \cite{Ye2024LatentAP}.

\paragraph{Tackling distribution shift.}
Previous LfV works have attempted to address LfV distribution shifts using specific inductive biases (see Section \ref{sec:reps_address_shifts}).
However, these have not scaled to heterogenous internet video data.
We advocate for implicitly addressing distribution shifts by scaling flexible methods to large, diverse internet video data, and, when possible, additionally training on (real and simulated) robot data.
Compositional approaches are another option for obtaining generalization under distribution shift \cite{Du2024CompositionalGM}.
Nevertheless, addressing LfV distribution shift is an open problem.

\paragraph{Mitigating computational demands.}
The high computational demands of training on video data can be mitigated via improved efficient architectures, taking inspiration from relevant LLM research \cite{Wang2023VideoMAEVS,liu2023ring,balavzevic2024memory,jiang2024mixtral}.
The ever decreasing cost of compute will also help here \cite{moore1998cramming}.
Improved methods for learning latent spaces in which the video model can operate (versus operating directly in pixel space) can aid computational efficiency in addition to mitigating issues related to noise and redundancy in video \cite{bardes2023v,Yan2021VideoGPTVG}.

\subsection{Is Scaling Enough?}
\label{sec:opportunities_other}

Observing results in other machine learning domains \cite{achiam2023gpt,betker2023improving,Yang2023LearningIR,sutton2019bitter}, it seems likely that scaling current robot learning approaches to larger datasets will yield improvements in robot capabilities.
This survey advocates for the use of internet video data to help us achieve this scaling.
However, scaling may be insufficient to take us all the way to generalist robots \cite{ieee_solve_robotics}.
Indeed, there may be unique challenges in robotics that do not exist in other domains where scaling has proven successful \cite{moravec1988}.
In this section, we note challenges that may face LfV approaches that rely on scaling the data under the current paradigm.
Instead, solving these challenges may require algorithmic advances.

\begin{itemize}
    \item \textit{LfV is also reliant on robot data.}
    LfV approaches augment a robot dataset with (internet) video data.
    However, any LfV approach will still be bottlenecked by the quantity, quality, and coverage of its robot data.
    The difficulty of collecting robot data with sufficient coverage over generalist robot settings may be the main barrier to scaling up robotics as seen in other domains, even for LfV approaches.

    \item \textit{Safety and reliability.} Safety is paramount when deploying robots in the real world --- failures can lead to damage to property, costly operational disruptions, or physical harm to people \cite{Layne2023RobotrelatedFA}.
    As such, physical robots may require higher success rates than are required in other generative AI tasks.
    Unfortunately, deep learning models are known to be brittle and often generalise poorly to unseen scenarios \cite{Goodfellow2014ExplainingAH,Hendrycks2019BenchmarkingNN}.
    Scaling to internet data may provide a robot model with improved `common-sense', improving its reliability and generalization \cite{zhao2023large,Brohan2023RT2VM}.
    However, this may not fundamentally solve the problem --- it may only improve generalization within the distribution of the internet data, but not beyond it.
    

    
    \item \textit{Long-horizon tasks and memory.}
    Generalist robots may need to operate over long time horizons.
    This presents two challenges.
    First, longer-horizon tasks may necessitate improved planning, reasoning, and subtask success rates.
    Second, longer-horizon tasks require improved memory capabilities.
    Current solutions to long-term memory involve scaling up the context-length of the model \cite{Liu2024WorldMO}, or storing and retrieving from simple memory databases \cite{park2023generative}.
    It is unclear whether these solutions will be sufficient to fundamentally address the problem.
    
    \item \textit{Latency.}
    Large foundation models currently have long inference times that limit their ability to perform real-time high-frequency robot control \cite{Reed2022AGA,Brohan2022RT1RT}. Techniques to reduce inference latency will be beneficial here \cite{jiang2024mixtral,song2024germ}.

    \item \textit{Continual learning and adaptation.}
    A generalist robot may regularly encounter novel scenarios.
    It must be able to adapt to these scenarios appropriately.
    Scaling can improve generalization and in-context `meta-learning' abilities \cite{Brown2020LanguageMA}.
    Nevertheless, obtaining true continual learning \cite{abel2024definition} abilities is still an open problem, and may require further algorithmic advances.
    
    \item \textit{Reasoning.}
    There exists both speculation and empirical evidence regarding the inability of deep learning methods to perform true `reasoning' \cite{huang2023large,anwar2024foundational}.
    Such reasoning abilities may be crucial for a generalist robot.
    If deep learning is fundamentally limited in this regard, then scaling to internet video data may be insufficient to achieve general-purpose robots.

\end{itemize}

Nevertheless, `Is scaling enough?' is ultimately an empirical question.
Given deep learning will likely be a core component of generalist robot research for the foreseeable future, there is value in work pushing scaling-based approaches as far as possible.
This will allow us to empirically answer this question and find fundamental and practical limitations that do indeed exist.
Identified limitations can be tackled in a targeted manner by combining deep learning with alternative approaches (Section \ref{sec:background}).

\section{Conclusion}
\label{sec:conclusion}
Developing general-purpose robots is a grand challenge in robotics.
However, current robot learning approaches are bottlenecked by a lack of robot data.
Learning from Video (LfV) methods seek to alleviate this issue by augmenting their training dataset with video data.
These methods are promising as internet video comes in vast quantities and contains information highly relevant a general-purpose robot.

In this survey, we have outlined fundamental LfV concepts and conducted a comprehensive review of the LfV literature.
We emphasised methods with the potential to scale to large, heterogeneous internet video datasets:
following the success of scaling deep learning to internet data in other domains \cite{OpenAI2023GPT4TR,betker2023improving}, these LfV methods are promising for improving the generality of our robots.

The key takeaways of this survey are summarised below.

\begin{itemize}
    \item \textit{Fundamental LfV concepts.}
    We explicitly establish the benefits that can be obtained from LfV (Section \ref{sec:benefits}), and key LfV challenges that stand in the way of these benefits (Section \ref{sec:challenges}).
    These fundamental concepts should guide future LfV research agendas. 
    
    \item \textit{Robot foundation models from internet video.} Video foundation model techniques are promising for extracting knowledge from large, heterogeneous internet video datasets (Section \ref{sec:video_FMs}).
    In Section \ref{sec:opportunities_video_fms}, we discussed directions for leveraging video foundation model techniques for robotics.
    Developing models customized for robotics is a particularly promising direction.
    Here, recent monolithic any-to-any sequence models offer a clear path forward.
    It will also be beneficial to pursue compositional approaches in parallel.
    
    \item \textit{Scalable approaches for learning policies and dynamics models.} 
    Our main analysis of the LfV-for-robotics literature (Section \ref{sec:applications}) yielded important takeaways.
    First, future LfV research should avoid inductive biases that limit scalability (i.e., those seen in Section \ref{sec:reps_address_shifts}), and adopt scalable learning techniques similar to those used for video foundation models (Section \ref{sec:video_FMs}).
    Though a reliance on language captions may limit scalability to an extent, this may become less of an issue as automated captioning methods improve (Section \ref{sec:dataset_curating}).
    Second, we should focus on methods that can best obtain the key LfV benefits (e.g., generalisation beyond the available robot data). This includes targeting the learning of policies (Section \ref{sec:policy}) and dynamics models (Section \ref{sec:dynamics}) from video data.
    Improved benchmarks that quantitatively measure LfV benefits (see Section \ref{sec:benchmarks_existing}) will facilitate these efforts.
    
    \item \textit{Action representations.} In Section \ref{sec:alt_actions}, we outlined methods for extracting action representations from video.
    These can serve to mitigate missing action labels in video.
    These methods are promising, but further work is required to scale them to large, realistic internet video data.
    
    \item \textit{Improved datasets.} Improving and scaling our video datasets (as per the desiderata outlined in Section \ref{sec:dataset_desiderata}) is a reliable way to improve our LfV models.
    Suitable methods for curating internet video data are outlined in Section \ref{sec:dataset_curating}.
    
    \item \textit{Is scaling enough?} Exploiting internet video will likely drive significant advances in robotics.
    However, the generalist robot setting presents challenges that may not be solved via naive scaling of deep learning (Section \ref{sec:opportunities_other}).
    As LfV and other research attempts to address the data bottleneck in robotics, other complementary and alternative research agendas should be explored in parallel (see Section \ref{sec:background}).
\end{itemize}

The analyses, taxonomies, and directions presented in this survey should serve as a valuable reference for future LfV research.
We hope this can catalyze further research in the area, and accelerate our progress towards developing general-purpose robots.

\begin{acks}
Daniel C.H. Tan is supported by the Agency for Science, Technology, and Research (A*STAR), Singapore. Fernando Acero and Nathan Herr are supported by the UKRI CDT in Foundational AI [EP/S021566/1] at the UCL Centre for Artificial Intelligence. We thank Anthony Hu for useful feedback in the early stages of the paper.
\end{acks}

\printbibliography

@article{Yang2023FoundationMF,
  title={Foundation Models for Decision Making: Problems, Methods, and Opportunities},
  author={Sherry Yang and Ofir Nachum and Yilun Du and Jason Wei and P. Abbeel and Dale Schuurmans},
  journal={ArXiv},
  year={2023},
  volume={abs/2303.04129},
  url={https://api.semanticscholar.org/CorpusID:257378587}
}

@misc{ho2022imagen,
  title={Imagen video: High definition video generation with diffusion models},
  author={Ho, Jonathan and Chan, William and Saharia, Chitwan and Whang, Jay and Gao, Ruiqi and Gritsenko, Alexey and Kingma, Diederik P and Poole, Ben and Norouzi, Mohammad and Fleet, David J and others},
  journal={arXiv preprint arXiv:2210.02303},
  year={2022}
}

@misc{singer2022make,
  title={Make-a-video: Text-to-video generation without text-video data},
  author={Singer, Uriel and Polyak, Adam and Hayes, Thomas and Yin, Xi and An, Jie and Zhang, Songyang and Hu, Qiyuan and Yang, Harry and Ashual, Oron and Gafni, Oran and others},
  journal={arXiv preprint arXiv:2209.14792},
  year={2022}
}

@inproceedings{Torabi2019RecentAI,
  title={Recent Advances in Imitation Learning from Observation},
  author={Faraz Torabi and Garrett Warnell and Peter Stone},
  booktitle={International Joint Conference on Artificial Intelligence},
  year={2019},
  url={https://api.semanticscholar.org/CorpusID:173188327}
}

@article{Ravichandar2020RecentAI,
  title={Recent Advances in Robot Learning from Demonstration},
  author={Harish Ravichandar and Athanasios S. Polydoros and Sonia Chernova and Aude Billard},
  journal={Annu. Rev. Control. Robotics Auton. Syst.},
  year={2020},
  volume={3},
  pages={297-330},
  url={https://api.semanticscholar.org/CorpusID:208958394}
}

@misc{zhang2024mm,
  title={Mm-llms: Recent advances in multimodal large language models},
  author={Zhang, Duzhen and Yu, Yahan and Li, Chenxing and Dong, Jiahua and Su, Dan and Chu, Chenhui and Yu, Dong},
  journal={arXiv preprint arXiv:2401.13601},
  year={2024}
}

@misc{tang2023video,
  title={Video understanding with large language models: A survey},
  author={Tang, Yunlong and Bi, Jing and Xu, Siting and Song, Luchuan and Liang, Susan and Wang, Teng and Zhang, Daoan and An, Jie and Lin, Jingyang and Zhu, Rongyi and others},
  journal={arXiv preprint arXiv:2312.17432},
  year={2023}
}

@article{schiappa2023self,
  title={Self-supervised learning for videos: A survey},
  author={Schiappa, Madeline C and Rawat, Yogesh S and Shah, Mubarak},
  journal={ACM Computing Surveys},
  volume={55},
  number={13s},
  pages={1--37},
  year={2023},
  publisher={ACM New York, NY}
}

@article{ho2022video,
  title={Video diffusion models},
  author={Ho, Jonathan and Salimans, Tim and Gritsenko, Alexey and Chan, William and Norouzi, Mohammad and Fleet, David J},
  journal={Advances in Neural Information Processing Systems},
  volume={35},
  pages={8633--8646},
  year={2022}
}

@inproceedings{Zhao2022WhatMR,
  title={What Makes Representation Learning from Videos Hard for Control?},
  author={Tony Zhao and Siddharth Karamcheti and Thomas Kollar and Chelsea Finn and Percy Liang},
  year={2022},
  url={https://api.semanticscholar.org/CorpusID:252635608}
}

@article{Burns2023WhatMP,
  title={What Makes Pre-Trained Visual Representations Successful for Robust Manipulation?},
  author={Kaylee Burns and Zach Witzel and Jubayer Ibn Hamid and Tianhe Yu and Chelsea Finn and Karol Hausman},
  journal={ArXiv},
  year={2023},
  volume={abs/2312.12444},
  url={https://api.semanticscholar.org/CorpusID:266369884}
}

@inproceedings{Hu2023ForPV,
  title={For Pre-Trained Vision Models in Motor Control, Not All Policy Learning Methods are Created Equal},
  author={Yingdong Hu and Renhao Wang and Li Li and Yang Gao},
  booktitle={International Conference on Machine Learning},
  year={2023},
  url={https://api.semanticscholar.org/CorpusID:258048578}
}

@article{Yan2021VideoGPTVG,
  title={VideoGPT: Video Generation using VQ-VAE and Transformers},
  author={Wilson Yan and Yunzhi Zhang and P. Abbeel and A. Srinivas},
  journal={ArXiv},
  year={2021},
  volume={abs/2104.10157},
  url={https://api.semanticscholar.org/CorpusID:233307257}
}

@inproceedings{ge2022long,
  title={Long video generation with time-agnostic vqgan and time-sensitive transformer},
  author={Ge, Songwei and Hayes, Thomas and Yang, Harry and Yin, Xi and Pang, Guan and Jacobs, David and Huang, Jia-Bin and Parikh, Devi},
  booktitle={European Conference on Computer Vision},
  pages={102--118},
  year={2022},
  organization={Springer}
}

@misc{gu2023mamba,
  title={Mamba: Linear-time sequence modeling with selective state spaces},
  author={Gu, Albert and Dao, Tri},
  journal={arXiv preprint arXiv:2312.00752},
  year={2023}
}

@misc{narang2022pathways,
  title={Pathways language model (palm): Scaling to 540 billion parameters for breakthrough performance},
  author={Narang, Sharan and Chowdhery, Aakanksha},
  journal={Google AI Blog},
  year={2022}
}

@article{radford2019language,
  title={Language models are unsupervised multitask learners},
  author={Radford, Alec and Wu, Jeffrey and Child, Rewon and Luan, David and Amodei, Dario and Sutskever, Ilya and others},
  journal={OpenAI blog},
  volume={1},
  number={8},
  pages={9},
  year={2019}
}

@misc{kim2024openvla,
  title={OpenVLA: An Open-Source Vision-Language-Action Model},
  author={Kim, Moo Jin and Pertsch, Karl and Karamcheti, Siddharth and Xiao, Ted and Balakrishna, Ashwin and Nair, Suraj and Rafailov, Rafael and Foster, Ethan and Lam, Grace and Sanketi, Pannag and others},
  journal={arXiv preprint arXiv:2406.09246},
  year={2024}
}

@misc{balavzevic2024memory,
  title={Memory Consolidation Enables Long-Context Video Understanding},
  author={Bala{\v{z}}evi{\'c}, Ivana and Shi, Yuge and Papalampidi, Pinelopi and Chaabouni, Rahma and Koppula, Skanda and H{\'e}naff, Olivier J},
  journal={arXiv preprint arXiv:2402.05861},
  year={2024}
}

@inproceedings{park2023generative,
  title={Generative agents: Interactive simulacra of human behavior},
  author={Park, Joon Sung and O'Brien, Joseph and Cai, Carrie Jun and Morris, Meredith Ringel and Liang, Percy and Bernstein, Michael S},
  booktitle={Proceedings of the 36th Annual ACM Symposium on User Interface Software and Technology},
  pages={1--22},
  year={2023}
}

@misc{haarnoja2018soft,
  title={Soft actor-critic algorithms and applications},
  author={Haarnoja, Tuomas and Zhou, Aurick and Hartikainen, Kristian and Tucker, George and Ha, Sehoon and Tan, Jie and Kumar, Vikash and Zhu, Henry and Gupta, Abhishek and Abbeel, Pieter and others},
  journal={arXiv preprint arXiv:1812.05905},
  year={2018}
}

@misc{schulman2017proximal,
  title={Proximal policy optimization algorithms},
  author={Schulman, John and Wolski, Filip and Dhariwal, Prafulla and Radford, Alec and Klimov, Oleg},
  journal={arXiv preprint arXiv:1707.06347},
  year={2017}
}

@misc{jang2024all,
  title={All Neural Networks, All Autonomous, All 1X Speed},
  author={Jang, Eric},
  year={2024},
  month={Feb},
  day={8},
  howpublished={\url{https://www.1x.tech/discover/all-neural-networks-all-autonomous-all-1x-speed}},
  note={Accessed: 2024-04-10}
}

@misc{sohn2024introducing,
  author = {Sohn, Andrew and Nagabandi, Anusha and Florensa, Carlos and Adelberg, Daniel and Wu, Di and Farooq, Hassan and Clavera, Ignasi and Welborn, Jeremy and Chen, Juyue and Mishra, Nikhil and Chen, Peter and Qian, Peter and Abbeel, Pieter and Duan, Rocky and Vijay, Varun and Liu, Yang},
  title = {Introducing RFM-1: Giving robots human-like reasoning capabilities},
  year = {2024},
  month = {March},
  day = {29},
  howpublished = {\url{https://covariant.ai/insights/introducing-rfm-1-giving-robots-human-like-reasoning-capabilities/}},
  note = {Accessed: 2024-03-29}
}

@article{moore1998cramming,
  title={Cramming more components onto integrated circuits},
  author={Moore, Gordon E},
  journal={Proceedings of the IEEE},
  volume={86},
  number={1},
  pages={82--85},
  year={1998},
  publisher={Ieee}
}

@article{vaswani2017attention,
  title={Attention is all you need},
  author={Vaswani, Ashish and Shazeer, Noam and Parmar, Niki and Uszkoreit, Jakob and Jones, Llion and Gomez, Aidan N and Kaiser, {\L}ukasz and Polosukhin, Illia},
  journal={Advances in neural information processing systems},
  volume={30},
  year={2017}
}

@misc{kondratyuk2023videopoet,
  title={Videopoet: A large language model for zero-shot video generation},
  author={Kondratyuk, Dan and Yu, Lijun and Gu, Xiuye and Lezama, Jos{\'e} and Huang, Jonathan and Hornung, Rachel and Adam, Hartwig and Akbari, Hassan and Alon, Yair and Birodkar, Vighnesh and others},
  journal={arXiv preprint arXiv:2312.14125},
  year={2023}
}

@misc{hoppe2022diffusion,
  title={Diffusion models for video prediction and infilling},
  author={H{\"o}ppe, Tobias and Mehrjou, Arash and Bauer, Stefan and Nielsen, Didrik and Dittadi, Andrea},
  journal={arXiv preprint arXiv:2206.07696},
  year={2022}
}

@misc{Seo2022HARPAL,
  title={HARP: Autoregressive Latent Video Prediction with High-Fidelity Image Generator},
  author={Younggyo Seo and Kimin Lee and Fangchen Liu and Stephen James and P. Abbeel},
  journal={2022 IEEE International Conference on Image Processing (ICIP)},
  year={2022},
  pages={3943-3947},
  url={https://api.semanticscholar.org/CorpusID:252280733}
}

@inproceedings{Yu2023LanguageMB,
  title={Language Model Beats Diffusion -- Tokenizer is Key to Visual Generation},
  author={Lijun Yu and Jos{\'e} Lezama and Nitesh Bharadwaj Gundavarapu and Luca Versari and Kihyuk Sohn and David C. Minnen and Yong Cheng and Agrim Gupta and Xiuye Gu and Alexander G. Hauptmann and Boqing Gong and Ming-Hsuan Yang and Irfan Essa and David A. Ross and Lu Jiang},
  year={2023},
  url={https://api.semanticscholar.org/CorpusID:263830733}
}

@article{Villegas2022PhenakiVL,
  title={Phenaki: Variable Length Video Generation From Open Domain Textual Description},
  author={Ruben Villegas and Mohammad Babaeizadeh and Pieter-Jan Kindermans and Hernan Moraldo and Han Zhang and Mohammad Taghi Saffar and Santiago Castro and Julius Kunze and D. Erhan},
  journal={ArXiv},
  year={2022},
  volume={abs/2210.02399},
  url={https://api.semanticscholar.org/CorpusID:252715594}
}

@misc{oord2018representation,
  title={Representation learning with contrastive predictive coding},
  author={Oord, Aaron van den and Li, Yazhe and Vinyals, Oriol},
  journal={arXiv preprint arXiv:1807.03748},
  year={2018}
}

@article{Wang2023ManipulateBS,
  title={Manipulate by Seeing: Creating Manipulation Controllers from Pre-Trained Representations},
  author={Jianren Wang and Sudeep Dasari and Mohan Kumar Srirama and Shubham Tulsiani and Abhi Gupta},
  journal={ArXiv},
  year={2023},
  volume={abs/2303.08135},
  url={https://api.semanticscholar.org/CorpusID:257505038}
}

@inproceedings{sieb2020graph,
  title={Graph-structured visual imitation},
  author={Sieb, Maximilian and Xian, Zhou and Huang, Audrey and Kroemer, Oliver and Fragkiadaki, Katerina},
  booktitle={Conference on Robot Learning},
  pages={979--989},
  year={2020},
  organization={PMLR}
}

@inproceedings{schonberger2016structure,
  title={Structure-from-motion revisited},
  author={Schonberger, Johannes L and Frahm, Jan-Michael},
  booktitle={Proceedings of the IEEE conference on computer vision and pattern recognition},
  pages={4104--4113},
  year={2016}
}

@misc{wulfmeier2023foundations,
  title={Foundations for Transfer in Reinforcement Learning: A Taxonomy of Knowledge Modalities},
  author={Wulfmeier, Markus and Byravan, Arunkumar and Bechtle, Sarah and Hausman, Karol and Heess, Nicolas},
  journal={arXiv preprint arXiv:2312.01939},
  year={2023}
}

@misc{peng2020learning,
  title={Learning agile robotic locomotion skills by imitating animals},
  author={Peng, Xue Bin and Coumans, Erwin and Zhang, Tingnan and Lee, Tsang-Wei and Tan, Jie and Levine, Sergey},
  journal={arXiv preprint arXiv:2004.00784},
  year={2020}
}

@misc{ieee_solve_robotics,
  author = {Nishanth Kumar},
  title = {Will Scaling Solve Robotics?},
  year = {2024},
  url = {https://spectrum.ieee.org/solve-robotics},
  note = {Accessed: 2024-10-08}
}

@misc{li2024sds,
  title={SDS--See it, Do it, Sorted: Quadruped Skill Synthesis from Single Video Demonstration},
  author={Li, Jeffrey and Stamatopoulou, Maria and Kanoulas, Dimitrios},
  journal={arXiv preprint arXiv:2410.11571},
  year={2024}
}

@misc{huang2023voxposer,
  title={Voxposer: Composable 3d value maps for robotic manipulation with language models},
  author={Huang, Wenlong and Wang, Chen and Zhang, Ruohan and Li, Yunzhu and Wu, Jiajun and Fei-Fei, Li},
  journal={arXiv preprint arXiv:2307.05973},
  year={2023}
}

@misc{dalal2023imitating,
  title={Imitating task and motion planning with visuomotor transformers},
  author={Dalal, Murtaza and Mandlekar, Ajay and Garrett, Caelan and Handa, Ankur and Salakhutdinov, Ruslan and Fox, Dieter},
  journal={arXiv preprint arXiv:2305.16309},
  year={2023}
}

@article{yang2020multi,
  title={Multi-expert learning of adaptive legged locomotion},
  author={Yang, Chuanyu and Yuan, Kai and Zhu, Qiuguo and Yu, Wanming and Li, Zhibin},
  journal={Science Robotics},
  volume={5},
  number={49},
  pages={eabb2174},
  year={2020},
  publisher={American Association for the Advancement of Science}
}

@inproceedings{sermanet2018time,
  title={Time-contrastive networks: Self-supervised learning from video},
  author={Sermanet, Pierre and Lynch, Corey and Chebotar, Yevgen and Hsu, Jasmine and Jang, Eric and Schaal, Stefan and Levine, Sergey and Brain, Google},
  booktitle={2018 IEEE international conference on robotics and automation (ICRA)},
  pages={1134--1141},
  year={2018},
  organization={IEEE}
}

@misc{Ji2022HierarchicalRL,
  title={Hierarchical Reinforcement Learning for Precise Soccer Shooting Skills using a Quadrupedal Robot},
  author={Yandong Ji and Zhongyu Li and Yinan Sun and Xue Bin Peng and Sergey Levine and Glen Berseth and Koushil Sreenath},
  journal={2022 IEEE/RSJ International Conference on Intelligent Robots and Systems (IROS)},
  year={2022},
  pages={1479-1486},
  url={https://api.semanticscholar.org/CorpusID:251253136}
}

@misc{Lee2021AdversarialSC,
  title={Adversarial Skill Chaining for Long-Horizon Robot Manipulation via Terminal State Regularization},
  author={Youngwoon Lee and Joseph J. Lim and Anima Anandkumar and Yuke Zhu},
  booktitle={Conference on Robot Learning},
  year={2021},
  url={https://api.semanticscholar.org/CorpusID:237351177}
}

@misc{hu2023toward,
  title={Toward general-purpose robots via foundation models: A survey and meta-analysis},
  author={Hu, Yafei and Xie, Quanting and Jain, Vidhi and Francis, Jonathan and Patrikar, Jay and Keetha, Nikhil and Kim, Seungchan and Xie, Yaqi and Zhang, Tianyi and Zhao, Zhibo and others},
  journal={arXiv preprint arXiv:2312.08782},
  year={2023}
}

@misc{li2024evaluating,
  title={Evaluating Real-World Robot Manipulation Policies in Simulation},
  author={Li, Xuanlin and Hsu, Kyle and Gu, Jiayuan and Pertsch, Karl and Mees, Oier and Walke, Homer Rich and Fu, Chuyuan and Lunawat, Ishikaa and Sieh, Isabel and Kirmani, Sean and others},
  journal={arXiv preprint arXiv:2405.05941},
  year={2024}
}

@misc{prudencio2023survey,
  title={A survey on offline reinforcement learning: Taxonomy, review, and open problems},
  author={Prudencio, Rafael Figueiredo and Maximo, Marcos ROA and Colombini, Esther Luna},
  journal={IEEE Transactions on Neural Networks and Learning Systems},
  year={2023},
  publisher={IEEE}
}

@article{Schmeckpeper2020ReinforcementLW,
  title={Reinforcement Learning with Videos: Combining Offline Observations with Interaction},
  author={Karl Schmeckpeper and Oleh Rybkin and Kostas Daniilidis and Sergey Levine and Chelsea Finn},
  journal={ArXiv},
  year={2020},
  volume={abs/2011.06507},
  url={https://api.semanticscholar.org/CorpusID:226306712}
}

@misc{huang2023large,
  title={Large language models cannot self-correct reasoning yet},
  author={Huang, Jie and Chen, Xinyun and Mishra, Swaroop and Zheng, Huaixiu Steven and Yu, Adams Wei and Song, Xinying and Zhou, Denny},
  journal={arXiv preprint arXiv:2310.01798},
  year={2023}
}

@article{abel2024definition,
  title={A definition of continual reinforcement learning},
  author={Abel, David and Barreto, Andr{\'e} and Van Roy, Benjamin and Precup, Doina and van Hasselt, Hado P and Singh, Satinder},
  journal={Advances in Neural Information Processing Systems},
  volume={36},
  year={2024}
}

@misc{bommasani2021opportunities,
  title={On the opportunities and risks of foundation models},
  author={Bommasani, Rishi and Hudson, Drew A and Adeli, Ehsan and Altman, Russ and Arora, Simran and von Arx, Sydney and Bernstein, Michael S and Bohg, Jeannette and Bosselut, Antoine and Brunskill, Emma and others},
  journal={arXiv preprint arXiv:2108.07258},
  year={2021}
}

@misc{dubey2024llama,
  title={The llama 3 herd of models},
  author={Dubey, Abhimanyu and Jauhri, Abhinav and Pandey, Abhinav and Kadian, Abhishek and Al-Dahle, Ahmad and Letman, Aiesha and Mathur, Akhil and Schelten, Alan and Yang, Amy and Fan, Angela and others},
  journal={arXiv preprint arXiv:2407.21783},
  year={2024}
}

@article{Brown2020LanguageMA,
  title={Language Models are Few-Shot Learners},
  author={Tom B. Brown and Benjamin Mann and Nick Ryder and Melanie Subbiah and Jared Kaplan and Prafulla Dhariwal and Arvind Neelakantan and Pranav Shyam and Girish Sastry and Amanda Askell and Sandhini Agarwal and Ariel Herbert-Voss and Gretchen Krueger and T. J. Henighan and Rewon Child and Aditya Ramesh and Daniel M. Ziegler and Jeff Wu and Clemens Winter and Christopher Hesse and Mark Chen and Eric Sigler and Mateusz Litwin and Scott Gray and Benjamin Chess and Jack Clark and Christopher Berner and Sam McCandlish and Alec Radford and Ilya Sutskever and Dario Amodei},
  journal={ArXiv},
  year={2020},
  volume={abs/2005.14165},
  url={https://api.semanticscholar.org/CorpusID:218971783}
}

@article{OpenAI2023GPT4TR,
  title={GPT-4 Technical Report},
  author={OpenAI},
  journal={ArXiv},
  year={2023},
  volume={abs/2303.08774},
  url={https://api.semanticscholar.org/CorpusID:257532815}
}

@misc{maaz2023video,
  title={Video-chatgpt: Towards detailed video understanding via large vision and language models},
  author={Maaz, Muhammad and Rasheed, Hanoona and Khan, Salman and Khan, Fahad Shahbaz},
  journal={arXiv preprint arXiv:2306.05424},
  year={2023}
}

@article{lecun2022path,
  title={A path towards autonomous machine intelligence version 0.9. 2, 2022-06-27},
  author={LeCun, Yann},
  journal={Open Review},
  volume={62},
  number={1},
  year={2022}
}

@misc{makoviychuk2021isaac,
  title={Isaac gym: High performance gpu-based physics simulation for robot learning},
  author={Makoviychuk, Viktor and Wawrzyniak, Lukasz and Guo, Yunrong and Lu, Michelle and Storey, Kier and Macklin, Miles and Hoeller, David and Rudin, Nikita and Allshire, Arthur and Handa, Ankur and others},
  journal={arXiv preprint arXiv:2108.10470},
  year={2021}
}

@inproceedings{todorov2012mujoco,
  title={Mujoco: A physics engine for model-based control},
  author={Todorov, Emanuel and Erez, Tom and Tassa, Yuval},
  booktitle={2012 IEEE/RSJ international conference on intelligent robots and systems},
  pages={5026--5033},
  year={2012},
  organization={IEEE}
}

@misc{Rombach2021HighResolutionIS,
  title={High-Resolution Image Synthesis with Latent Diffusion Models},
  author={Robin Rombach and A. Blattmann and Dominik Lorenz and Patrick Esser and Bj{\"o}rn Ommer},
  journal={2022 IEEE/CVF Conference on Computer Vision and Pattern Recognition (CVPR)},
  year={2021},
  pages={10674-10685},
  url={https://api.semanticscholar.org/CorpusID:245335280}
}

@misc{Grauman2021Ego4DAT,
  title={Ego4D: Around the World in 3,000 Hours of Egocentric Video},
  author={Kristen Grauman and Andrew Westbury and Eugene Byrne and Zachary Q. Chavis and Antonino Furnari and Rohit Girdhar and Jackson Hamburger and Hao Jiang and Miao Liu and Xingyu Liu and Miguel Martin and Tushar Nagarajan and Ilija Radosavovic and Santhosh K. Ramakrishnan and Fiona Ryan and Jayant Sharma et. al.},
  journal={2022 IEEE/CVF Conference on Computer Vision and Pattern Recognition (CVPR)},
  year={2021},
  pages={18973-18990},
  url={https://api.semanticscholar.org/CorpusID:238856888}
}

@inproceedings{bain2021frozen,
  title={Frozen in time: A joint video and image encoder for end-to-end retrieval},
  author={Bain, Max and Nagrani, Arsha and Varol, G{\"u}l and Zisserman, Andrew},
  booktitle={Proceedings of the IEEE/CVF International Conference on Computer Vision},
  pages={1728--1738},
  year={2021}
}

@misc{Goyal2017TheS,
  title={The “Something Something” Video Database for Learning and Evaluating Visual Common Sense},
  author={Raghav Goyal and Samira Ebrahimi Kahou and Vincent Michalski and Joanna Materzynska and Susanne Westphal and Heuna Kim and Valentin Haenel and Ingo Fr{\"u}nd and Peter N. Yianilos and Moritz Mueller-Freitag and Florian Hoppe and Christian Thurau and Ingo Bax and Roland Memisevic},
  journal={2017 IEEE International Conference on Computer Vision (ICCV)},
  year={2017},
  pages={5843-5851},
  url={https://api.semanticscholar.org/CorpusID:834612}
}

@inproceedings{nagrani2022learning,
  title={Learning audio-video modalities from image captions},
  author={Nagrani, Arsha and Seo, Paul Hongsuck and Seybold, Bryan and Hauth, Anja and Manen, Santiago and Sun, Chen and Schmid, Cordelia},
  booktitle={European Conference on Computer Vision},
  pages={407--426},
  year={2022},
  organization={Springer}
}

@misc{grauman2023ego,
  title={Ego-exo4d: Understanding skilled human activity from first-and third-person perspectives},
  author={Grauman, Kristen and Westbury, Andrew and Torresani, Lorenzo and Kitani, Kris and Malik, Jitendra and Afouras, Triantafyllos and Ashutosh, Kumar and Baiyya, Vijay and Bansal, Siddhant and Boote, Bikram and others},
  journal={arXiv preprint arXiv:2311.18259},
  year={2023}
}

@misc{Miech2019HowTo100MLA,
  title={HowTo100M: Learning a Text-Video Embedding by Watching Hundred Million Narrated Video Clips},
  author={Antoine Miech and Dimitri Zhukov and Jean-Baptiste Alayrac and Makarand Tapaswi and Ivan Laptev and Josef Sivic},
  journal={2019 IEEE/CVF International Conference on Computer Vision (ICCV)},
  year={2019},
  pages={2630-2640},
  url={https://api.semanticscholar.org/CorpusID:182952863}
}

@misc{Shan2020UnderstandingHH,
  title={Understanding Human Hands in Contact at Internet Scale},
  author={Dandan Shan and Jiaqi Geng and Michelle Shu and David F. Fouhey},
  journal={2020 IEEE/CVF Conference on Computer Vision and Pattern Recognition (CVPR)},
  year={2020},
  pages={9866-9875},
  url={https://api.semanticscholar.org/CorpusID:215413188}
}

@inproceedings{ding2023mose,
  title={MOSE: A new dataset for video object segmentation in complex scenes},
  author={Ding, Henghui and Liu, Chang and He, Shuting and Jiang, Xudong and Torr, Philip HS and Bai, Song},
  booktitle={Proceedings of the IEEE/CVF International Conference on Computer Vision},
  pages={20224--20234},
  year={2023}
}

@article{Zhu2023GuidingOR,
  title={Guiding Online Reinforcement Learning with Action-Free Offline Pretraining},
  author={Deyao Zhu and Yuhui Wang and J{\"u}rgen Schmidhuber and Mohamed Elhoseiny},
  journal={ArXiv},
  year={2023},
  volume={abs/2301.12876},
  url={https://api.semanticscholar.org/CorpusID:256390557}
}

@inproceedings{kirillov2023segment,
  title={Segment anything},
  author={Kirillov, Alexander and Mintun, Eric and Ravi, Nikhila and Mao, Hanzi and Rolland, Chloe and Gustafson, Laura and Xiao, Tete and Whitehead, Spencer and Berg, Alexander C and Lo, Wan-Yen and others},
  booktitle={Proceedings of the IEEE/CVF International Conference on Computer Vision},
  pages={4015--4026},
  year={2023}
}

@misc{klissarov2023motif,
  title={Motif: Intrinsic motivation from artificial intelligence feedback},
  author={Klissarov, Martin and D'Oro, Pierluca and Sodhani, Shagun and Raileanu, Roberta and Bacon, Pierre-Luc and Vincent, Pascal and Zhang, Amy and Henaff, Mikael},
  journal={arXiv preprint arXiv:2310.00166},
  year={2023}
}

@misc{yang2023robot,
  title={Robot Fine-Tuning Made Easy: Pre-Training Rewards and Policies for Autonomous Real-World Reinforcement Learning},
  author={Yang, Jingyun and Mark, Max Sobol and Vu, Brandon and Sharma, Archit and Bohg, Jeannette and Finn, Chelsea},
  journal={arXiv preprint arXiv:2310.15145},
  year={2023}
}

@misc{baumli2023vision,
  title={Vision-language models as a source of rewards},
  author={Baumli, Kate and Baveja, Satinder and Behbahani, Feryal and Chan, Harris and Comanici, Gheorghe and Flennerhag, Sebastian and Gazeau, Maxime and Holsheimer, Kristian and Horgan, Dan and Laskin, Michael and others},
  journal={arXiv preprint arXiv:2312.09187},
  year={2023}
}

@misc{du2023vision,
  title={Vision-language models as success detectors},
  author={Du, Yuqing and Konyushkova, Ksenia and Denil, Misha and Raju, Akhil and Landon, Jessica and Hill, Felix and de Freitas, Nando and Cabi, Serkan},
  journal={arXiv preprint arXiv:2303.07280},
  year={2023}
}

@misc{huang2023diffusion,
  title={Diffusion Reward: Learning Rewards via Conditional Video Diffusion},
  author={Huang, Tao and Jiang, Guangqi and Ze, Yanjie and Xu, Huazhe},
  journal={arXiv preprint arXiv:2312.14134},
  year={2023}
}

@article{Du2023LearningUP,
  title={Learning Universal Policies via Text-Guided Video Generation},
  author={Yilun Du and Mengjiao Yang and Bo Dai and Hanjun Dai and Ofir Nachum and Joshua B. Tenenbaum and Dale Schuurmans and P. Abbeel},
  journal={ArXiv},
  year={2023},
  volume={abs/2302.00111},
  url={https://api.semanticscholar.org/CorpusID:256459809}
}

@article{Escontrela2023VideoPM,
  title={Video Prediction Models as Rewards for Reinforcement Learning},
  author={Alejandro Escontrela and Ademi Adeniji and Wilson Yan and Ajay Jain and Xue Bin Peng and Ken Goldberg and Youngwoon Lee and Danijar Hafner and P. Abbeel},
  journal={ArXiv},
  year={2023},
  volume={abs/2305.14343},
  url={https://api.semanticscholar.org/CorpusID:258841355}
}

@article{Reed2022AGA,
  title={A Generalist Agent},
  author={Scott Reed and Konrad Zolna and Emilio Parisotto and Sergio Gomez Colmenarejo and Alexander Novikov and Gabriel Barth-Maron and Mai Gimenez and Yury Sulsky and Jackie Kay and Jost Tobias Springenberg and Tom Eccles and Jake Bruce and Ali Razavi and Ashley D. Edwards and Nicolas Manfred Otto Heess and Yutian Chen and Raia Hadsell and Oriol Vinyals and Mahyar Bordbar and Nando de Freitas},
  journal={Trans. Mach. Learn. Res.},
  year={2022},
  volume={2022},
  url={https://api.semanticscholar.org/CorpusID:248722148}
}

@article{betker2023improving,
  title={Improving image generation with better captions},
  author={Betker, James and Goh, Gabriel and Jing, Li and Brooks, Tim and Wang, Jianfeng and Li, Linjie and Ouyang, Long and Zhuang, Juntang and Lee, Joyce and Guo, Yufei and others},
  journal={Computer Science. https://cdn. openai. com/papers/dall-e-3. pdf},
  volume={2},
  number={3},
  pages={8},
  year={2023}
}

@misc{yu2023scaling,
  title={Scaling robot learning with semantically imagined experience},
  author={Yu, Tianhe and Xiao, Ted and Stone, Austin and Tompson, Jonathan and Brohan, Anthony and Wang, Su and Singh, Jaspiar and Tan, Clayton and Peralta, Jodilyn and Ichter, Brian and others},
  journal={arXiv preprint arXiv:2302.11550},
  year={2023}
}

@article{JyothirS2023GradientbasedPW,
  title={Gradient-based Planning with World Models},
  author={V JyothirS and Siddhartha Jalagam and Yann LeCun and Vlad Sobal},
  journal={ArXiv},
  year={2023},
  volume={abs/2312.17227},
  url={https://api.semanticscholar.org/CorpusID:266573170}
}

@misc{edwards2019perceptual,
  title={Perceptual values from observation},
  author={Edwards, Ashley D and Isbell, Charles L},
  journal={arXiv preprint arXiv:1905.07861},
  year={2019}
}

@misc{lin2023video,
  title={Video-llava: Learning united visual representation by alignment before projection},
  author={Lin, Bin and Zhu, Bin and Ye, Yang and Ning, Munan and Jin, Peng and Yuan, Li},
  journal={arXiv preprint arXiv:2311.10122},
  year={2023}
}

@misc{yu2023language,
  title={Language to rewards for robotic skill synthesis},
  author={Yu, Wenhao and Gileadi, Nimrod and Fu, Chuyuan and Kirmani, Sean and Lee, Kuang-Huei and Arenas, Montse Gonzalez and Chiang, Hao-Tien Lewis and Erez, Tom and Hasenclever, Leonard and Humplik, Jan and others},
  journal={arXiv preprint arXiv:2306.08647},
  year={2023}
}

@misc{xu2024manifoundation,
  title={ManiFoundation Model for General-Purpose Robotic Manipulation of Contact Synthesis with Arbitrary Objects and Robots},
  author={Xu, Zhixuan and Gao, Chongkai and Liu, Zixuan and Yang, Gang and Tie, Chenrui and Zheng, Haozhuo and Zhou, Haoyu and Peng, Weikun and Wang, Debang and Chen, Tianyi and others},
  journal={arXiv preprint arXiv:2405.06964},
  year={2024}
}

@misc{chisari2024learning,
  title={Learning Robotic Manipulation Policies from Point Clouds with Conditional Flow Matching},
  author={Chisari, Eugenio and Heppert, Nick and Argus, Max and Welschehold, Tim and Brox, Thomas and Valada, Abhinav},
  journal={arXiv preprint arXiv:2409.07343},
  year={2024}
}

@misc{levy2024learning,
  title={Learning to Walk from Three Minutes of Real-World Data with Semi-structured Dynamics Models},
  author={Levy, Jacob and Westenbroek, Tyler and Fridovich-Keil, David},
  journal={arXiv preprint arXiv:2410.09163},
  year={2024}
}

@misc{lutter2019deep,
  title={Deep lagrangian networks: Using physics as model prior for deep learning},
  author={Lutter, Michael and Ritter, Christian and Peters, Jan},
  journal={arXiv preprint arXiv:1907.04490},
  year={2019}
}

@misc{Wang2018NerveNetLS,
  title={NerveNet: Learning Structured Policy with Graph Neural Networks},
  author={Tingwu Wang and Renjie Liao and Jimmy Ba and Sanja Fidler},
  booktitle={International Conference on Learning Representations},
  year={2018},
  url={https://api.semanticscholar.org/CorpusID:65051725}
}

@misc{sferrazza2024body,
  title={Body Transformer: Leveraging Robot Embodiment for Policy Learning},
  author={Sferrazza, Carmelo and Huang, Dun-Ming and Liu, Fangchen and Lee, Jongmin and Abbeel, Pieter},
  journal={arXiv preprint arXiv:2408.06316},
  year={2024}
}

@misc{Marza2022MultiObjectNW,
  title={Multi-Object Navigation with dynamically learned neural implicit representations},
  author={Pierre Marza and La{\"e}titia Matignon and Olivier Simonin and Christian Wolf},
  journal={2023 IEEE/CVF International Conference on Computer Vision (ICCV)},
  year={2022},
  pages={10970-10981},
  url={https://api.semanticscholar.org/CorpusID:252815543}
}

@misc{Chaplot2020NeuralTS,
  title={Neural Topological SLAM for Visual Navigation},
  author={Devendra Singh Chaplot and Ruslan Salakhutdinov and Abhinav Kumar Gupta and Saurabh Gupta},
  journal={2020 IEEE/CVF Conference on Computer Vision and Pattern Recognition (CVPR)},
  year={2020},
  pages={12872-12881},
  url={https://api.semanticscholar.org/CorpusID:214754592}
}

@misc{peng2024q,
  title={Q-slam: Quadric representations for monocular slam},
  author={Peng, Chensheng and Xu, Chenfeng and Wang, Yue and Ding, Mingyu and Yang, Heng and Tomizuka, Masayoshi and Keutzer, Kurt and Pavone, Marco and Zhan, Wei},
  journal={arXiv preprint arXiv:2403.08125},
  year={2024}
}

@misc{liang2024environment,
  title={Environment Curriculum Generation via Large Language Models},
  author={Liang, William and Wang, Sam and Wang, Hung-Ju and Bastani, Osbert and Jayaraman, Dinesh and Ma, Yecheng Jason},
  booktitle={8th Annual Conference on Robot Learning},
  year={2024}
}

@misc{mishra2023generative,
  title={Generative skill chaining: Long-horizon skill planning with diffusion models},
  author={Mishra, Utkarsh Aashu and Xue, Shangjie and Chen, Yongxin and Xu, Danfei},
  booktitle={Conference on Robot Learning},
  pages={2905--2925},
  year={2023},
  organization={PMLR}
}

@misc{kaufmann2023survey,
  title={A survey of reinforcement learning from human feedback},
  author={Kaufmann, Timo and Weng, Paul and Bengs, Viktor and H{\"u}llermeier, Eyke},
  journal={arXiv preprint arXiv:2312.14925},
  year={2023}
}

@article{yuan2023hierarchical,
  title={Hierarchical generative modelling for autonomous robots},
  author={Yuan, Kai and Sajid, Noor and Friston, Karl and Li, Zhibin},
  journal={Nature Machine Intelligence},
  volume={5},
  number={12},
  pages={1402--1414},
  year={2023},
  publisher={Nature Publishing Group UK London}
}

@misc{liu2023ring,
  title={Ring attention with blockwise transformers for near-infinite context},
  author={Liu, Hao and Zaharia, Matei and Abbeel, Pieter},
  journal={arXiv preprint arXiv:2310.01889},
  year={2023}
}

@article{Layne2023RobotrelatedFA,
  title={Robot-related fatalities at work in the United States, 1992-2017.},
  author={Larry A Layne},
  journal={American journal of industrial medicine},
  year={2023},
  url={https://api.semanticscholar.org/CorpusID:257370737}
}

@inproceedings{li2024visual,
  title={Visual Robotic Manipulation with Depth-Aware Pretraining},
  author={Li, Jinming and Wang, Wanying and Peng, Yaxin and Shen, Chaomin and Zhu, Yichen and Xu, Zhiyuan},
  booktitle={2024 IEEE International Conference on Robotics and Biomimetics (ROBIO)},
  pages={843--850},
  year={2024},
  organization={IEEE}
}

@article{zhao2023large,
  title={Large language models as commonsense knowledge for large-scale task planning},
  author={Zhao, Zirui and Lee, Wee Sun and Hsu, David},
  journal={Advances in Neural Information Processing Systems},
  volume={36},
  pages={31967--31987},
  year={2023}
}

@misc{Liu2024WorldMO,
  title={World Model on Million-Length Video And Language With Blockwise RingAttention},
  author={Hao Liu and Wilson Yan and Matei Zaharia and Pieter Abbeel},
  year={2024},
  url={https://api.semanticscholar.org/CorpusID:268385142}
}

@misc{jiang2024mixtral,
  title={Mixtral of experts},
  author={Jiang, Albert Q and Sablayrolles, Alexandre and Roux, Antoine and Mensch, Arthur and Savary, Blanche and Bamford, Chris and Chaplot, Devendra Singh and Casas, Diego de las and Hanna, Emma Bou and Bressand, Florian and others},
  journal={arXiv preprint arXiv:2401.04088},
  year={2024}
}

@misc{song2024germ,
  title={GeRM: A Generalist Robotic Model with Mixture-of-experts for Quadruped Robot},
  author={Song, Wenxuan and Zhao, Han and Ding, Pengxiang and Cui, Can and Lyu, Shangke and Fan, Yaning and Wang, Donglin},
  journal={arXiv preprint arXiv:2403.13358},
  year={2024}
}

@misc{zhou2024robodreamer,
  title={RoboDreamer: Learning Compositional World Models for Robot Imagination},
  author={Zhou, Siyuan and Du, Yilun and Chen, Jiaben and Li, Yandong and Yeung, Dit-Yan and Gan, Chuang},
  journal={arXiv preprint arXiv:2404.12377},
  year={2024}
}

@misc{anwar2024foundational,
  title={Foundational challenges in assuring alignment and safety of large language models},
  author={Anwar, Usman and Saparov, Abulhair and Rando, Javier and Paleka, Daniel and Turpin, Miles and Hase, Peter and Lubana, Ekdeep Singh and Jenner, Erik and Casper, Stephen and Sourbut, Oliver and others},
  journal={arXiv preprint arXiv:2404.09932},
  year={2024}
}

@misc{radosavovic2024humanoid,
  title={Humanoid Locomotion as Next Token Prediction},
  author={Radosavovic, Ilija and Zhang, Bike and Shi, Baifeng and Rajasegaran, Jathushan and Kamat, Sarthak and Darrell, Trevor and Sreenath, Koushil and Malik, Jitendra},
  journal={arXiv preprint arXiv:2402.19469},
  year={2024}
}

@misc{li2024behavior,
  title={Behavior-1k: A human-centered, embodied ai benchmark with 1,000 everyday activities and realistic simulation},
  author={Li, Chengshu and Zhang, Ruohan and Wong, Josiah and Gokmen, Cem and Srivastava, Sanjana and Mart{\'\i}n-Mart{\'\i}n, Roberto and Wang, Chen and Levine, Gabrael and Ai, Wensi and Martinez, Benjamin and others},
  journal={arXiv preprint arXiv:2403.09227},
  year={2024}
}

@misc{pumacay2024colosseum,
  title={THE COLOSSEUM: A Benchmark for Evaluating Generalization for Robotic Manipulation},
  author={Pumacay, Wilbert and Singh, Ishika and Duan, Jiafei and Krishna, Ranjay and Thomason, Jesse and Fox, Dieter},
  journal={arXiv preprint arXiv:2402.08191},
  year={2024}
}

@inproceedings{blank2024scaling,
  title={Scaling Robot Policy Learning via Zero-Shot Labeling with Foundation Models},
  author={Blank, Nils and Reuss, Moritz and Wenzel, Fabian and Mees, Oier and Lioutikov, Rudolf},
  booktitle={2nd Workshop on Mobile Manipulation and Embodied Intelligence at ICRA 2024},
    year={2024}
}

@misc{faldor2024omni,
  title={OMNI-EPIC: Open-endedness via Models of human Notions of Interestingness with Environments Programmed in Code},
  author={Faldor, Maxence and Zhang, Jenny and Cully, Antoine and Clune, Jeff},
  journal={arXiv preprint arXiv:2405.15568},
  year={2024}
}

@misc{ranasinghe2024understanding,
  title={Understanding Long Videos in One Multimodal Language Model Pass},
  author={Ranasinghe, Kanchana and Li, Xiang and Kahatapitiya, Kumara and Ryoo, Michael S},
  journal={arXiv preprint arXiv:2403.16998},
  year={2024}
}

@misc{shang2024traveler,
  title={TraveLER: A Multi-LMM Agent Framework for Video Question-Answering},
  author={Shang, Chuyi and You, Amos and Subramanian, Sanjay and Darrell, Trevor and Herzig, Roei},
  journal={arXiv preprint arXiv:2404.01476},
  year={2024}
}

@misc{liu2024enhancing,
  title={Enhancing Robotic Manipulation with AI Feedback from Multimodal Large Language Models},
  author={Liu, Jinyi and Yuan, Yifu and Hao, Jianye and Ni, Fei and Fu, Lingzhi and Chen, Yibin and Zheng, Yan},
  journal={arXiv preprint arXiv:2402.14245},
  year={2024}
}

@misc{he2024large,
  title={Large-Scale Actionless Video Pre-Training via Discrete Diffusion for Efficient Policy Learning},
  author={He, Haoran and Bai, Chenjia and Pan, Ling and Zhang, Weinan and Zhao, Bin and Li, Xuelong},
  journal={arXiv preprint arXiv:2402.14407},
  year={2024}
}

@misc{shi2024yell,
  title={Yell At Your Robot: Improving On-the-Fly from Language Corrections},
  author={Shi, Lucy Xiaoyang and Hu, Zheyuan and Zhao, Tony Z and Sharma, Archit and Pertsch, Karl and Luo, Jianlan and Levine, Sergey and Finn, Chelsea},
  journal={arXiv preprint arXiv:2403.12910},
  year={2024}
}

@misc{jain2024vid2robot,
  title={Vid2Robot: End-to-end Video-conditioned Policy Learning with Cross-Attention Transformers},
  author={Jain, Vidhi and Attarian, Maria and Joshi, Nikhil J and Wahid, Ayzaan and Driess, Danny and Vuong, Quan and Sanketi, Pannag R and Sermanet, Pierre and Welker, Stefan and Chan, Christine and others},
  journal={arXiv preprint arXiv:2403.12943},
  year={2024}
}

@misc{belkhale2024rt,
  title={Rt-h: Action hierarchies using language},
  author={Belkhale, Suneel and Ding, Tianli and Xiao, Ted and Sermanet, Pierre and Vuong, Quon and Tompson, Jonathan and Chebotar, Yevgen and Dwibedi, Debidatta and Sadigh, Dorsa},
  journal={arXiv preprint arXiv:2403.01823},
  year={2024}
}

@inproceedings{lin2024spawnnet,
  title={SpawnNet: Learning Generalizable Visuomotor Skills from Pre-trained Network},
  author={Lin, Xingyu and So, John and Mahalingam, Sashwat and Liu, Fangchen and Abbeel, Pieter},
  booktitle={2024 IEEE International Conference on Robotics and Automation (ICRA)},
  pages={4781--4787},
  year={2024},
  organization={IEEE}
}

@misc{xiang2024pandora,
  title={Pandora: Towards General World Model with Natural Language Actions and Video States},
  author={Xiang, Jiannan and Liu, Guangyi and Gu, Yi and Gao, Qiyue and Ning, Yuting and Zha, Yuheng and Feng, Zeyu and Tao, Tianhua and Hao, Shibo and Shi, Yemin and others},
  journal={arXiv preprint arXiv:2406.09455},
  year={2024}
}

@misc{li2024ag2manip,
  title={Ag2Manip: Learning Novel Manipulation Skills with Agent-Agnostic Visual and Action Representations},
  author={Li, Puhao and Liu, Tengyu and Li, Yuyang and Han, Muzhi and Geng, Haoran and Wang, Shu and Zhu, Yixin and Zhu, Song-Chun and Huang, Siyuan},
  journal={arXiv preprint arXiv:2404.17521},
  year={2024}
}

@misc{wang2024internvideo2,
  title={Internvideo2: Scaling video foundation models for multimodal video understanding},
  author={Wang, Yi and Li, Kunchang and Li, Xinhao and Yu, Jiashuo and He, Yinan and Chen, Guo and Pei, Baoqi and Zheng, Rongkun and Xu, Jilan and Wang, Zun and others},
  journal={arXiv preprint arXiv:2403.15377},
  year={2024}
}

@misc{wang2023robogen,
  title={Robogen: Towards unleashing infinite data for automated robot learning via generative simulation},
  author={Wang, Yufei and Xian, Zhou and Chen, Feng and Wang, Tsun-Hsuan and Wang, Yian and Fragkiadaki, Katerina and Erickson, Zackory and Held, David and Gan, Chuang},
  journal={arXiv preprint arXiv:2311.01455},
  year={2023}
}

@misc{jang2024visual,
  title={Visual Representation Learning with Stochastic Frame Prediction},
  author={Jang, Huiwon and Kim, Dongyoung and Kim, Junsu and Shin, Jinwoo and Abbeel, Pieter and Seo, Younggyo},
  journal={arXiv preprint arXiv:2406.07398},
  year={2024}
}

@misc{nasiriany2024robocasa,
  title={RoboCasa: Large-Scale Simulation of Everyday Tasks for Generalist Robots},
  author={Nasiriany, Soroush and Maddukuri, Abhiram and Zhang, Lance and Parikh, Adeet and Lo, Aaron and Joshi, Abhishek and Mandlekar, Ajay and Zhu, Yuke},
  journal={arXiv preprint arXiv:2406.02523},
  year={2024}
}

@misc{zhen20243d,
  title={3D-VLA: A 3D Vision-Language-Action Generative World Model},
  author={Zhen, Haoyu and Qiu, Xiaowen and Chen, Peihao and Yang, Jincheng and Yan, Xin and Du, Yilun and Hong, Yining and Gan, Chuang},
  journal={arXiv preprint arXiv:2403.09631},
  year={2024}
}

@misc{bousmalis2023robocat,
  title={Robocat: A self-improving foundation agent for robotic manipulation},
  author={Bousmalis, Konstantinos and Vezzani, Giulia and Rao, Dushyant and Devin, Coline and Lee, Alex X and Bauza, Maria and Davchev, Todor and Zhou, Yuxiang and Gupta, Agrim and Raju, Akhil and others},
  journal={arXiv preprint arXiv:2306.11706},
  year={2023}
}

@misc{ahn2024autort,
  title={Autort: Embodied foundation models for large scale orchestration of robotic agents},
  author={Ahn, Michael and Dwibedi, Debidatta and Finn, Chelsea and Arenas, Montse Gonzalez and Gopalakrishnan, Keerthana and Hausman, Karol and Ichter, Brian and Irpan, Alex and Joshi, Nikhil and Julian, Ryan and others},
  journal={arXiv preprint arXiv:2401.12963},
  year={2024}
}

@misc{team2023human,
  title={Human-timescale adaptation in an open-ended task space},
  author={Team, Adaptive Agent and Bauer, Jakob and Baumli, Kate and Baveja, Satinder and Behbahani, Feryal and Bhoopchand, Avishkar and Bradley-Schmieg, Nathalie and Chang, Michael and Clay, Natalie and Collister, Adrian and others},
  journal={arXiv preprint arXiv:2301.07608},
  year={2023}
}

@article{Sontakke2023RoboCLIPOD,
  title={RoboCLIP: One Demonstration is Enough to Learn Robot Policies},
  author={Sumedh Anand Sontakke and Jesse Zhang and S'ebastien M. R. Arnold and Karl Pertsch and Erdem Biyik and Dorsa Sadigh and Chelsea Finn and Laurent Itti},
  journal={ArXiv},
  year={2023},
  volume={abs/2310.07899},
  url={https://api.semanticscholar.org/CorpusID:263909538}
}

@inproceedings{Aytar2018PlayingHE,
  title={Playing hard exploration games by watching YouTube},
  author={Yusuf Aytar and Tobias Pfaff and David Budden and Tom Le Paine and Ziyun Wang and Nando de Freitas},
  booktitle={Neural Information Processing Systems},
  year={2018},
  url={https://api.semanticscholar.org/CorpusID:44061126}
}

@inproceedings{Qin2021DexMVIL,
  title={DexMV: Imitation Learning for Dexterous Manipulation from Human Videos},
  author={Yuzhe Qin and Yueh-Hua Wu and Shaowei Liu and Hanwen Jiang and Ruihan Yang and Yang Fu and Xiaolong Wang},
  booktitle={European Conference on Computer Vision},
  year={2021},
  url={https://api.semanticscholar.org/CorpusID:236986915}
}

@article{Chen2021LearningGR,
  title={Learning Generalizable Robotic Reward Functions from "In-The-Wild" Human Videos},
  author={Annie S. Chen and Suraj Nair and Chelsea Finn},
  journal={ArXiv},
  year={2021},
  volume={abs/2103.16817},
  url={https://api.semanticscholar.org/CorpusID:232428118}
}

@article{Mandikal2022DexVIPLD,
  title={DexVIP: Learning Dexterous Grasping with Human Hand Pose Priors from Video},
  author={Priyanka Mandikal and Kristen Grauman},
  journal={ArXiv},
  year={2022},
  volume={abs/2202.00164},
  url={https://api.semanticscholar.org/CorpusID:237369373}
}

@article{Bahl2022HumantoRobotII,
  title={Human-to-Robot Imitation in the Wild},
  author={Shikhar Bahl and Abhi Gupta and Deepak Pathak},
  journal={ArXiv},
  year={2022},
  volume={abs/2207.09450},
  url={https://api.semanticscholar.org/CorpusID:248941578}
}

@article{Shao2020Concept2RobotLM,
  title={Concept2Robot: Learning manipulation concepts from instructions and human demonstrations},
  author={Lin Shao and Toki Migimatsu and Qiang Zhang and Karen Yang and Jeannette Bohg},
  journal={The International Journal of Robotics Research},
  year={2020},
  volume={40},
  pages={1419 - 1434},
  url={https://api.semanticscholar.org/CorpusID:220069237}
}

@article{Torabi2018GenerativeAI,
  title={Generative Adversarial Imitation from Observation},
  author={Faraz Torabi and Garrett Warnell and Peter Stone},
  journal={ArXiv},
  year={2018},
  volume={abs/1807.06158},
  url={https://api.semanticscholar.org/CorpusID:49863329}
}

@article{kuttler2020nethack,
  title={The nethack learning environment},
  author={K{\"u}ttler, Heinrich and Nardelli, Nantas and Miller, Alexander and Raileanu, Roberta and Selvatici, Marco and Grefenstette, Edward and Rockt{\"a}schel, Tim},
  journal={Advances in Neural Information Processing Systems},
  volume={33},
  pages={7671--7684},
  year={2020}
}

@article{fan2022minedojo,
  title={Minedojo: Building open-ended embodied agents with internet-scale knowledge},
  author={Fan, Linxi and Wang, Guanzhi and Jiang, Yunfan and Mandlekar, Ajay and Yang, Yuncong and Zhu, Haoyi and Tang, Andrew and Huang, De-An and Zhu, Yuke and Anandkumar, Anima},
  journal={Advances in Neural Information Processing Systems},
  volume={35},
  pages={18343--18362},
  year={2022}
}

@inproceedings{deng2009imagenet,
  title={Imagenet: A large-scale hierarchical image database},
  author={Deng, Jia and Dong, Wei and Socher, Richard and Li, Li-Jia and Li, Kai and Fei-Fei, Li},
  booktitle={2009 IEEE conference on computer vision and pattern recognition},
  pages={248--255},
  year={2009},
  organization={Ieee}
}

@article{Ding2023CLIP4MCAR,
  title={CLIP4MC: An RL-Friendly Vision-Language Model for Minecraft},
  author={Ziluo Ding and Hao Luo and Ke Li and Junpeng Yue and Tiejun Huang and Zongqing Lu},
  journal={ArXiv},
  year={2023},
  volume={abs/2303.10571},
  url={https://api.semanticscholar.org/CorpusID:257632482}
}

@article{Dasari2023AnUL,
  title={An Unbiased Look at Datasets for Visuo-Motor Pre-Training},
  author={Sudeep Dasari and Mohan Kumar Srirama and Unnat Jain and Abhinav Gupta},
  journal={ArXiv},
  year={2023},
  volume={abs/2310.09289},
  url={https://api.semanticscholar.org/CorpusID:263914653}
}

@inproceedings{Ma2023LIVLR,
  title={LIV: Language-Image Representations and Rewards for Robotic Control},
  author={Yecheng Jason Ma and William Jiahua Liang and Vaidehi Som and Vikash Kumar and Amy Zhang and Osbert Bastani and Dinesh Jayaraman},
  booktitle={International Conference on Machine Learning},
  year={2023},
  url={https://api.semanticscholar.org/CorpusID:258999195}
}

@misc{Bahl2023AffordancesFH,
  title={Affordances from Human Videos as a Versatile Representation for Robotics},
  author={Shikhar Bahl and Russell Mendonca and Lili Chen and Unnat Jain and Deepak Pathak},
  journal={2023 IEEE/CVF Conference on Computer Vision and Pattern Recognition (CVPR)},
  year={2023},
  pages={01-13},
  url={https://api.semanticscholar.org/CorpusID:258180471}
}

@article{ho2016generative,
  title={Generative adversarial imitation learning},
  author={Ho, Jonathan and Ermon, Stefano},
  journal={Advances in neural information processing systems},
  volume={29},
  year={2016}
}

@inproceedings{wang2023optimal,
  title={Optimal goal-reaching reinforcement learning via quasimetric learning},
  author={Wang, Tongzhou and Torralba, Antonio and Isola, Phillip and Zhang, Amy},
  booktitle={International Conference on Machine Learning},
  pages={36411--36430},
  year={2023},
  organization={PMLR}
}

@article{Ma2022VIPTU,
  title={VIP: Towards Universal Visual Reward and Representation via Value-Implicit Pre-Training},
  author={Yecheng Jason Ma and Shagun Sodhani and Dinesh Jayaraman and Osbert Bastani and Vikash Kumar and Amy Zhang},
  journal={ArXiv},
  year={2022},
  volume={abs/2210.00030},
  url={https://api.semanticscholar.org/CorpusID:252683397}
}

@inproceedings{Zakka2021XIRLCI,
  title={XIRL: Cross-embodiment Inverse Reinforcement Learning},
  author={Kevin Zakka and Andy Zeng and Peter R. Florence and Jonathan Tompson and Jeannette Bohg and Debidatta Dwibedi},
  booktitle={Conference on Robot Learning},
  year={2021},
  url={https://api.semanticscholar.org/CorpusID:235368061}
}

@misc{Chen2021AnES,
  title={An Empirical Study of Training Self-Supervised Vision Transformers},
  author={Xinlei Chen and Saining Xie and Kaiming He},
  journal={2021 IEEE/CVF International Conference on Computer Vision (ICCV)},
  year={2021},
  pages={9620-9629},
  url={https://api.semanticscholar.org/CorpusID:233024948}
}

@inproceedings{liu2023learning,
  title={Learning to identify critical states for reinforcement learning from videos},
  author={Liu, Haozhe and Zhuge, Mingchen and Li, Bing and Wang, Yuhui and Faccio, Francesco and Ghanem, Bernard and Schmidhuber, J{\"u}rgen},
  booktitle={Proceedings of the IEEE/CVF International Conference on Computer Vision},
  pages={1955--1965},
  year={2023}
}

@inproceedings{Parisi2022TheUE,
  title={The Unsurprising Effectiveness of Pre-Trained Vision Models for Control},
  author={Simone Parisi and Aravind Rajeswaran and Senthil Purushwalkam and Abhinav Kumar Gupta},
  booktitle={International Conference on Machine Learning},
  year={2022},
  url={https://api.semanticscholar.org/CorpusID:247292805}
}

@inproceedings{Nair2022R3MAU,
  title={R3M: A Universal Visual Representation for Robot Manipulation},
  author={Suraj Nair and Aravind Rajeswaran and Vikash Kumar and Chelsea Finn and Abhi Gupta},
  booktitle={Conference on Robot Learning},
  year={2022},
  url={https://api.semanticscholar.org/CorpusID:247618840}
}

@misc{Yu2022MAGVITMG,
  title={MAGVIT: Masked Generative Video Transformer},
  author={Lijun Yu and Yong Cheng and Kihyuk Sohn and Jos{\'e} Lezama and Han Zhang and Huiwen Chang and Alexander G. Hauptmann and Ming-Hsuan Yang and Yuan Hao and Irfan Essa and Lu Jiang},
  journal={2023 IEEE/CVF Conference on Computer Vision and Pattern Recognition (CVPR)},
  year={2022},
  pages={10459-10469},
  url={https://api.semanticscholar.org/CorpusID:254563906}
}

@article{Karamcheti2023LanguageDrivenRL,
  title={Language-Driven Representation Learning for Robotics},
  author={Siddharth Karamcheti and Suraj Nair and Annie S. Chen and Thomas Kollar and Chelsea Finn and Dorsa Sadigh and Percy Liang},
  journal={ArXiv},
  year={2023},
  volume={abs/2302.12766},
  url={https://api.semanticscholar.org/CorpusID:257205716}
}

@article{Tong2022VideoMAEMA,
  title={VideoMAE: Masked Autoencoders are Data-Efficient Learners for Self-Supervised Video Pre-Training},
  author={Zhan Tong and Yibing Song and Jue Wang and Limin Wang},
  journal={ArXiv},
  year={2022},
  volume={abs/2203.12602},
  url={https://api.semanticscholar.org/CorpusID:247619234}
}

@misc{Wang2023VideoMAEVS,
  title={VideoMAE V2: Scaling Video Masked Autoencoders with Dual Masking},
  author={Limin Wang and Bingkun Huang and Zhiyu Zhao and Zhan Tong and Yinan He and Yi Wang and Yali Wang and Yu Qiao},
  journal={2023 IEEE/CVF Conference on Computer Vision and Pattern Recognition (CVPR)},
  year={2023},
  pages={14549-14560},
  url={https://api.semanticscholar.org/CorpusID:257805127}
}

@misc{Wang2022MaskedVD,
  title={Masked Video Distillation: Rethinking Masked Feature Modeling for Self-supervised Video Representation Learning},
  author={Rui Wang and Dongdong Chen and Zuxuan Wu and Yinpeng Chen and Xiyang Dai and Mengchen Liu and Lu Yuan and Yu-Gang Jiang},
  journal={2023 IEEE/CVF Conference on Computer Vision and Pattern Recognition (CVPR)},
  year={2022},
  pages={6312-6322},
  url={https://api.semanticscholar.org/CorpusID:254408955}
}

@misc{stroud2020learning,
  title={Learning video representations from textual web supervision},
  author={Stroud, Jonathan C and Lu, Zhichao and Sun, Chen and Deng, Jia and Sukthankar, Rahul and Schmid, Cordelia and Ross, David A},
  journal={arXiv preprint arXiv:2007.14937},
  year={2020}
}

@article{zellers2021merlot,
  title={Merlot: Multimodal neural script knowledge models},
  author={Zellers, Rowan and Lu, Ximing and Hessel, Jack and Yu, Youngjae and Park, Jae Sung and Cao, Jize and Farhadi, Ali and Choi, Yejin},
  journal={Advances in Neural Information Processing Systems},
  volume={34},
  pages={23634--23651},
  year={2021}
}

@misc{Girdhar2022OmniMAESM,
  title={OmniMAE: Single Model Masked Pretraining on Images and Videos},
  author={Rohit Girdhar and Alaaeldin El-Nouby and Mannat Singh and Kalyan Vasudev Alwala and Armand Joulin and Ishan Misra},
  journal={2023 IEEE/CVF Conference on Computer Vision and Pattern Recognition (CVPR)},
  year={2022},
  pages={10406-10417},
  url={https://api.semanticscholar.org/CorpusID:249712367}
}

@inproceedings{Yan2022VideoCoCaVM,
  title={VideoCoCa: Video-Text Modeling with Zero-Shot Transfer from Contrastive Captioners},
  author={Shen Yan and Tao Zhu and Zirui Wang and Yuan Cao and Mi Zhang and Soham Ghosh and Yonghui Wu and Jiahui Yu},
  year={2022},
  url={https://api.semanticscholar.org/CorpusID:254535696}
}

@article{Wang2023InternVidAL,
  title={InternVid: A Large-scale Video-Text Dataset for Multimodal Understanding and Generation},
  author={Yi Wang and Yinan He and Yizhuo Li and Kunchang Li and Jiashuo Yu and Xin Jian Ma and Xinyuan Chen and Yaohui Wang and Ping Luo and Ziwei Liu and Yali Wang and Limin Wang and Y. Qiao},
  journal={ArXiv},
  year={2023},
  volume={abs/2307.06942},
  url={https://api.semanticscholar.org/CorpusID:259847783}
}

@article{Papalampidi2023ASR,
  title={A Simple Recipe for Contrastively Pre-training Video-First Encoders Beyond 16 Frames},
  author={Pinelopi Papalampidi and Skanda Koppula and Shreya Pathak and Justin Chiu and Joseph Heyward and Viorica Patraucean and Jiajun Shen and Antoine Miech and Andrew Zisserman and Aida Nematzdeh},
  journal={ArXiv},
  year={2023},
  volume={abs/2312.07395},
  url={https://api.semanticscholar.org/CorpusID:266174654}
}

@inproceedings{Xu2021VideoCLIPCP,
  title={VideoCLIP: Contrastive Pre-training for Zero-shot Video-Text Understanding},
  author={Hu Xu and Gargi Ghosh and Po-Yao (Bernie) Huang and Dmytro Okhonko and Armen Aghajanyan and Florian Metze Luke Zettlemoyer Christoph Feichtenhofer},
  booktitle={Conference on Empirical Methods in Natural Language Processing},
  year={2021},
  url={https://api.semanticscholar.org/CorpusID:238215257}
}

@misc{Li2023UnmaskedTT,
  title={Unmasked Teacher: Towards Training-Efficient Video Foundation Models},
  author={Kunchang Li and Yali Wang and Yizhuo Li and Yi Wang and Yinan He and Limin Wang and Yu Qiao},
  journal={2023 IEEE/CVF International Conference on Computer Vision (ICCV)},
  year={2023},
  pages={19891-19903},
  url={https://api.semanticscholar.org/CorpusID:257771777}
}

@misc{bardes2023v,
  title={V-JEPA: Latent Video Prediction for Visual Representation Learning},
  author={Bardes, Adrien and Garrido, Quentin and Ponce, Jean and Chen, Xinlei and Rabbat, Michael and LeCun, Yann and Assran, Mido and Ballas, Nicolas},
  year={2023}
}

@article{Majumdar2023WhereAW,
  title={Where are we in the search for an Artificial Visual Cortex for Embodied Intelligence?},
  author={Arjun Majumdar and Karmesh Yadav and Sergio Arnaud and Yecheng Jason Ma and Claire Chen and Sneha Silwal and Aryan Jain and Vincent-Pierre Berges and P. Abbeel and Jitendra Malik and Dhruv Batra and Yixin Lin and Oleksandr Maksymets and Aravind Rajeswaran and Franziska Meier},
  journal={ArXiv},
  year={2023},
  volume={abs/2303.18240},
  url={https://api.semanticscholar.org/CorpusID:257901087}
}

@article{schuhmann2022laion,
  title={Laion-5b: An open large-scale dataset for training next generation image-text models},
  author={Schuhmann, Christoph and Beaumont, Romain and Vencu, Richard and Gordon, Cade and Wightman, Ross and Cherti, Mehdi and Coombes, Theo and Katta, Aarush and Mullis, Clayton and Wortsman, Mitchell and others},
  journal={Advances in Neural Information Processing Systems},
  volume={35},
  pages={25278--25294},
  year={2022}
}

@inproceedings{gao2022simvp,
  title={Simvp: Simpler yet better video prediction},
  author={Gao, Zhangyang and Tan, Cheng and Wu, Lirong and Li, Stan Z},
  booktitle={Proceedings of the IEEE/CVF conference on computer vision and pattern recognition},
  pages={3170--3180},
  year={2022}
}

@article{Seo2022ReinforcementLW,
  title={Reinforcement Learning with Action-Free Pre-Training from Videos},
  author={Younggyo Seo and Kimin Lee and Stephen James and P. Abbeel},
  journal={ArXiv},
  year={2022},
  volume={abs/2203.13880},
  url={https://api.semanticscholar.org/CorpusID:247762941}
}

@misc{sermanet2023robovqa,
  title={RoboVQA: Multimodal Long-Horizon Reasoning for Robotics},
  author={Sermanet, Pierre and Ding, Tianli and Zhao, Jeffrey and Xia, Fei and Dwibedi, Debidatta and Gopalakrishnan, Keerthana and Chan, Christine and Dulac-Arnold, Gabriel and Maddineni, Sharath and Joshi, Nikhil J and others},
  journal={arXiv preprint arXiv:2311.00899},
  year={2023}
}

@article{Jing2023ExploringVP,
  title={Exploring Visual Pre-training for Robot Manipulation: Datasets, Models and Methods},
  author={Ya Jing and Xuelin Zhu and Xingbin Liu and Qie Sima and Taozheng Yang and Yunhai Feng and Tao Kong},
  journal={ArXiv},
  year={2023},
  volume={abs/2308.03620},
  url={https://api.semanticscholar.org/CorpusID:254198890}
}

@article{Wang2022VRL3AD,
  title={VRL3: A Data-Driven Framework for Visual Deep Reinforcement Learning},
  author={Che Wang and Xufang Luo and Keith W. Ross and Dongsheng Li},
  journal={ArXiv},
  year={2022},
  volume={abs/2202.10324},
  url={https://api.semanticscholar.org/CorpusID:247011862}
}

@article{Hafner2023MasteringDD,
  title={Mastering Diverse Domains through World Models},
  author={Danijar Hafner and J. Paukonis and Jimmy Ba and Timothy P. Lillicrap},
  journal={ArXiv},
  year={2023},
  volume={abs/2301.04104},
  url={https://api.semanticscholar.org/CorpusID:255569874}
}

@article{Silwal2023WhatDW,
  title={What do we learn from a large-scale study of pre-trained visual representations in sim and real environments?},
  author={Sneha Silwal and Karmesh Yadav and Tingfan Wu and Jay Vakil and Arjun Majumdar and Sergio Arnaud and Claire Chen and Vincent-Pierre Berges and Dhruv Batra and Aravind Rajeswaran and Mrinal Kalakrishnan and Franziska Meier and Oleksandr Maksymets},
  journal={ArXiv},
  year={2023},
  volume={abs/2310.02219},
  url={https://api.semanticscholar.org/CorpusID:263608779}
}

@inproceedings{Rybkin2018LearningWY,
  title={Learning what you can do before doing anything},
  author={Oleh Rybkin and Karl Pertsch and Konstantinos G. Derpanis and Kostas Daniilidis and Andrew Jaegle},
  booktitle={International Conference on Learning Representations},
  year={2018},
  url={https://api.semanticscholar.org/CorpusID:60441438}
}

@inproceedings{lynch2020learning,
  title={Learning latent plans from play},
  author={Lynch, Corey and Khansari, Mohi and Xiao, Ted and Kumar, Vikash and Tompson, Jonathan and Levine, Sergey and Sermanet, Pierre},
  booktitle={Conference on robot learning},
  pages={1113--1132},
  year={2020},
  organization={PMLR}
}

@inproceedings{RoseteBeas2022LatentPF,
  title={Latent Plans for Task-Agnostic Offline Reinforcement Learning},
  author={Erick Rosete-Beas and Oier Mees and Gabriel Kalweit and Joschka Boedecker and Wolfram Burgard},
  booktitle={Conference on Robot Learning},
  year={2022},
  url={https://api.semanticscholar.org/CorpusID:252367910}
}

@article{Cui2022FromPT,
  title={From Play to Policy: Conditional Behavior Generation from Uncurated Robot Data},
  author={Zichen Jeff Cui and Yibin Wang and Nur Muhammad (Mahi) Shafiullah and Lerrel Pinto},
  journal={ArXiv},
  year={2022},
  volume={abs/2210.10047},
  url={https://api.semanticscholar.org/CorpusID:252968170}
}

@inproceedings{Schmeckpeper2019LearningPM,
  title={Learning Predictive Models From Observation and Interaction},
  author={Karl Schmeckpeper and Annie Xie and Oleh Rybkin and Stephen Tian and Kostas Daniilidis and Sergey Levine and Chelsea Finn},
  booktitle={European Conference on Computer Vision},
  year={2019},
  url={https://api.semanticscholar.org/CorpusID:209515451}
}

@misc{rong2020frankmocap,
  title={Frankmocap: Fast monocular 3d hand and body motion capture by regression and integration},
  author={Rong, Yu and Shiratori, Takaaki and Joo, Hanbyul},
  journal={arXiv preprint arXiv:2008.08324},
  year={2020}
}

@misc{Yuan2021DMotionRV,
  title={DMotion: Robotic Visuomotor Control with Unsupervised Forward Model Learned from Videos},
  author={Haoqi Yuan and Ruihai Wu and Andrew Zhao and Hanwang Zhang and Zihan Ding and Hao Dong},
  journal={2021 IEEE/RSJ International Conference on Intelligent Robots and Systems (IROS)},
  year={2021},
  pages={7135-7142},
  url={https://api.semanticscholar.org/CorpusID:232146710}
}

@article{Mendonca2023StructuredWM,
  title={Structured World Models from Human Videos},
  author={Russell Mendonca and Shikhar Bahl and Deepak Pathak},
  journal={ArXiv},
  year={2023},
  volume={abs/2308.10901},
  url={https://api.semanticscholar.org/CorpusID:259336798}
}

@article{Wu2023PretrainingCW,
  title={Pre-training Contextualized World Models with In-the-wild Videos for Reinforcement Learning},
  author={Jialong Wu and Haoyu Ma and Chao Deng and Mingsheng Long},
  journal={ArXiv},
  year={2023},
  volume={abs/2305.18499},
  url={https://api.semanticscholar.org/CorpusID:258967679}
}

@article{Baker2022VideoP,
  title={Video PreTraining (VPT): Learning to Act by Watching Unlabeled Online Videos},
  author={Bowen Baker and Ilge Akkaya and Peter Zhokhov and Joost Huizinga and Jie Tang and Adrien Ecoffet and Brandon Houghton and Raul Sampedro and Jeff Clune},
  journal={ArXiv},
  year={2022},
  volume={abs/2206.11795},
  url={https://api.semanticscholar.org/CorpusID:249953673}
}

@inproceedings{Shaw2022VideoDexLD,
  title={VideoDex: Learning Dexterity from Internet Videos},
  author={Kenneth Shaw and Shikhar Bahl and Deepak Pathak},
  booktitle={Conference on Robot Learning},
  year={2022},
  url={https://api.semanticscholar.org/CorpusID:254408735}
}

@article{Wang2023MimicPlayLI,
  title={MimicPlay: Long-Horizon Imitation Learning by Watching Human Play},
  author={Chen Wang and Linxi (Jim) Fan and Jiankai Sun and Ruohan Zhang and Li Fei-Fei and Danfei Xu and Yuke Zhu and Anima Anandkumar},
  journal={ArXiv},
  year={2023},
  volume={abs/2302.12422},
  url={https://api.semanticscholar.org/CorpusID:257205825}
}

@inproceedings{Pertsch2022CrossDomainTV,
  title={Cross-Domain Transfer via Semantic Skill Imitation},
  author={Karl Pertsch and Ruta Desai and Vikash Kumar and Franziska Meier and Joseph J. Lim and Dhruv Batra and Akshara Rai},
  booktitle={Conference on Robot Learning},
  year={2022},
  url={https://api.semanticscholar.org/CorpusID:254636470}
}

@article{Thomas2023PLEXMT,
  title={PLEX: Making the Most of the Available Data for Robotic Manipulation Pretraining},
  author={Garrett Thomas and Ching-An Cheng and Ricky Loynd and Vibhav Vineet and Mihai Jalobeanu and Andrey Kolobov},
  journal={ArXiv},
  year={2023},
  volume={abs/2303.08789},
  url={https://api.semanticscholar.org/CorpusID:257532588}
}

@article{Yuan2024GeneralFA,
  title={General Flow as Foundation Affordance for Scalable Robot Learning},
  author={Chengbo Yuan and Chuan Wen and Tong Zhang and Yang Gao},
  journal={ArXiv},
  year={2024},
  volume={abs/2401.11439},
  url={https://api.semanticscholar.org/CorpusID:267069070}
}

@article{Nasiriany2024PIVOTIV,
  title={PIVOT: Iterative Visual Prompting Elicits Actionable Knowledge for VLMs},
  author={Soroush Nasiriany and Fei Xia and Wenhao Yu and Ted Xiao and Jacky Liang and Ishita Dasgupta and Annie Xie and Danny Driess and Ayzaan Wahid and Zhuo Xu and Quan Ho Vuong and Tingnan Zhang and Tsang-Wei Edward Lee and Kuang-Huei Lee and Peng Xu and Sean Kirmani and Yuke Zhu and Andy Zeng and Karol Hausman and Nicolas Manfred Otto Heess and Chelsea Finn and Sergey Levine and Brian Ichter},
  journal={ArXiv},
  year={2024},
  volume={abs/2402.07872},
  url={https://api.semanticscholar.org/CorpusID:267627797}
}

@article{Wen2023AnypointTM,
  title={Any-point Trajectory Modeling for Policy Learning},
  author={Chuan Wen and Xingyu Lin and John So and Kai Chen and Qi Dou and Yang Gao and Pieter Abbeel},
  journal={ArXiv},
  year={2023},
  volume={abs/2401.00025},
  url={https://api.semanticscholar.org/CorpusID:266693687}
}

@misc{karaev2023cotracker,
  title={Cotracker: It is better to track together},
  author={Karaev, Nikita and Rocco, Ignacio and Graham, Benjamin and Neverova, Natalia and Vedaldi, Andrea and Rupprecht, Christian},
  journal={arXiv preprint arXiv:2307.07635},
  year={2023}
}

@article{Black2023ZeroShotRM,
  title={Zero-Shot Robotic Manipulation with Pretrained Image-Editing Diffusion Models},
  author={Kevin Black and Mitsuhiko Nakamoto and Pranav Atreya and Homer Walke and Chelsea Finn and Aviral Kumar and Sergey Levine},
  journal={ArXiv},
  year={2023},
  volume={abs/2310.10639},
  url={https://api.semanticscholar.org/CorpusID:264172455}
}

@article{Hilton2023ScalingLF,
  title={Scaling laws for single-agent reinforcement learning},
  author={Jacob Hilton and Jie Tang and John Schulman},
  journal={ArXiv},
  year={2023},
  volume={abs/2301.13442},
  url={https://api.semanticscholar.org/CorpusID:256416224}
}

@article{Sartor2024NeuralSL,
  title={Neural Scaling Laws for Embodied AI},
  author={Sebastian Sartor and Neil Thompson},
  journal={ArXiv},
  year={2024},
  volume={abs/2405.14005},
  url={https://api.semanticscholar.org/CorpusID:269983385}
}

@article{Mu2023EmbodiedGPTVP,
  title={EmbodiedGPT: Vision-Language Pre-Training via Embodied Chain of Thought},
  author={Yao Mu and Qinglong Zhang and Mengkang Hu and Wen Wang and Mingyu Ding and Jun Jin and Bin Wang and Jifeng Dai and Y. Qiao and Ping Luo},
  journal={ArXiv},
  year={2023},
  volume={abs/2305.15021},
  url={https://api.semanticscholar.org/CorpusID:258865718}
}

@article{Du2023VideoLP,
  title={Video Language Planning},
  author={Yilun Du and Mengjiao Yang and Peter R. Florence and Fei Xia and Ayzaan Wahid and Brian Ichter and Pierre Sermanet and Tianhe Yu and Pieter Abbeel and Josh Tenenbaum and Leslie Pack Kaelbling and Andy Zeng and Jonathan Tompson},
  journal={ArXiv},
  year={2023},
  volume={abs/2310.10625},
  url={https://api.semanticscholar.org/CorpusID:264172935}
}

@article{Yang2023LearningIR,
  title={Learning Interactive Real-World Simulators},
  author={Mengjiao Yang and Yilun Du and Kamyar Ghasemipour and Jonathan Tompson and Dale Schuurmans and Pieter Abbeel},
  journal={ArXiv},
  year={2023},
  volume={abs/2310.06114},
  url={https://api.semanticscholar.org/CorpusID:263830899}
}

@misc{numina_math_datasets,
    title = {NuminaMath TIR},
    author = {Li, Jia and Beeching, Edward and Tunstall, Lewis and Lipkin, Ben and Soletskyi, Roman and Huang, Shengyi Costa and Rasul, Kashif and Yu, Longhui and Jiang, Albert and Shen, Ziju and Qin, Zihan and Dong, Bin and Zhou, Li and Fleureau, Yann and Lample, Guillaume and Polu, Stanislas},
    year = {2024},
    publisher = {Numina},
    journal = {Hugging Face repository},
    howpublished = {\url{https://huggingface.co/AI-MO/NuminaMath-TIR}}
}

@misc{wang2024helpsteer2,
  title={HelpSteer2: Open-source dataset for training top-performing reward models},
  author={Wang, Zhilin and Dong, Yi and Delalleau, Olivier and Zeng, Jiaqi and Shen, Gerald and Egert, Daniel and Zhang, Jimmy J and Sreedhar, Makesh Narsimhan and Kuchaiev, Oleksii},
  journal={arXiv preprint arXiv:2406.08673},
  year={2024}
}

@article{yu2020mopo,
  title={Mopo: Model-based offline policy optimization},
  author={Yu, Tianhe and Thomas, Garrett and Yu, Lantao and Ermon, Stefano and Zou, James Y and Levine, Sergey and Finn, Chelsea and Ma, Tengyu},
  journal={Advances in Neural Information Processing Systems},
  volume={33},
  pages={14129--14142},
  year={2020}
}

@misc{achiam2023gpt,
  title={Gpt-4 technical report},
  author={Achiam, Josh and Adler, Steven and Agarwal, Sandhini and Ahmad, Lama and Akkaya, Ilge and Aleman, Florencia Leoni and Almeida, Diogo and Altenschmidt, Janko and Altman, Sam and Anadkat, Shyamal and others},
  journal={arXiv preprint arXiv:2303.08774},
  year={2023}
}

@misc{Shang2021SelfSupervisedDR,
  title={Self-Supervised Disentangled Representation Learning for Third-Person Imitation Learning},
  author={Jinghuan Shang and Michael S. Ryoo},
  journal={2021 IEEE/RSJ International Conference on Intelligent Robots and Systems (IROS)},
  year={2021},
  pages={214-221},
  url={https://api.semanticscholar.org/CorpusID:236772575}
}

@inproceedings{damen2018scaling,
  title={Scaling egocentric vision: The epic-kitchens dataset},
  author={Damen, Dima and Doughty, Hazel and Farinella, Giovanni Maria and Fidler, Sanja and Furnari, Antonino and Kazakos, Evangelos and Moltisanti, Davide and Munro, Jonathan and Perrett, Toby and Price, Will and others},
  booktitle={Proceedings of the European conference on computer vision (ECCV)},
  pages={720--736},
  year={2018}
}

@misc{damen2022rescaling,
  title={Rescaling egocentric vision: Collection, pipeline and challenges for epic-kitchens-100},
  author={Damen, Dima and Doughty, Hazel and Farinella, Giovanni Maria and Furnari, Antonino and Kazakos, Evangelos and Ma, Jian and Moltisanti, Davide and Munro, Jonathan and Perrett, Toby and Price, Will and others},
  journal={International Journal of Computer Vision},
  pages={1--23},
  year={2022},
  publisher={Springer}
}

@inproceedings{tomar2023video,
  title={Video-Guided Skill Discovery},
  author={Tomar, Manan and Ghosh, Dibya and Myers, Vivek and Dragan, Anca and Taylor, Matthew E and Bachman, Philip and Levine, Sergey},
  booktitle={ICML 2023 Workshop The Many Facets of Preference-Based Learning},
  year={2023}
}

@article{Xu2023XSkillCE,
  title={XSkill: Cross Embodiment Skill Discovery},
  author={Mengda Xu and Zhenjia Xu and Cheng Chi and Manuela M. Veloso and Shuran Song},
  journal={ArXiv},
  year={2023},
  volume={abs/2307.09955},
  url={https://api.semanticscholar.org/CorpusID:259982636}
}

@article{Smith2019AVIDLM,
  title={AVID: Learning Multi-Stage Tasks via Pixel-Level Translation of Human Videos},
  author={Laura Smith and Nikita Dhawan and Marvin Zhang and P. Abbeel and Sergey Levine},
  journal={ArXiv},
  year={2019},
  volume={abs/1912.04443},
  url={https://api.semanticscholar.org/CorpusID:209140723}
}

@article{chen2024rh20t,
  title={Rh20t-p: A primitive-level robotic dataset towards composable generalization agents},
  author={Chen, Zeren and Shi, Zhelun and Lu, Xiaoya and He, Lehan and Qian, Sucheng and Yin, Zhenfei and Ouyang, Wanli and Shao, Jing and Qiao, Yu and Lu, Cewu and others},
  journal={arXiv preprint arXiv:2403.19622},
  year={2024}
}

@article{ju2024miradata,
  title={Miradata: A large-scale video dataset with long durations and structured captions},
  author={Ju, Xuan and Gao, Yiming and Zhang, Zhaoyang and Yuan, Ziyang and Wang, Xintao and Zeng, Ailing and Xiong, Yu and Xu, Qiang and Shan, Ying},
  journal={Advances in Neural Information Processing Systems},
  volume={37},
  pages={48955--48970},
  year={2024}
}

@misc{Xiong2021LearningBW,
  title={Learning by Watching: Physical Imitation of Manipulation Skills from Human Videos},
  author={Haoyu Xiong and Quanzhou Li and Yun-Chun Chen and Homanga Bharadhwaj and Samarth Sinha and Animesh Garg},
  journal={2021 IEEE/RSJ International Conference on Intelligent Robots and Systems (IROS)},
  year={2021},
  pages={7827-7834},
  url={https://api.semanticscholar.org/CorpusID:231632575}
}

@article{Qin2022FromOH,
  title={From One Hand to Multiple Hands: Imitation Learning for Dexterous Manipulation From Single-Camera Teleoperation},
  author={Yuzhe Qin and Hao Su and Xiaolong Wang},
  journal={IEEE Robotics and Automation Letters},
  year={2022},
  volume={7},
  pages={10873-10881},
  url={https://api.semanticscholar.org/CorpusID:248392006}
}

@misc{stadie2017third,
  title={Third-person imitation learning},
  author={Stadie, Bradly C and Abbeel, Pieter and Sutskever, Ilya},
  journal={arXiv preprint arXiv:1703.01703},
  year={2017}
}

@article{Bharadhwaj2023TowardsGZ,
  title={Towards Generalizable Zero-Shot Manipulation via Translating Human Interaction Plans},
  author={Homanga Bharadhwaj and Abhi Gupta and Vikash Kumar and Shubham Tulsiani},
  journal={ArXiv},
  year={2023},
  volume={abs/2312.00775},
  url={https://api.semanticscholar.org/CorpusID:265551754}
}

@article{darkhalil2022epic,
  title={Epic-kitchens visor benchmark: Video segmentations and object relations},
  author={Darkhalil, Ahmad and Shan, Dandan and Zhu, Bin and Ma, Jian and Kar, Amlan and Higgins, Richard and Fidler, Sanja and Fouhey, David and Damen, Dima},
  journal={Advances in Neural Information Processing Systems},
  volume={35},
  pages={13745--13758},
  year={2022}
}

@inproceedings{zhang2022fine,
  title={Fine-grained egocentric hand-object segmentation: Dataset, model, and applications},
  author={Zhang, Lingzhi and Zhou, Shenghao and Stent, Simon and Shi, Jianbo},
  booktitle={European Conference on Computer Vision},
  pages={127--145},
  year={2022},
  organization={Springer}
}

@article{Peng2018SFVRL,
  title={SFV: Reinforcement Learning of Physical Skills from Videos},
  author={Xue Bin Peng and Angjoo Kanazawa and Jitendra Malik and P. Abbeel and Sergey Levine},
  journal={ACM Trans. Graph.},
  year={2018},
  volume={37},
  pages={178},
  url={https://api.semanticscholar.org/CorpusID:52937281}
}

@article{Sivakumar2022RoboticTL,
  title={Robotic Telekinesis: Learning a Robotic Hand Imitator by Watching Humans on Youtube},
  author={Aravind Sivakumar and Kenneth Shaw and Deepak Pathak},
  journal={ArXiv},
  year={2022},
  volume={abs/2202.10448},
  url={https://api.semanticscholar.org/CorpusID:247011104}
}

@inproceedings{Kumar2022GraphIR,
  title={Graph Inverse Reinforcement Learning from Diverse Videos},
  author={Sateesh Kumar and Jonathan Zamora and Nicklas Hansen and Rishabh Jangir and Xiaolong Wang},
  booktitle={Conference on Robot Learning},
  year={2022},
  url={https://api.semanticscholar.org/CorpusID:251223373}
}

@article{Nagarajan2021ShapingEA,
  title={Shaping embodied agent behavior with activity-context priors from egocentric video},
  author={Tushar Nagarajan and Kristen Grauman},
  journal={ArXiv},
  year={2021},
  volume={abs/2110.07692},
  url={https://api.semanticscholar.org/CorpusID:239009498}
}

@misc{Karnan2021VOILAVI,
  title={VOILA: Visual-Observation-Only Imitation Learning for Autonomous Navigation},
  author={Haresh Karnan and Garrett Warnell and Xuesu Xiao and Peter Stone},
  journal={2022 International Conference on Robotics and Automation (ICRA)},
  year={2021},
  pages={2497-2503},
  url={https://api.semanticscholar.org/CorpusID:234790310}
}

@article{Brohan2023RT2VM,
  title={RT-2: Vision-Language-Action Models Transfer Web Knowledge to Robotic Control},
  author={Anthony Brohan and Noah Brown and Justice Carbajal and Yevgen Chebotar and Krzysztof Choromanski and Tianli Ding and Danny Driess and Chelsea Finn and Peter R. Florence and Chuyuan Fu and Montse Gonzalez Arenas and Keerthana Gopalakrishnan and Kehang Han and Karol Hausman and Alexander Herzog and Jasmine Hsu and Brian Ichter and Alex Irpan and Nikhil J. Joshi and Ryan C. Julian and Dmitry Kalashnikov and Yuheng Kuang and Isabel Leal and Sergey Levine and Henryk Michalewski and Igor Mordatch and Karl Pertsch and Kanishka Rao and Krista Reymann and Michael S. Ryoo and Grecia Salazar and Pannag R. Sanketi and Pierre Sermanet and Jaspiar Singh and Anika Singh and Radu Soricut and Huong Tran and Vincent Vanhoucke and Quan Ho Vuong and Ayzaan Wahid and Stefan Welker and Paul Wohlhart and Ted Xiao and Tianhe Yu and Brianna Zitkovich},
  journal={ArXiv},
  year={2023},
  volume={abs/2307.15818},
  url={https://api.semanticscholar.org/CorpusID:260293142}
}

@inproceedings{liang2023code,
  title={Code as policies: Language model programs for embodied control},
  author={Liang, Jacky and Huang, Wenlong and Xia, Fei and Xu, Peng and Hausman, Karol and Ichter, Brian and Florence, Pete and Zeng, Andy},
  booktitle={2023 IEEE International Conference on Robotics and Automation (ICRA)},
  pages={9493--9500},
  year={2023},
  organization={IEEE}
}

@inproceedings{shah2023lm,
  title={Lm-nav: Robotic navigation with large pre-trained models of language, vision, and action},
  author={Shah, Dhruv and Osi{\'n}ski, B{\l}a{\.z}ej and Levine, Sergey and others},
  booktitle={Conference on robot learning},
  pages={492--504},
  year={2023},
  organization={PMLR}
}

@article{Kim2023GivingRA,
  title={Giving Robots a Hand: Learning Generalizable Manipulation with Eye-in-Hand Human Video Demonstrations},
  author={Moo Jin Kim and Jiajun Wu and Chelsea Finn},
  journal={ArXiv},
  year={2023},
  volume={abs/2307.05959},
  url={https://api.semanticscholar.org/CorpusID:259836885}
}

@article{park2024hiql,
  title={Hiql: Offline goal-conditioned rl with latent states as actions},
  author={Park, Seohong and Ghosh, Dibya and Eysenbach, Benjamin and Levine, Sergey},
  journal={Advances in Neural Information Processing Systems},
  volume={36},
  year={2024}
}

@inproceedings{Torabi2018BehavioralCF,
  title={Behavioral Cloning from Observation},
  author={Faraz Torabi and Garrett Warnell and Peter Stone},
  booktitle={International Joint Conference on Artificial Intelligence},
  year={2018},
  url={https://api.semanticscholar.org/CorpusID:23206414}
}

@article{Chang2020SemanticVN,
  title={Semantic Visual Navigation by Watching YouTube Videos},
  author={Matthew Chang and Arjun Gupta and Saurabh Gupta},
  journal={ArXiv},
  year={2020},
  volume={abs/2006.10034},
  url={https://api.semanticscholar.org/CorpusID:219721405}
}

@book{sutton2018reinforcement,
  title={Reinforcement learning: An introduction},
  author={Sutton, Richard S and Barto, Andrew G},
  year={2018},
  publisher={MIT press}
}

@misc{guan2024task,
  title={" Task Success" is not Enough: Investigating the Use of Video-Language Models as Behavior Critics for Catching Undesirable Agent Behaviors},
  author={Guan, Lin and Zhou, Yifan and Liu, Denis and Zha, Yantian and Amor, Heni Ben and Kambhampati, Subbarao},
  journal={arXiv preprint arXiv:2402.04210},
  year={2024}
}

@article{Chang2022LearningVF,
  title={Learning Value Functions from Undirected State-only Experience},
  author={Matthew Chang and Arjun Gupta and Saurabh Gupta},
  journal={ArXiv},
  year={2022},
  volume={abs/2204.12458},
  url={https://api.semanticscholar.org/CorpusID:245064979}
}

@article{Ali2022UnmannedAV,
  title={Unmanned Aerial Vehicles: A Literature Review},
  author={Shaaban Ali and Osama Hassan and Anand Gopalakrishnan and Aboobacker Sidheeq Varamb Muriyan and Sobers Lx Francis},
  journal={Journal of Hunan University Natural Sciences},
  year={2022},
  url={https://api.semanticscholar.org/CorpusID:252183358}
}

@article{Biswal2020DevelopmentOQ,
  title={Development of quadruped walking robots: A review},
  author={Priyaranjan Biswal and Prases Kumar Mohanty},
  journal={Ain Shams Engineering Journal},
  year={2020},
  url={https://api.semanticscholar.org/CorpusID:229445610}
}

@article{Urrea2025,
  author    = {Claudio Urrea and John Kern},
  title     = {Recent Advances and Challenges in Industrial Robotics: A Systematic Review of Technological Trends and Emerging Applications},
  journal   = {Processes},
  volume    = {13},
  number    = {3},
  pages     = {832},
  year      = {2025},
  doi       = {10.3390/pr13030832},
  url       = {https://www.mdpi.com/2227-9717/13/3/832}
}

@article{Schilling1995MobileRF,
  title={Mobile robots for planetary exploration},
  author={Klaus Schilling and Christoph C. Jungius},
  journal={Control Engineering Practice},
  year={1995},
  volume={4},
  pages={513-524},
  url={https://api.semanticscholar.org/CorpusID:109713134}
}

@article{Edwards2018ImitatingLP,
  title={Imitating Latent Policies from Observation},
  author={Ashley D. Edwards and Himanshu Sahni and Yannick Schroecker and Charles Lee Isbell},
  journal={ArXiv},
  year={2018},
  volume={abs/1805.07914},
  url={https://api.semanticscholar.org/CorpusID:29156793}
}

@inproceedings{Edwards2020EstimatingQS,
  title={Estimating Q(s, s') with Deep Deterministic Dynamics Gradients},
  author={Ashley D. Edwards and Himanshu Sahni and Rosanne Liu and Jane Hung and Ankit Jain and Rui Wang and Adrien Ecoffet and Thomas Miconi and Charles Lee Isbell and Jason Yosinski},
  booktitle={International Conference on Machine Learning},
  year={2020},
  url={https://api.semanticscholar.org/CorpusID:211258729}
}

@inproceedings{Pertsch2019KeyframingTF,
  title={Keyframing the Future: Keyframe Discovery for Visual Prediction and Planning},
  author={Karl Pertsch and Oleh Rybkin and Jingyun Yang and Shenghao Zhou and Konstantinos G. Derpanis and Kostas Daniilidis and Joseph J. Lim and Andrew Jaegle},
  booktitle={Conference on Learning for Dynamics \& Control},
  year={2019},
  url={https://api.semanticscholar.org/CorpusID:218571383}
}

@article{Lifshitz2023STEVE1AG,
  title={STEVE-1: A Generative Model for Text-to-Behavior in Minecraft},
  author={Shalev Lifshitz and Keiran Paster and Harris Chan and Jimmy Ba and Sheila A. McIlraith},
  journal={ArXiv},
  year={2023},
  volume={abs/2306.00937},
  url={https://api.semanticscholar.org/CorpusID:258999563}
}

@article{Cai2023GROOTLT,
  title={GROOT: Learning to Follow Instructions by Watching Gameplay Videos},
  author={Shaofei Cai and Bowei Zhang and Zihao Wang and Xiaojian Ma and Anji Liu and Yitao Liang},
  journal={ArXiv},
  year={2023},
  volume={abs/2310.08235},
  url={https://api.semanticscholar.org/CorpusID:263908999}
}

@article{Adeniji2023LanguageRM,
  title={Language Reward Modulation for Pretraining Reinforcement Learning},
  author={Ademi Adeniji and Amber Xie and Carmelo Sferrazza and Younggyo Seo and Stephen James and P. Abbeel},
  journal={ArXiv},
  year={2023},
  volume={abs/2308.12270},
  url={https://api.semanticscholar.org/CorpusID:261075941}
}

@misc{shafiullah2023bringing,
  title={On bringing robots home},
  author={Shafiullah, Nur Muhammad Mahi and Rai, Anant and Etukuru, Haritheja and Liu, Yiqian and Misra, Ishan and Chintala, Soumith and Pinto, Lerrel},
  journal={arXiv preprint arXiv:2311.16098},
  year={2023}
}

@misc{ChaneSane2023LearningVP,
  title={Learning Video-Conditioned Policies for Unseen Manipulation Tasks},
  author={Elliot Chane-Sane and Cordelia Schmid and Ivan Laptev},
  journal={2023 IEEE International Conference on Robotics and Automation (ICRA)},
  year={2023},
  pages={909-916},
  url={https://api.semanticscholar.org/CorpusID:258588267}
}

@article{Ye2023FoundationRL,
  title={Foundation Reinforcement Learning: towards Embodied Generalist Agents with Foundation Prior Assistance},
  author={Weirui Ye and Yunsheng Zhang and Mengchen Wang and Shengjie Wang and Xianfan Gu and Pieter Abbeel and Yang Gao},
  journal={ArXiv},
  year={2023},
  volume={abs/2310.02635},
  url={https://api.semanticscholar.org/CorpusID:263620344}
}

@inproceedings{xu2022gmflow,
  title={Gmflow: Learning optical flow via global matching},
  author={Xu, Haofei and Zhang, Jing and Cai, Jianfei and Rezatofighi, Hamid and Tao, Dacheng},
  booktitle={Proceedings of the IEEE/CVF conference on computer vision and pattern recognition},
  pages={8121--8130},
  year={2022}
}

@article{Ko2023LearningTA,
  title={Learning to Act from Actionless Videos through Dense Correspondences},
  author={Po-Chen Ko and Jiayuan Mao and Yilun Du and Shao-Hua Sun and Josh Tenenbaum},
  journal={ArXiv},
  year={2023},
  volume={abs/2310.08576},
  url={https://api.semanticscholar.org/CorpusID:263908842}
}

@inproceedings{Bruce2024GenieGI,
  title={Genie: Generative Interactive Environments},
  author={Jake Bruce and Michael Dennis and Ashley Edwards and Jack Parker-Holder and Yuge Shi and Edward Hughes and Matthew Lai and Aditi Mavalankar and Richie Steigerwald and Chris Apps and Yusuf Aytar and Sarah Bechtle and Feryal M. P. Behbahani and Stephanie Chan and Nicolas Manfred Otto Heess and Lucy Gonzalez and Simon Osindero and Sherjil Ozair and Scott Reed and Jingwei Zhang and Konrad Zolna and Jeff Clune and Nando de Freitas and Satinder Singh and Tim Rocktaschel},
  year={2024},
  url={https://api.semanticscholar.org/CorpusID:267897982}
}

@inproceedings{Radosavovic2022RealWorldRL,
  title={Real-World Robot Learning with Masked Visual Pre-training},
  author={Ilija Radosavovic and Tete Xiao and Stephen James and P. Abbeel and Jitendra Malik and Trevor Darrell},
  booktitle={Conference on Robot Learning},
  year={2022},
  url={https://api.semanticscholar.org/CorpusID:252718704}
}

@misc{jin2023unified,
  title={Unified language-vision pretraining with dynamic discrete visual tokenization},
  author={Jin, Yang and Xu, Kun and Chen, Liwei and Liao, Chao and Tan, Jianchao and Chen, Bin and Lei, Chenyi and Liu, An and Song, Chengru and Lei, Xiaoqiang and others},
  journal={arXiv preprint arXiv:2309.04669},
  year={2023}
}

@article{Yang2023ProbabilisticAO,
  title={Probabilistic Adaptation of Text-to-Video Models},
  author={Mengjiao Yang and Yilun Du and Bo Dai and Dale Schuurmans and Joshua B. Tenenbaum and P. Abbeel},
  journal={ArXiv},
  year={2023},
  volume={abs/2306.01872},
  url={https://api.semanticscholar.org/CorpusID:259075709}
}

@article{Ajay2023CompositionalFM,
  title={Compositional Foundation Models for Hierarchical Planning},
  author={Anurag Ajay and Seung-Jun Han and Yilun Du and Shaung Li and Abhishek Gupta and T. Jaakkola and Josh Tenenbaum and Leslie Pack Kaelbling and Akash Srivastava and Pulkit Agrawal},
  journal={ArXiv},
  year={2023},
  volume={abs/2309.08587},
  url={https://api.semanticscholar.org/CorpusID:262012485}
}

@article{Schmidt2023LearningTA,
  title={Learning to Act without Actions},
  author={Dominik Schmidt and Minqi Jiang},
  journal={ArXiv},
  year={2023},
  volume={abs/2312.10812},
  url={https://api.semanticscholar.org/CorpusID:266359570}
}

@misc{gupta2022maskvit,
  title={Maskvit: Masked visual pre-training for video prediction},
  author={Gupta, Agrim and Tian, Stephen and Zhang, Yunzhi and Wu, Jiajun and Mart{\'\i}n-Mart{\'\i}n, Roberto and Fei-Fei, Li},
  journal={arXiv preprint arXiv:2206.11894},
  year={2022}
}

@misc{Chang2022MaskGITMG,
  title={MaskGIT: Masked Generative Image Transformer},
  author={Huiwen Chang and Han Zhang and Lu Jiang and Ce Liu and William T. Freeman},
  journal={2022 IEEE/CVF Conference on Computer Vision and Pattern Recognition (CVPR)},
  year={2022},
  pages={11305-11315},
  url={https://api.semanticscholar.org/CorpusID:246680316}
}

@article{Hu2023GAIA1AG,
  title={GAIA-1: A Generative World Model for Autonomous Driving},
  author={Anthony Hu and Lloyd Russell and Hudson Yeo and Zak Murez and George Fedoseev and Alex Kendall and Jamie Shotton and Gianluca Corrado},
  journal={ArXiv},
  year={2023},
  volume={abs/2309.17080},
  url={https://api.semanticscholar.org/CorpusID:263310665}
}

@inproceedings{ghosh2023reinforcement,
  title={Reinforcement learning from passive data via latent intentions},
  author={Ghosh, Dibya and Bhateja, Chethan Anand and Levine, Sergey},
  booktitle={International Conference on Machine Learning},
  pages={11321--11339},
  year={2023},
  organization={PMLR}
}

@article{Du2024CompositionalGM,
  title={Compositional Generative Modeling: A Single Model is Not All You Need},
  author={Yilun Du and Leslie Pack Kaelbling},
  journal={ArXiv},
  year={2024},
  volume={abs/2402.01103},
  url={https://api.semanticscholar.org/CorpusID:267406745}
}

@article{Bhateja2023RoboticOR,
  title={Robotic Offline RL from Internet Videos via Value-Function Pre-Training},
  author={Chethan Bhateja and Derek Guo and Dibya Ghosh and Anika Singh and Manan Tomar and Quan Ho Vuong and Yevgen Chebotar and Sergey Levine and Aviral Kumar},
  journal={ArXiv},
  year={2023},
  volume={abs/2309.13041},
  url={https://api.semanticscholar.org/CorpusID:262217278}
}

@misc{kaplan2020scaling,
  title={Scaling laws for neural language models},
  author={Kaplan, Jared and McCandlish, Sam and Henighan, Tom and Brown, Tom B and Chess, Benjamin and Child, Rewon and Gray, Scott and Radford, Alec and Wu, Jeffrey and Amodei, Dario},
  journal={arXiv preprint arXiv:2001.08361},
  year={2020}
}

@article{Ho2020DenoisingDP,
  title={Denoising Diffusion Probabilistic Models},
  author={Jonathan Ho and Ajay Jain and P. Abbeel},
  journal={ArXiv},
  year={2020},
  volume={abs/2006.11239},
  url={https://api.semanticscholar.org/CorpusID:219955663}
}

@misc{zeng2022socratic,
  title={Socratic models: Composing zero-shot multimodal reasoning with language},
  author={Zeng, Andy and Attarian, Maria and Ichter, Brian and Choromanski, Krzysztof and Wong, Adrian and Welker, Stefan and Tombari, Federico and Purohit, Aveek and Ryoo, Michael and Sindhwani, Vikas and others},
  journal={arXiv preprint arXiv:2204.00598},
  year={2022}
}

@article{Tam2022SemanticEF,
  title={Semantic Exploration from Language Abstractions and Pretrained Representations},
  author={Allison C. Tam and Neil C. Rabinowitz and Andrew Kyle Lampinen and Nicholas A. Roy and Stephanie C. Y. Chan and DJ Strouse and Jane X. Wang and Andrea Banino and Felix Hill},
  journal={ArXiv},
  year={2022},
  volume={abs/2204.05080},
  url={https://api.semanticscholar.org/CorpusID:248085427}
}

@inproceedings{Huang2022InnerME,
  title={Inner Monologue: Embodied Reasoning through Planning with Language Models},
  author={Wenlong Huang and F. Xia and Ted Xiao and Harris Chan and Jacky Liang and Peter R. Florence and Andy Zeng and Jonathan Tompson and Igor Mordatch and Yevgen Chebotar and Pierre Sermanet and Noah Brown and Tomas Jackson and Linda Luu and Sergey Levine and Karol Hausman and Brian Ichter},
  booktitle={Conference on Robot Learning},
  year={2022},
  url={https://api.semanticscholar.org/CorpusID:250451569}
}

@misc{chen2023video,
  title={Video chatcaptioner: Towards the enriched spatiotemporal descriptions},
  author={Chen, Jun and Zhu, Deyao and Haydarov, Kilichbek and Li, Xiang and Elhoseiny, Mohamed},
  journal={arXiv preprint arXiv:2304.04227},
  year={2023}
}

@misc{team2023octo,
  title={Octo: An open-source generalist robot policy},
  author={Team, Octo Model and Ghosh, Dibya and Walke, Homer and Pertsch, Karl and Black, Kevin and Mees, Oier and Dasari, Sudeep and Hejna, Joey and Xu, Charles and Luo, Jianlan and others},
  year={2023}
}

@article{deitke2022,
  title={ProcTHOR: Large-Scale Embodied AI Using Procedural Generation},
  author={Deitke, Matt and VanderBilt, Eli and Herrasti, Alvaro and Weihs, Luca and Ehsani, Kiana and Salvador, Jordi and Han, Winson and Kolve, Eric and Kembhavi, Aniruddha and Mottaghi, Roozbeh},
  journal={Advances in Neural Information Processing Systems},
  volume={35},
  pages={5982--5994},
  year={2022}
}

@inproceedings{ha2023scaling,
  title={Scaling up and distilling down: Language-guided robot skill acquisition},
  author={Ha, Huy and Florence, Pete and Song, Shuran},
  booktitle={Conference on Robot Learning},
  pages={3766--3777},
  year={2023},
  organization={PMLR}
}

@article{Goodfellow2014ExplainingAH,
  title={Explaining and Harnessing Adversarial Examples},
  author={Ian J. Goodfellow and Jonathon Shlens and Christian Szegedy},
  journal={CoRR},
  year={2014},
  volume={abs/1412.6572},
  url={https://api.semanticscholar.org/CorpusID:6706414}
}

@inproceedings{bharadhwaj2024position,
  title={Position: Scaling Simulation is Neither Necessary Nor Sufficient for In-the-Wild Robot Manipulation},
  author={Bharadhwaj, Homanga},
  booktitle={Forty-first International Conference on Machine Learning},
  year={2024}
}

@inproceedings{wan2023unidexgrasp++,
  title={Unidexgrasp++: Improving dexterous grasping policy learning via geometry-aware curriculum and iterative generalist-specialist learning},
  author={Wan, Weikang and Geng, Haoran and Liu, Yun and Shan, Zikang and Yang, Yaodong and Yi, Li and Wang, He},
  booktitle={Proceedings of the IEEE/CVF International Conference on Computer Vision},
  pages={3891--3902},
  year={2023}
}

@article{schrittwieser2020mastering,
  title={Mastering atari, go, chess and shogi by planning with a learned model},
  author={Schrittwieser, Julian and Antonoglou, Ioannis and Hubert, Thomas and Simonyan, Karen and Sifre, Laurent and Schmitt, Simon and Guez, Arthur and Lockhart, Edward and Hassabis, Demis and Graepel, Thore and others},
  journal={Nature},
  volume={588},
  number={7839},
  pages={604--609},
  year={2020},
  publisher={Nature Publishing Group}
}

@misc{Yang2024SimEndoGSED,
  title={SimEndoGS: Efficient Data-driven Scene Simulation using Robotic Surgery Videos via Physics-embedded 3D Gaussians},
  author={Zhenya Yang and Kai Chen and Yonghao Long and Qi Dou},
  year={2024},
  url={https://api.semanticscholar.org/CorpusID:269502486}
}

@misc{tang2024automate,
  title={AutoMate: Specialist and Generalist Assembly Policies over Diverse Geometries},
  author={Tang, Bingjie and Akinola, Iretiayo and Xu, Jie and Wen, Bowen and Handa, Ankur and Van Wyk, Karl and Fox, Dieter and Sukhatme, Gaurav S and Ramos, Fabio and Narang, Yashraj},
  journal={arXiv preprint arXiv:2407.08028},
  year={2024}
}

@inproceedings{Horan2021WhenIU,
  title={When is Unsupervised Disentanglement Possible?},
  author={Daniel P. Horan and Eitan Richardson and Yair Weiss},
  year={2021},
  url={https://api.semanticscholar.org/CorpusID:248497880}
}

@article{ParkerHolder2022EvolvingCW,
  title={Evolving Curricula with Regret-Based Environment Design},
  author={Jack Parker-Holder and Minqi Jiang and Michael Dennis and Mikayel Samvelyan and Jakob Nicolaus Foerster and Edward Grefenstette and Tim Rocktaschel},
  journal={ArXiv},
  year={2022},
  volume={abs/2203.01302},
  url={https://api.semanticscholar.org/CorpusID:247218125}
}

@misc{Ye2024LatentAP,
  title={Latent Action Pretraining from Videos},
  author={Seonghyeon Ye and Joel Jang and Byeongguk Jeon and Se June Joo and Jianwei Yang and Baolin Peng and Ajay Mandlekar and Reuben Tan and Yu-Wei Chao and Bill Yuchen Lin and Lars Lid{\'e}n and Kimin Lee and Jianfeng Gao and Luke Zettlemoyer and Dieter Fox and Minjoon Seo},
  year={2024},
  url={https://api.semanticscholar.org/CorpusID:273351190}
}

@book{moravec1988,
  author    = {Moravec, Hans},
  title     = {Mind Children: The Future of Robot and Human Intelligence},
  year      = {1988},
  publisher = {Harvard University Press},
  address   = {Cambridge, MA},
}

@article{Hendrycks2019BenchmarkingNN,
  title={Benchmarking Neural Network Robustness to Common Corruptions and Perturbations},
  author={Dan Hendrycks and Thomas G. Dietterich},
  journal={ArXiv},
  year={2019},
  volume={abs/1903.12261},
  url={https://api.semanticscholar.org/CorpusID:56657912}
}

@article{kumar2024robohive,
  title={RoboHive: A Unified Framework for Robot Learning},
  author={Kumar, Vikash and Shah, Rutav and Zhou, Gaoyue and Moens, Vincent and Caggiano, Vittorio and Gupta, Abhishek and Rajeswaran, Aravind},
  journal={Advances in Neural Information Processing Systems},
  volume={36},
  year={2024}
}

@article{liu2024libero,
  title={Libero: Benchmarking knowledge transfer for lifelong robot learning},
  author={Liu, Bo and Zhu, Yifeng and Gao, Chongkai and Feng, Yihao and Liu, Qiang and Zhu, Yuke and Stone, Peter},
  journal={Advances in Neural Information Processing Systems},
  volume={36},
  year={2024}
}

@misc{gu2023maniskill2,
  title={Maniskill2: A unified benchmark for generalizable manipulation skills},
  author={Gu, Jiayuan and Xiang, Fanbo and Li, Xuanlin and Ling, Zhan and Liu, Xiqiang and Mu, Tongzhou and Tang, Yihe and Tao, Stone and Wei, Xinyue and Yao, Yunchao and others},
  journal={arXiv preprint arXiv:2302.04659},
  year={2023}
}

@misc{madan2024foundation,
  title={Foundation Models for Video Understanding: A Survey},
  author={Madan, Neelu and Moegelmose, Andreas and Modi, Rajat and Rawat, Yogesh S and Moeslund, Thomas B},
  journal={arXiv preprint arXiv:2405.03770},
  year={2024}
}

@misc{gavenski2024imitation,
  title={Imitation Learning: A Survey of Learning Methods, Environments and Metrics},
  author={Gavenski, Nathan and Rodrigues, Odinaldo and Luck, Michael},
  journal={arXiv preprint arXiv:2404.19456},
  year={2024}
}

@misc{sutton2019bitter,
  title={The Bitter Lesson},
  author={Richard Sutton},
  year={2019},
  journal={Incomplete Ideas},
  url={http://www.incompleteideas.net/IncIdeas/BitterLesson.html}
}

@inproceedings{yu2020meta,
  title={Meta-world: A benchmark and evaluation for multi-task and meta reinforcement learning},
  author={Yu, Tianhe and Quillen, Deirdre and He, Zhanpeng and Julian, Ryan and Hausman, Karol and Finn, Chelsea and Levine, Sergey},
  booktitle={Conference on robot learning},
  pages={1094--1100},
  year={2020},
  organization={PMLR}
}

@inproceedings{ahn2020robel,
  title={Robel: Robotics benchmarks for learning with low-cost robots},
  author={Ahn, Michael and Zhu, Henry and Hartikainen, Kristian and Ponte, Hugo and Gupta, Abhishek and Levine, Sergey and Kumar, Vikash},
  booktitle={Conference on robot learning},
  pages={1300--1313},
  year={2020},
  organization={PMLR}
}

@misc{kolve2017ai2,
  title={Ai2-thor: An interactive 3d environment for visual ai},
  author={Kolve, Eric and Mottaghi, Roozbeh and Han, Winson and VanderBilt, Eli and Weihs, Luca and Herrasti, Alvaro and Deitke, Matt and Ehsani, Kiana and Gordon, Daniel and Zhu, Yuke and others},
  journal={arXiv preprint arXiv:1712.05474},
  year={2017}
}

@misc{puig2023habitat,
  title={Habitat 3.0: A co-habitat for humans, avatars and robots},
  author={Puig, Xavier and Undersander, Eric and Szot, Andrew and Cote, Mikael Dallaire and Yang, Tsung-Yen and Partsey, Ruslan and Desai, Ruta and Clegg, Alexander William and Hlavac, Michal and Min, So Yeon and others},
  journal={arXiv preprint arXiv:2310.13724},
  year={2023}
}

@misc{xie2023decomposing,
  title={Decomposing the generalization gap in imitation learning for visual robotic manipulation},
  author={Xie, Annie and Lee, Lisa and Xiao, Ted and Finn, Chelsea},
  journal={arXiv preprint arXiv:2307.03659},
  year={2023}
}

@misc{lynch2023interactive,
  title={Interactive language: Talking to robots in real time},
  author={Lynch, Corey and Wahid, Ayzaan and Tompson, Jonathan and Ding, Tianli and Betker, James and Baruch, Robert and Armstrong, Travis and Florence, Pete},
  journal={IEEE Robotics and Automation Letters},
  year={2023},
  publisher={IEEE}
}

@article{mees2022calvin,
  title={Calvin: A benchmark for language-conditioned policy learning for long-horizon robot manipulation tasks},
  author={Mees, Oier and Hermann, Lukas and Rosete-Beas, Erick and Burgard, Wolfram},
  journal={IEEE Robotics and Automation Letters},
  volume={7},
  number={3},
  pages={7327--7334},
  year={2022},
  publisher={IEEE}
}

@misc{Tobin2017DomainRF,
  title={Domain randomization for transferring deep neural networks from simulation to the real world},
  author={Joshua Tobin and Rachel Fong and Alex Ray and Jonas Schneider and Wojciech Zaremba and P. Abbeel},
  journal={2017 IEEE/RSJ International Conference on Intelligent Robots and Systems (IROS)},
  year={2017},
  pages={23-30},
  url={https://api.semanticscholar.org/CorpusID:2413610}
}

@inproceedings{zhao2020sim,
  title={Sim-to-real transfer in deep reinforcement learning for robotics: a survey},
  author={Zhao, Wenshuai and Queralta, Jorge Pe{\~n}a and Westerlund, Tomi},
  booktitle={2020 IEEE symposium series on computational intelligence (SSCI)},
  pages={737--744},
  year={2020},
  organization={IEEE}
}

@article{kaufmann2023champion,
  title={Champion-level drone racing using deep reinforcement learning},
  author={Kaufmann, Elia and Bauersfeld, Leonard and Loquercio, Antonio and M{\"u}ller, Matthias and Koltun, Vladlen and Scaramuzza, Davide},
  journal={Nature},
  volume={620},
  number={7976},
  pages={982--987},
  year={2023},
  publisher={Nature Publishing Group UK London}
}

@misc{zhuang2023robot,
  title={Robot parkour learning},
  author={Zhuang, Ziwen and Fu, Zipeng and Wang, Jianren and Atkeson, Christopher and Schwertfeger, Soeren and Finn, Chelsea and Zhao, Hang},
  journal={arXiv preprint arXiv:2309.05665},
  year={2023}
}

@misc{akkaya2019solving,
  title={Solving rubik's cube with a robot hand},
  author={Akkaya, Ilge and Andrychowicz, Marcin and Chociej, Maciek and Litwin, Mateusz and McGrew, Bob and Petron, Arthur and Paino, Alex and Plappert, Matthias and Powell, Glenn and Ribas, Raphael and others},
  journal={arXiv preprint arXiv:1910.07113},
  year={2019}
}

@misc{bai2022constitutional,
  title={Constitutional ai: Harmlessness from ai feedback},
  author={Bai, Yuntao and Kadavath, Saurav and Kundu, Sandipan and Askell, Amanda and Kernion, Jackson and Jones, Andy and Chen, Anna and Goldie, Anna and Mirhoseini, Azalia and McKinnon, Cameron and others},
  journal={arXiv preprint arXiv:2212.08073},
  year={2022}
}

@article{ouyang2022training,
  title={Training language models to follow instructions with human feedback},
  author={Ouyang, Long and Wu, Jeffrey and Jiang, Xu and Almeida, Diogo and Wainwright, Carroll and Mishkin, Pamela and Zhang, Chong and Agarwal, Sandhini and Slama, Katarina and Ray, Alex and others},
  journal={Advances in neural information processing systems},
  volume={35},
  pages={27730--27744},
  year={2022}
}

@misc{videoworldsimulators2024,
  title={Video generation models as world simulators},
  author={Tim Brooks and Bill Peebles and Connor Holmes and Will DePue and Yufei Guo and Li Jing and David Schnurr and Joe Taylor and Troy Luhman and Eric Luhman and Clarence Ng and Ricky Wang and Aditya Ramesh},
  year={2024},
  url={https://openai.com/research/video-generation-models-as-world-simulators},
}

\end{document}